%% file: main.tex
\documentclass[10pt,twocolumn,letterpaper]{article}

\PassOptionsToPackage{table}{xcolor}
\PassOptionsToPackage{breaklinks,colorlinks}{hyperref}

\usepackage[pagenumbers]{cvpr}

\input{preamble}

\usepackage{hyperref}
\definecolor{cvprblue}{rgb}{0.21,0.49,0.74}
\hypersetup{allcolors=cvprblue}

\def\paperID{xxx}
\def\confName{CVPR}
\def\confYear{2026}

\title{Hidden Monotonicity: Explaining Deep Neural Networks via their DC Decomposition}

\author{Jakob Paul Zimmermann\\
Technical University Berlin\\
{\tt\small \href{mailto:jakob.paul.zimmermann@campus.tu-berlin.de}{jakob.paul.zimmermann@campus.tu-berlin.de}}
\and
Georg Loho\\
Freie Universit\"at Berlin,\\University of Twente\\
{\tt\small \href{mailto:georg.loho@math.fu-berlin.de}{georg.loho@math.fu-berlin.de}}
}

\begin{document}
\maketitle
\begin{abstract}
  It has been demonstrated in various contexts that monotonicity leads to better explainability in neural networks.
  However, not every function can be well approximated by a monotone neural network.
  We demonstrate that monotonicity can still be used in two ways to boost explainability.

  First, we use an adaptation of the decomposition of a trained ReLU network into two monotone and convex parts, thereby overcoming numerical obstacles from an inherent blowup of the weights in this procedure.
  Our proposed saliency methods -- SplitCAM and SplitLRP --
  improve on
  state of the art results on both VGG16 and Resnet18 networks on ImageNet-S across all Quantus saliency metric categories.

  Second, we exhibit that training a model as the difference between two monotone neural networks results in a system with strong self-explainability properties.
\end{abstract}

\input{sections/introduction}
\input{sections/preliminaries}
\input{sections/method}
\input{sections/DICs}
\input{sections/table_main_generated_vgg}

\input{sections/methods_xai}
\input{sections/experiments}
\input{sections/conclusion}

\nocite{Amos2017}
\nocite{Brandenburg2024b}
\nocite{Haase2023}
\nocite{Arora2018}
\nocite{Sankaranarayanan2021CDINN}
\nocite{Zhang2018}
\nocite{Ziegler1995}

\bibliographystyle{ieeenat_fullname}
\bibliography{references}

\clearpage
\onecolumn
\setlength{\columnwidth}{\textwidth}
\setlength{\linewidth}{\textwidth}

\appendix \label{app}

\input{appendix/mathematical_perspective}
\clearpage
\input{appendix/saliency_analysis}
\clearpage
\input{appendix/ablation_studies}
\clearpage
\input{appendix/dcinns}
\clearpage
\input{appendix/maxout_networks}
\clearpage
\input{appendix/proofs}

\section*{References (Appendix)}
\addcontentsline{toc}{section}{References (Appendix)}

\appbibitem{AppAdebayo2018}{Julius Adebayo, Justin Gilmer, Michael Muelly, Ian Goodfellow, Moritz Hardt, and Been Kim. Sanity checks for saliency maps. In \emph{Advances in Neural Information Processing Systems}. Curran Associates, Inc., 2018.}

\appbibitem{AppBinder2023}{Alexander Binder, Leander Weber, Sebastian Lapuschkin, Gr\'egoire Montavon, Klaus-Robert M\"uller, and Wojciech Samek. Shortcomings of top-down randomization-based sanity checks for evaluations of deep neural network explanations. In \emph{Proceedings of the IEEE/CVF Conference on Computer Vision and Pattern Recognition (CVPR)}, pages 16143--16152, June 2023.}

\appbibitem{AppAmos2017}{Brandon Amos, Lei Xu, and J.~Zico Kolter. Input convex neural networks. In \emph{Proceedings of the 34th International Conference on Machine Learning}, pages 146--155. PMLR, 2017.}

\appbibitem{AppBrandenburg2024b}{Marie-Charlotte Brandenburg, Georg Loho, and Guido Montufar. The real tropical geometry of neural networks for binary classification. \emph{Transactions on Machine Learning Research}, 2024.}

\appbibitem{AppHaase2023}{Christian~Alexander Haase, Christoph Hertrich, and Georg Loho. Lower bounds on the depth of integral {ReLU} neural networks via lattice polytopes. In \emph{The Eleventh International Conference on Learning Representations}, 2023.}

\appbibitem{AppHertrich2025}{Christoph Hertrich and Georg Loho. Neural networks and (virtual) extended formulations, 2025.}

\appbibitem{AppArora2018}{Anirbit Mukherjee, Amitabh Basu, Raman Arora, and Poorya Mianjy. Understanding deep neural networks with rectified linear units. In \emph{International Conference on Learning Representations}, 2018.}

\appbibitem{AppSankaranarayanan2021CDINN}{Parameswaran Sankaranarayanan and Raghunathan Rengaswamy. Cdinn -- convex difference neural networks. \emph{Neurocomputing}, 495(C):153--168, 2022.}

\appbibitem{AppZhang2018}{Liwen Zhang, Gregory Naitzat, and Lek-Heng Lim. Tropical geometry of deep neural networks. In \emph{Proceedings of the 35th International Conference on Machine Learning}, pages 5824--5832. PMLR, 2018.}

\appbibitem{AppZiegler1995}{G{\"u}nter~M. Ziegler. \emph{Lectures on Polytopes}. Springer, New York, 1995.}

\end{document}

%% file: preamble.tex
\usepackage{times}
\usepackage{epsfig}
\usepackage{graphicx}
\usepackage{amsmath,amssymb,amsthm,mathtools}
\usepackage{array,booktabs,longtable}
\usepackage{subcaption}
\usepackage{mathrsfs}
\usepackage{bbold}
\usepackage{float}
\usepackage{tikz}
\usepackage{array,multirow}
\usepackage{makecell}
\usepackage{tcolorbox}
\usepackage{hhline}
\usepackage{placeins}
\usepackage{mathtools}

\newcounter{appbibcounter}
\newcommand{\appbibitem}[2]{%
  \refstepcounter{appbibcounter}%
  \par\noindent\hangindent=2em\hangafter=1%
  \hypertarget{appbib:#1}{[App\theappbibcounter]}\label{appbib:#1} #2%
  \par\vspace{0.5em}%
}
\newcommand{\citeapp}[1]{\hyperlink{appbib:#1}{[App\ref*{appbib:#1}]}}
\newcommand{\citepapp}[1]{\hyperlink{appbib:#1}{[App\ref*{appbib:#1}]}}
\newcommand{\citetapp}[1]{\hyperlink{appbib:#1}{[App\ref*{appbib:#1}]}}

\newtheorem{theorem}{Theorem}

\newtheorem{claim}{Claim}

\newtheorem{remark}{Remark}

\newcommand{\NN}{\mathbb{N}}
\newcommand{\RR}{\mathbb{R}}
\newcommand{\relu}{\text{ReLU}}

\newenvironment{proofref}[2]{%
  \begin{proof}[Proof of #1~\ref{#2}]%
}{%
  \end{proof}%
}

\newcommand{\TODO}[1]{\textbf{\color{red}[TODO: #1]}}
\renewcommand{\TODO}[1]{}

\usepackage{microtype}

\renewcommand{\paragraph}[1]{\vspace{.5em}\noindent\textbf{#1.}}

\usepackage{soul}
\setuldepth{foobar}

\renewcommand{\arraystretch}{1.0}

\usepackage{environ}
\NewEnviron{hideappendix}{%
  \setbox0=\vbox{\BODY}%
  \unvbox0
}

%% file: sections/introduction.tex
\section{Introduction}

Explaining the decisions of a trained neural network is among the deepest questions in machine learning.
Several methods have been proposed to make these decisions more interpretable, 
including visualization and computing several metrics. 
We present a technique to boost various of these approaches. 
The basic idea is to split a trained neural network in two (convex) monotone networks, allowing to benefit from the better interpretability of the latter type of networks. 
Explainability is crucial for debugging and improving models, detecting spurious correlations, and fostering trust in deployment. 
Methods of explainable artificial intelligence (XAI) have been used to localize relevant evidence, diagnose failure modes, and validate the reliability of explanations themselves \citep{ZeilerFergus2014,Montavon2017DeepTaylor,Ribeiro2016,Adebayo2018,Anders2020ClArC}. 

\begin{figure}[htbp]
    \centering
    \includegraphics[width=0.28\textwidth]{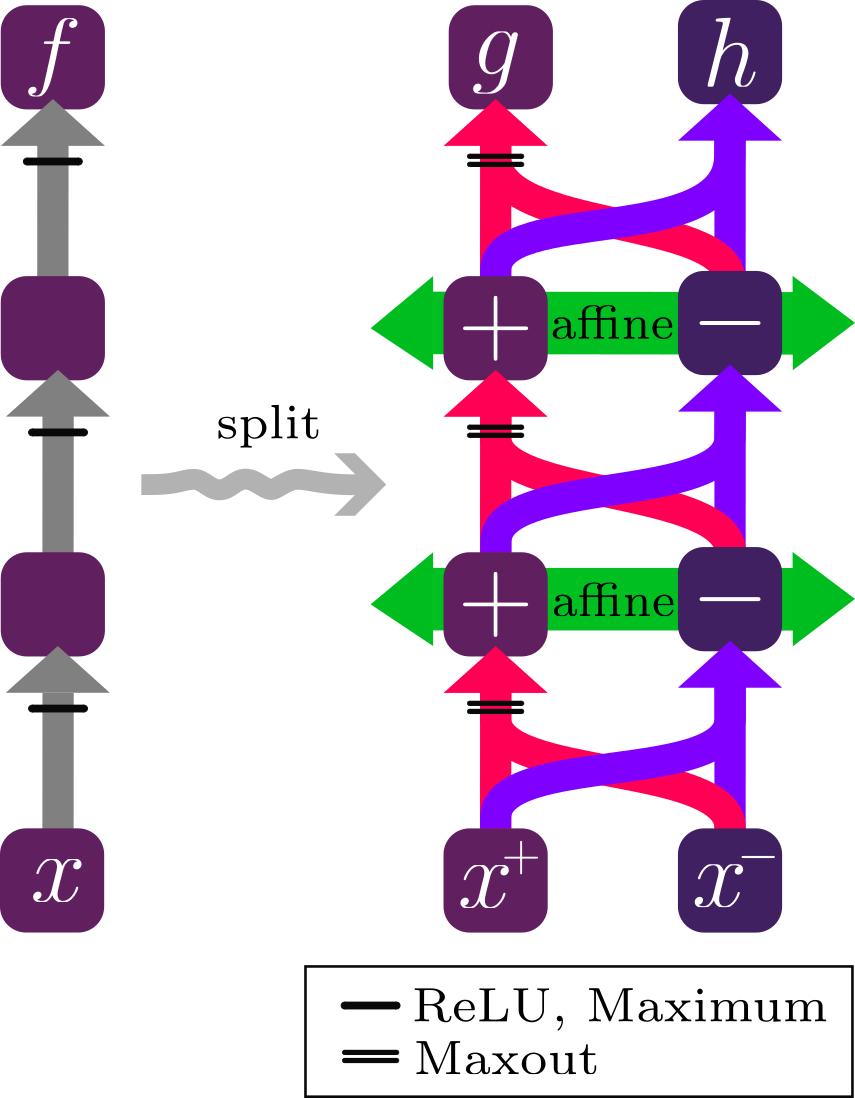}
    \caption{Illustration of the network splitting procedure. A ReLU network $f$ is decomposed into two monotone and convex streams $g$ and $h$, such that $f = g - h$.}
    \label{fig:network_split}
\end{figure}

\paragraph{Monotonicity as a tool for XAI}
Monotonicity has long been associated with interpretability, (adversarial) robustness, and fairness in machine learning \citep{Amos2017,Wang2020ShapeFairness,Liu2020CertifiedMonotone}. 
Networks that enforce non-negative weights
have been demonstrated to correctly express monotone dependences
and represent actual parts of the learned structure as no cancellation can arise~\citep{ayinde2018nonnegativity,lee1999nmf}.
However, while these models offer interpretability and fairness “by design”, they typically come at a cost in expressivity and cannot be easily applied to existing pretrained architectures.
In this work, we address this issue.
We use the idea of splitting any ReLU network into a pair of two \emph{split-streams}, structured sub-networks with non-negative weights, both representing monotone and convex functions $g$ and $h$ such that the originally represented function $f$ is their difference $f=g - h$ (see \cref{fig:network_split}).
The key idea arises from the identity
\begin{align}
    \relu(a-b) = \max\{a-b,0\} = \max\{a,b\} - b \enspace \label{eq:max_formulation}.
\end{align}
While this splitting was already present in earlier work~\citep{Siahkamari2020_DCRegression,Awasthi2023DCNN} as a special case of a Difference-of-Convex Decomposition (DC Decomposition), it has not been successfully applied in the context of explainability.
Our key contribution lies in the numerical stabilization of the forward and backward pass through this split representation by affine transformations of network activations and gradients.
This allows us to use it effectively as a tool for increasing explainability of an already trained model without retraining or altering predictions. 
Intuitively, the distinct positive and negative split-streams of the \emph{split-pair} preserve pre-activation information even when the original neuron output is zero. 

\paragraph{SplitCAM and SplitLRP}
Concretely, we develop a LayerCAM-inspired saliency method on this representation \citep{Jiang2021LayerCAM} and furthermore adapt Layer-wise Relevance Propagation to operate on the split-pair~\citep{Bach2015}.
Presenting state-of-the-art results for saliency maps with respect to VGG16 ~\citep{Simonyan2015} and ResNet18 \citep{Kaiming2016_resnet} models across selected metrics from Quantus~\citep{Hedstrom2023} across the desiderata \emph{faithfulness, robustness} and \emph{localization} we provide first evidence that the information flow through split networks yields more expressive explanations of model decisions
than those derived from the original architecture.
In some sense, the split-pair enforces a monotone, biologically plausible computational scheme:
non-negative activations and weights allow neurons only to increase evidence for a class rather than inhibit it, mirroring biological neural processing~\citep{lee1999nmf,Dayan2001_TheoreticalNeuroscience}.

\paragraph{Differences of networks}
One may wonder if these experimental results for the split models are a consequence of our specific splitting technique or really inherently in the difference of two monotone or convex networks.  
Therefore, while the previous section addressed the explainability of trained models, we also show that models directly trained as difference of convex and monotone models have remarkable properties in terms of explainability. 

We train DIC-models (difference-of-convex) and DM-models (difference-of-monotone). 
When the input to the $h$-stream is inverted, we observe a clear
separation of roles: the $g$-stream highlights present features, while the $h$-stream responds to
missing ones, yielding expressive and class-specific gradient explanations.
We demonstrate this behavior on MNIST, see
\cref{app:explaining_io_convex_nns}.
Our experiments serve as a proof of concept to demonstrate the improvements achieved by our Split framework when applied to pretrained ReLU networks.

\paragraph{Related work on convexity / monotonicity} 
DC-decompositions of neural networks have already appeared in various work. 
Most notably, the decomposition of a feedforward as well as convolutional neural network was presented in~\citep{Awasthi2023DCNN} so that DC-programming could be applied to train a network. On the other hand, using quadratic programming, basically the same decomposition was used in~\citep{Siahkamari2020_DCRegression} to efficiently fit a difference of convex functions to data. 
Additionally, \citep{Sankaranarayanan2021CDINN} introduce the network architecture based on a difference of two convex functions (CDiNN) and show that it behaves well for optimal control problems.  
More generally, there is a lot of work on finding and leveraging DC-decompositions in optimization~\citep{Melzer1986_PLasDC,Kripfganz1987_PWAasDC,Zalgaller2000_DCRepresentation,Griewank2020_PolyhedralDCDecomposition,Schlueter2021_ConvexDecompPAF,Koutschan2024_MinimalArity,Kazda2024_LP_DCPLA,Brandenburg2025_DecompositionPolyhedra}.
As a building block, it is important to mention the fundamental idea to use (partially) monotone networks, going back to~\citep{Sill1997_MonotonicNetworks}
and appearing in various contexts~\citep{Sivaraman2020_COMET,sivaprasad2021curious,Nguyen2023_MonoNet,Liu2020CertifiedMonotone,Daniels2010_MonotoneNN}.
While these are usually based on non-negative weights, the class of input convex neural networks (ICNN) is similar but more general~\citep{Amos2017}.

\paragraph{Related work on XAI} 
A large body of XAI research investigates pixel- or feature-level attributions in deep networks, including gradient saliency~\citep{Simonyan2014}, deconvolutional and guided-backprop visualizations~\citep{ZeilerFergus2014,Springenberg2015}, Layer-wise Relevance Propagation (LRP) and deep Taylor decomposition~\citep{Bach2015,Montavon2017DeepTaylor}, class-activation mapping (Grad-CAM)~\citep{Selvaraju2017}, axiomatic attributions such as Integrated Gradients and DeepLIFT~\citep{Sundararajan2017,Shrikumar2017}, noise-averaged gradients (SmoothGrad)~\citep{Smilkov2017}, and perturbation- or surrogate-based explanations (LIME, Meaningful Perturbations, SHAP, EVA, SharletX)~\citep{Ribeiro2016,FongVedaldi2017,LundbergLee2017, Kolek2023_ShearletX, Fel2023_EVA}.
Such tools have exposed spurious “Clever Hans” behaviors and guided mitigation strategies \citep{Anders2020ClArC,Kauffmann2020CleverHans,Kauffmann2018OneClassDTD}.

\paragraph{Contributions}
Our main contributions are as follows:
\begin{itemize}
    \item \textbf{A numerically stable DC decomposition of ReLU networks, including CNNs.}
    We propose an algorithm that transforms any pretrained ReLU network into two monotone and convex subnetworks without altering its outputs. This includes novel stabilization techniques for both the forward and backward pass.
    
    \item \textbf{New attribution methods — SplitCAM, SplitLRP, and SplitGrad.}
    We adapt LayerCAM and LRP to operate on the split networks and introduce a new gradient-based method, leveraging the improved signal flow in the DC Decomposition.
    
    \item \textbf{State-of-the-art explainability results.}
    Across VGG16 and ResNet18 on ImageNet-S \citep{Gao2022imagenets}, our methods outperform widely used baselines on the Quantus benchmark suite, including metrics from the desiderata faithfulness, localization and robustness. We complement our results with a comprehensive ablation study demonstrating the superiority of SplitCAM over classical LayerCAM as well as SplitGrad over the gradients of the original model, and rigorously analyze the contribution of our stabilization technique.
\end{itemize}

%% file: sections/preliminaries.tex
\section{Preliminaries}\label{sec:prelim}

We start with notation and basic concepts.

\subsection{Standard neural network notation}\label{subsec:relu_notation}

For simplicity, we start with
standard feedforward networks with $L$ hidden layers without bias.
Let $d^{(0)}, \ldots, d^{(L)}$ denote the layer widths, with input dimension $d^{(0)}$ and output dimension $d^{(L)}$, and
weight matrices $W^{(l)} \in \RR^{d^{(l)} \times d^{(l-1)}}$.
One can include bias by using an \emph{intercept input} of~$1$. 

Using an activation function $\mathtt{act}$, we denote by $a^{(0)} \coloneqq x \in \RR^{d^{(0)}}$ the input layer, by $z^{(l)} \coloneqq W^{(l)} a^{(l-1)}$ the pre-activation and by $a^{(l)} \coloneqq \mathtt{act}(z^{(l)})$ the activation.
Furthermore, we denote the function represented by the layers $l, \dots, L$ of the original network by $f^{(l)} : \ \RR ^ {d^{(l)}} \longrightarrow \RR$. 

\subsection{Convex and monotone neural networks}

We call a neural network with convex and monotone activation functions \emph{input-convex neural network} (ICNN) if all weights except possibly the first layer are non-negative~\citep{Amos2017}.
We call a neural network with monotone activation functions \emph{monotone} if all its weights are non-negative. 
It is well known that indeed ICNNs represent convex functions and monotone networks represent monotone functions; see \ref{app:fundamental}.
Note that both Maxout and ReLU are monotone and convex activation functions.

%% file: sections/method.tex
\section{Splitting ReLU networks}\label{sec:method}

To construct the subnetworks $g$ and $h$ with non-negative weights, we recall some of the techniques from~\cite{Zhang2018, Siahkamari2020_DCRegression,Awasthi2023DCNN}; additional proofs are deferred to \cref{app:fundamental}. 

\begin{figure}[t]
    \centering
    \includegraphics[width=.6\linewidth]{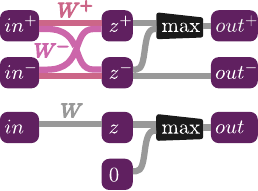}
    \caption{Visualization of the monotone and convex split-streams (top) in comparison to a standard ReLU layer (bottom). The negative weight matrix transitions positive input to negative preactivation and vice versa, whereas the positive part of the weight transitions from positive to positive and negative to negative.}
    \label{fig:splitstreams}
\end{figure}

\subsection{An algorithm for ReLU MLPs}\label{subsec:relu_mlp-split}

Each layer \(l \in \{1,\dots,L\}\) is replaced by a split-pair of two monotone, non-negative split-streams \(g\) and \(h\). We visualize this flow in \cref{fig:splitstreams}.
We split parameters into non-negative parts $W^{(l)} = W^{(l,+)} - W^{(l,-)}$ such that $W^{(l,+)}, \ W^{(l,-)} \geq 0$.
For more flexibility, we additionally split the input layer, replacing an input $x$ by two vectors $a^{(0,+)}$ and $a^{(0,-)}$ such that $x = a^{(0,+)} - a^{(0,-)}$. 

With this, for each \(l \in \{1,\dots,L\}\) we compute the split-stream pre-activations by
\begin{align}
      z^{(l,+)}
      &= W^{(l,+)} a^{(l-1,+)} + W^{(l,-)} a^{(l-1,-)}
      \label{eq:preact-plus} \\
      z^{(l,-)}
      &= W^{(l,-)} a^{(l-1,+)} + W^{(l,+)} a^{(l-1,-)}
      \label{eq:preact-minus}
\end{align}
By replacing the ReLU activation of the original model with a Maxout unit in the $g$-stream
one obtains the activations $a^{(l,+)} = \max \left\{ z^{(l,+)}, z^{(l,-)} \right\}$ and $a^{(l,-)} = z^{(l,-)}$.
Analogously to the notation $f^{(l)}$, we denote the split-streams $g$ and $h$ restricted to the layers $l,\dots,L$ by $g^{(l)} : \ \RR ^ {d^{(l)}} \times \RR ^ {d^{(l)}} \longrightarrow \RR$ and $h^{(l)} : \ \RR ^ {d^{(l)}} \times \RR ^ {d^{(l)}} \longrightarrow \RR$.

We prove correctness of the split in \cref{app:fundamental}.
Indeed, the difference between the outputs of the two split-streams equals the output of the original network evaluated at the difference of their inputs.

\begin{theorem}[Correctness] \label{thm:split_correct}
    For any $x^+, x^- \in \RR^{d^{(l)}}$
    \begin{align*}
        f^{(l)} \bigl( x^+ - x^- \bigr)
            = \bigl( g^{(l)} - h^{(l)} \bigr)
                \begin{bmatrix}
                    x^+ \\
                    x^-
                \end{bmatrix},
    \end{align*}
\end{theorem}

Here, the crucial observation for the next step is that we have flexibility in representing the input $x$ as a difference $x = x^+ - x^-$.

\subsection{Numerically stabilizing the forward pass} \label{sec:forward}

As our split-pair lacks cancellation by subtraction, network activations tend to explode in deep neural networks. 
To mitigate this, we propose shifting the activations into numerically stable regimes.

By \cref{thm:split_correct},
adding the same constant shift to both \(a^{(l,+)}\) and \(a^{(l,-)}\) preserves the difference of $g^{(l)}$ and $h^{(l)}$.
We show in \cref{app:fundamental} that adding a shift in each layer does not alter the gradients of \(g\) and \(h\). 
Indeed, the gradient flow is determined only by the indices, where the maximum between the split-streams preactivations are taken, and not by the preactivations actual value.
    
\paragraph{Implementation details for forward pass stabilization}
We keep activations in check by `re-centering' the pairs of
activations in the two split-streams whenever their magnitudes explode using different strategies based on thresholds, scaling, or shifting.
All intermediate computations use \texttt{torch.float64} for high numerical precision matrix multiplication.
Furthermore, we optionally cache the activations of the original model to slightly correct the activations of the split model on the fly, ensuring the \textbf{forward invariant:} $a^{(l,+)} - a^{(l,-)} = a^{(l)}$.

\subsection{Numerically stabilizing the backward pass} \label{sec:backward}

As the computation of the gradients of $g$ and $h$ involves only non-negative matrices, these quantities turn out to be very large for deep networks. 
Therefore, we subtract non-constant offsets in their iterative computation. 
While this does not result in the same numerical information as the gradient, it still captures the pattern of the activation function and preserves the relation of gradients for different inputs. 

So let $\delta^{(l,+,g)} \coloneqq \frac{\partial g^{(l)}}{\partial a^{(l,+)}}$ be the derivative of the `head' $g^{(l)}$ by the `positive part' $a^{(l,+)}$ of the split-stream activations of layer $l$, and let $\delta^{(l,-,g)}$, $\delta^{(l,+,h)}$ and $\delta^{(l,-,h)}$ be defined analogously. 
We call them \emph{local sensitivities}. 
We compute a \emph{modified gradient} by subtracting
an $\alpha^{(l)}$-multiple of the product of the absolute values of all subsequent weight matrices and then iterating this layer by layer. As we show mathematically in \cref{app:closed_formulas}, this reduces the absolute values of the local sensitivities.
We usually choose all the factors $\alpha^{(l)}$ to be the same value $\alpha$.
For $\alpha = 0.5$ the local sensitivities are just multiples of the original gradients, while for $\alpha = 0$ they resemble the actual split-pair gradients.
Hence, the most interesting regime is $0 < \alpha < 0.5$.

\paragraph{Implementation details for backward pass stabilization}
When backpropagating through ReLU and Maxpooling components, instead of using the maximum patterns of the split activations, \ie the indices, where the maximum of the split-streams preactivations are taken, we utilize the positivity patterns of the cached original network activations.
Furthermore, we cache the original network gradients and apply small corrections to ensure the \textbf{backward invariant}
    \begin{equation}
        (\delta^{(l,+,g)} - \delta^{(l,-,g)}) -
        (\delta^{(l,+,h)} - \delta^{(l,-,h)})
        = \frac{\partial f^{(l)}}{\partial a^{(l)}}.
        \label{eq:backward_invariant}
    \end{equation}

\subsection{Generalization to other network components} \label{subsec:other_components}

We represent every component as a \emph{chunk module} with explicit positive/negative inputs and outputs, and we keep linear and convolutional layers and ReLU as separate chunks in order to support multiple layers without intermediate activation.

A forward/backward cache accumulates inflowing activations and gradients per chunk, and a parallel cache mirrors the original model. The latter serves both analysis (layer-wise activation/gradient error) and optional on-the-fly correction (forward and backward) and reuses the original model’s positivity patterns in place of split-pair maxima in the backward pass.
Our converter maps arbitrary PyTorch graphs (for supported layers/functions) to the split form.

We give an overview of the adaptations to split further network components into non-negative parts:
\textbf{Convolutional operations} are handled analogously to linear layers by splitting each kernel into positive and negative parts.
\textbf{Bias terms} are decomposed as \(b=b^+-b^-\) and added separately to the two streams. 
\textbf{Additive operations}, including residual connections, simply add the corresponding split activations (and distribute gradients independently).
\textbf{Average Pooling} is a positive linear operator and is therefore applied to each stream without modification.
\textbf{Batch normalization} in evaluation mode reduces to an affine map and can be treated as a special convolution; handling training-mode statistics would require the multiplicative splitting described in \cref{app:splitting_multiplicative_maxout_networks}. We also support merging Batch normalization layers with adjacent linear or convolutional Layers.
\textbf{Dropout} in evaluation mode acts as identity, while in training mode we reconstruct the original dropout mask and apply it consistently to both activations and gradients of the split-streams.
\textbf{Alternative nonlinearities} such as ELU \cite{Clevert2016_elu} or Swish \cite{Ramachandran2017_swish} can be handled by replacing them with small ReLU subnetworks before applying the split procedure. 
For \textbf{Max pooling}, we use the convexity-preserving identity $\max_{j\in[q]}(a_j^+ - a_j^-) = \max_{j\in[q]}\Bigl\{a_j^+ + \sum_{i\neq j} a_i^- \Bigr\} - \sum_{i} a_i^-$.

Optionally, we replace this formulation with a ``winner-takes-it-all'' (\emph{wta}) rule that forwards the locations of both split-streams, where their difference takes the maximum. This however comes at the cost of monotonicity and convexity of the split-streams.

%% file: sections/DICs.tex
\subsection{Differences of networks analysis} \label{subsec:dc_network_training_methods}

\input{sections/counterfactual_features}

We train two base architectures on MNIST, \texttt{cnn\_maxout} and \texttt{mlp\_maxout}, and their difference-of-two variants.
The \textbf{DIC-model} (directly trained as the difference of two input-convex neural networks) denotes $g-h$ with both streams input-convex, while the \textbf{DM-model} (directly trained as the difference of two monotone networks) denotes $g-h$ with both streams monotone.
For each base model we also evaluate the \emph{inverted-input} variant (DIC\-IN / DM\-IN), where the $h$-stream processes $1-x$ so it can better focus on missing features.
Implementation-wise, positivity is enforced by parameterizing all constrained weights and affine terms via a nonnegativity transform, and batch norm parameters are constrained analogously.

On \texttt{mlp\_maxout}, DIC-models and DM-models slightly improve test accuracy over their single-stream counterparts.
The complete evaluation and more training details can be found in \cref{app:explaining_io_convex_nns}.
\cref{fig:mnist_split_ic_saliency_main} demonstrates the seperation of the two DIC models to focus on present and missing stroke features respectively.

%% file: sections/counterfactual_features.tex
\begin{figure}[t]
\centering

\begin{minipage}[c]{0.16\linewidth}
  \centering
  \includegraphics[width=\linewidth]{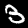}\\[-2pt]
  {\scriptsize Original}
\end{minipage}\hspace{0.01\linewidth}%
\begin{minipage}[c]{0.81\linewidth}
  \centering
  \begin{minipage}[b]{0.19\linewidth}\centering
    \includegraphics[width=\linewidth]{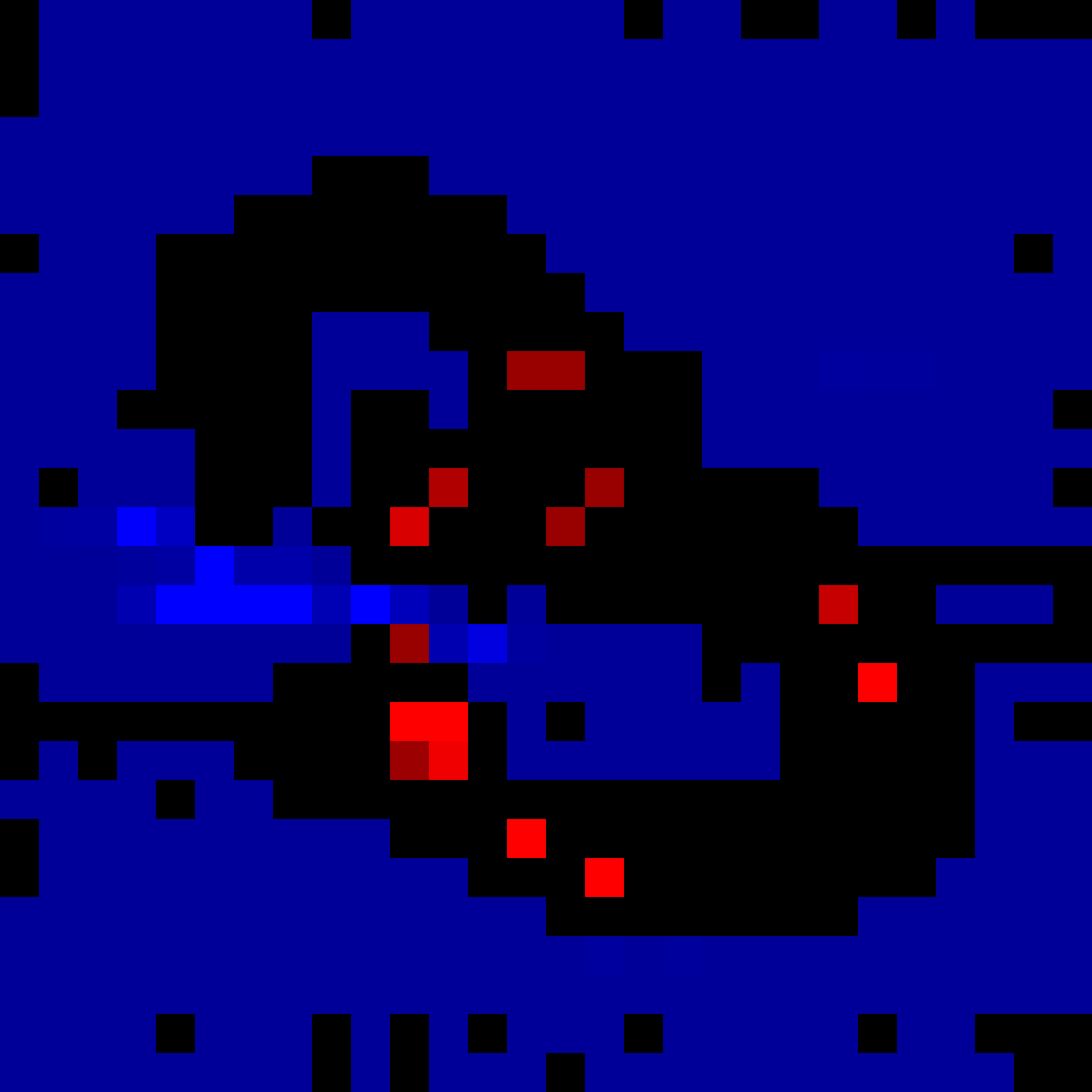}\\[-2pt]{\scriptsize Logit 0}
  \end{minipage}%
  \begin{minipage}[b]{0.19\linewidth}\centering
    \includegraphics[width=\linewidth]{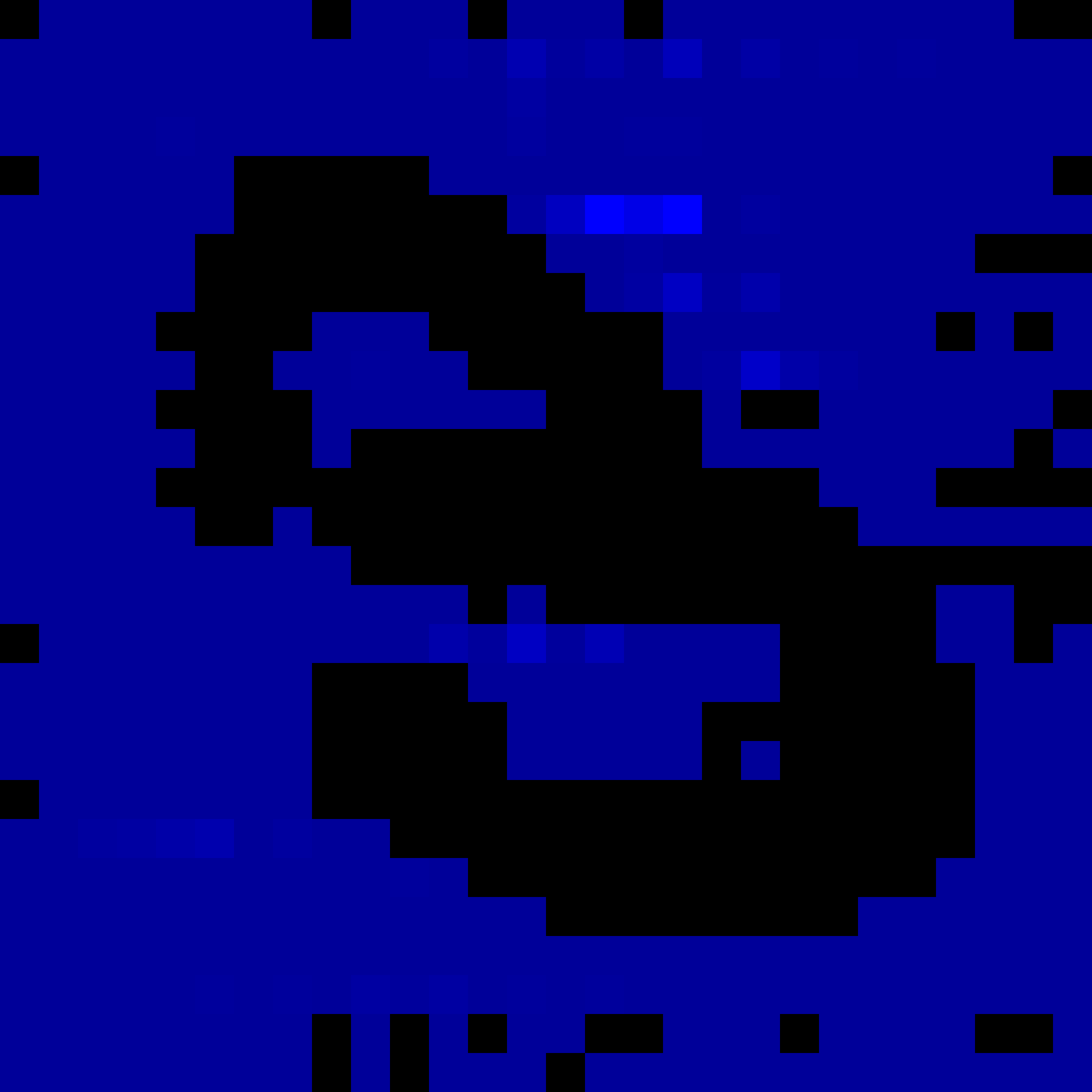}\\[-2pt]{\scriptsize Logit 1}
  \end{minipage}%
  \begin{minipage}[b]{0.19\linewidth}\centering
    \includegraphics[width=\linewidth]{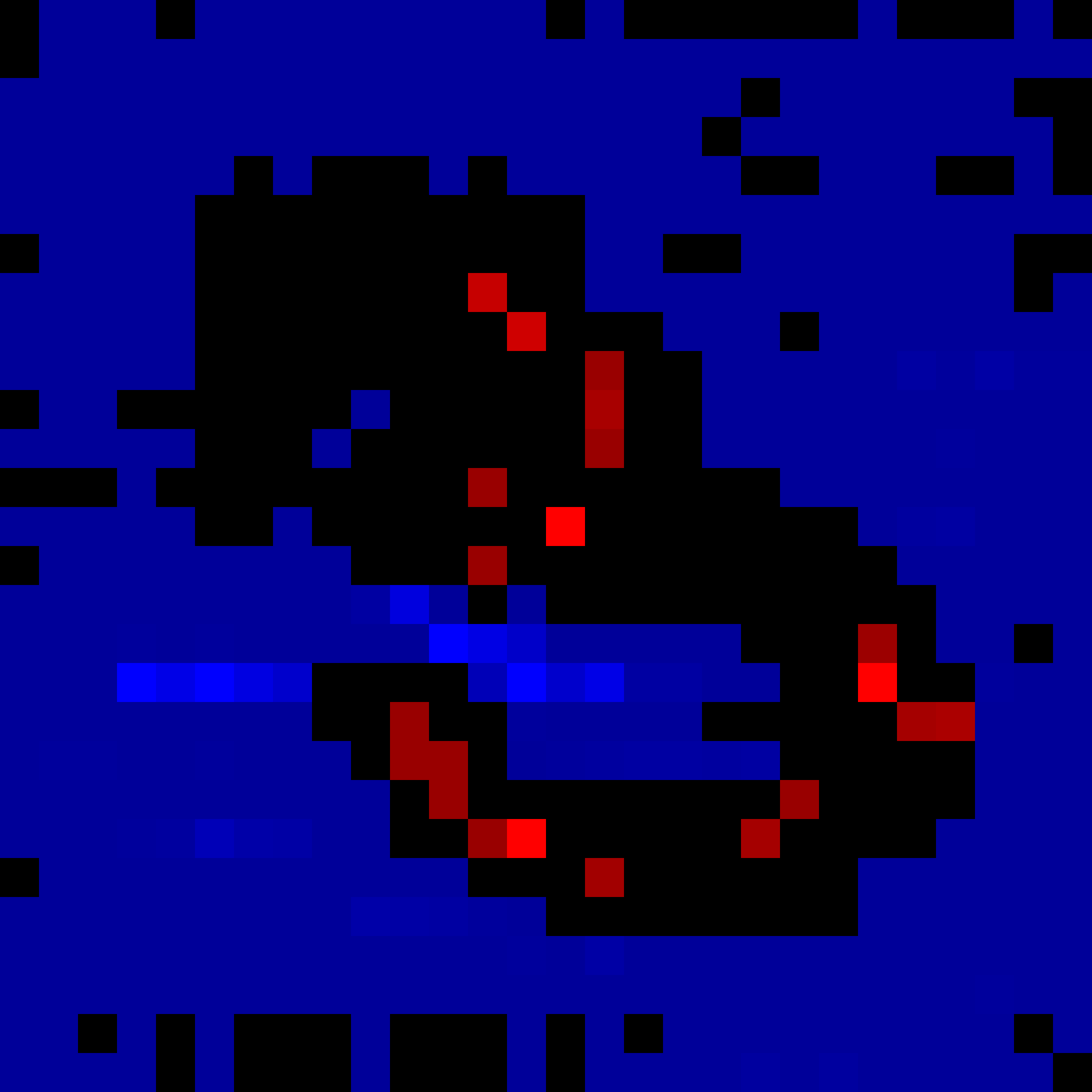}\\[-2pt]{\scriptsize Logit 2}
  \end{minipage}%
  \begin{minipage}[b]{0.19\linewidth}\centering
    \includegraphics[width=\linewidth]{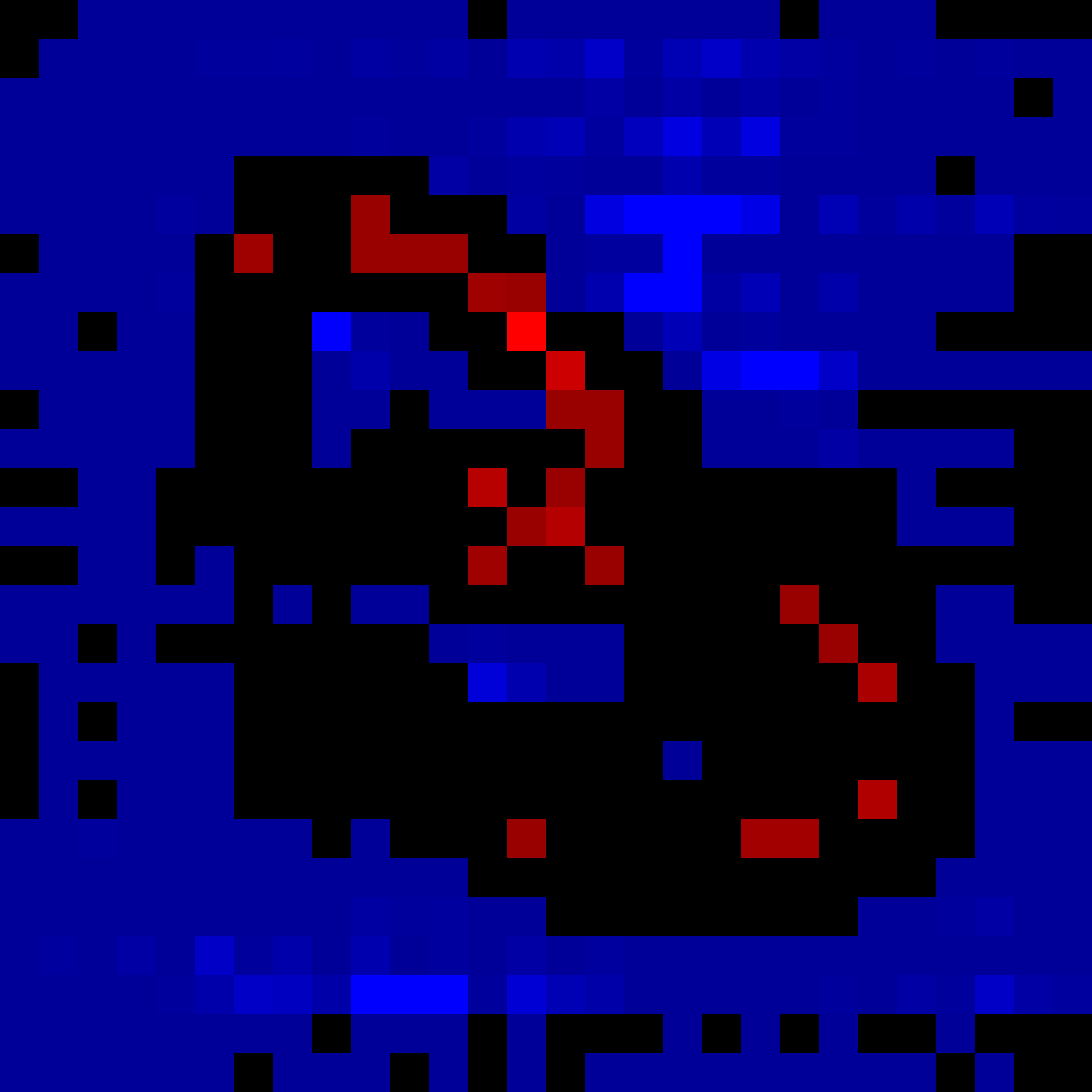}\\[-2pt]{\scriptsize Logit 3}
  \end{minipage}%
  \begin{minipage}[b]{0.19\linewidth}\centering
    \includegraphics[width=\linewidth]{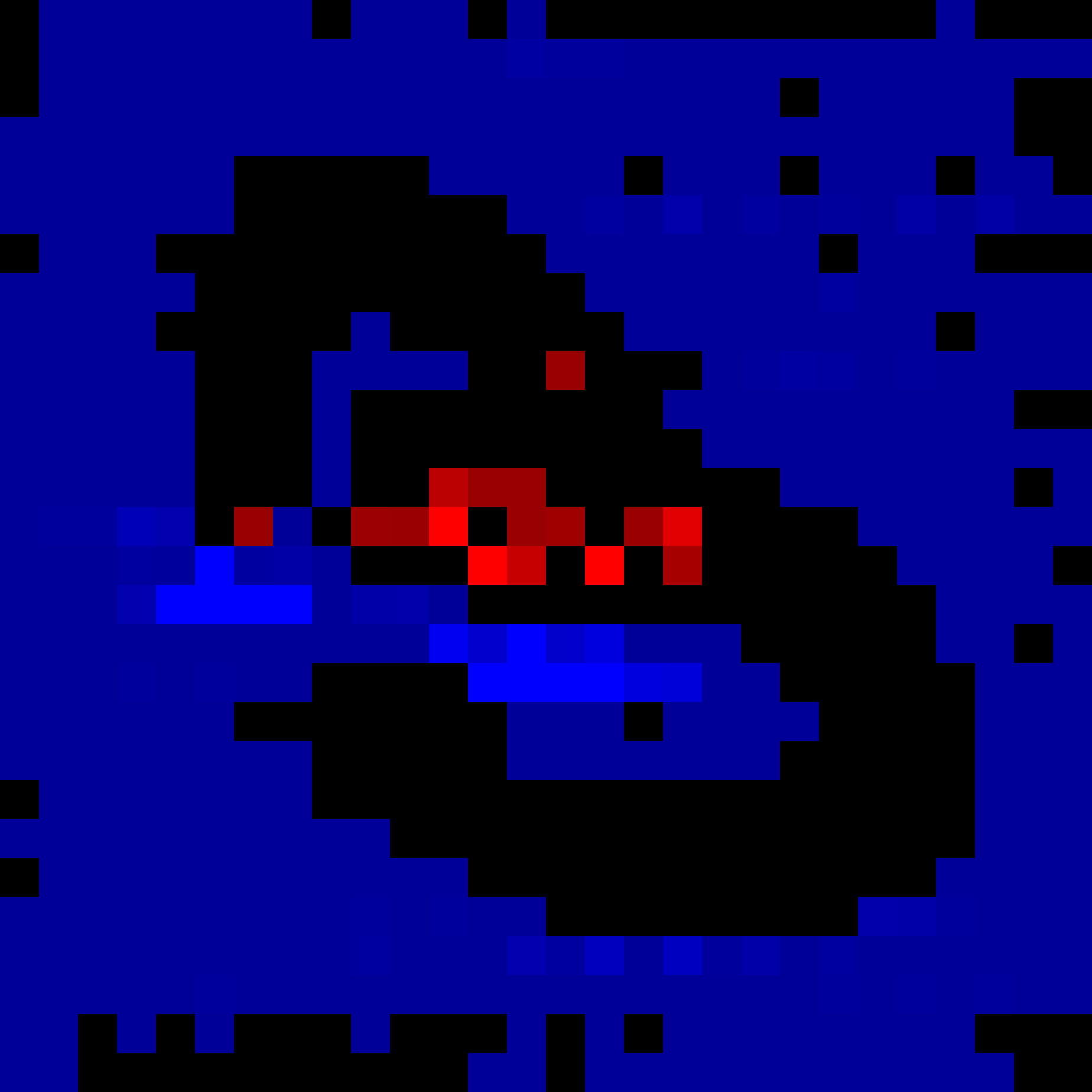}\\[-2pt]{\scriptsize Logit 4}
  \end{minipage}

  \vspace{0.12cm}

  \begin{minipage}[b]{0.19\linewidth}\centering
    \includegraphics[width=\linewidth]{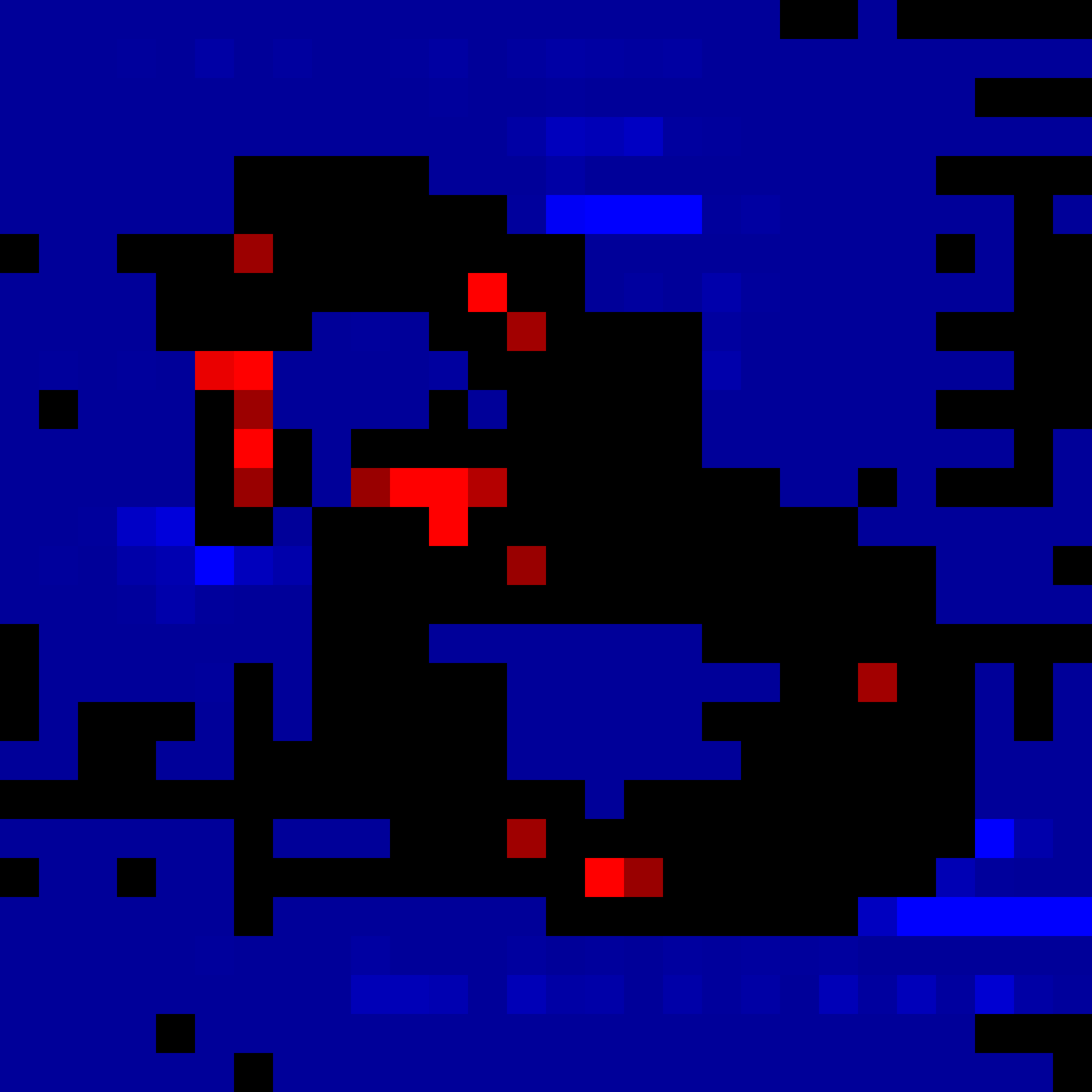}\\[-2pt]{\scriptsize Logit 5}
  \end{minipage}%
  \begin{minipage}[b]{0.19\linewidth}\centering
    \includegraphics[width=\linewidth]{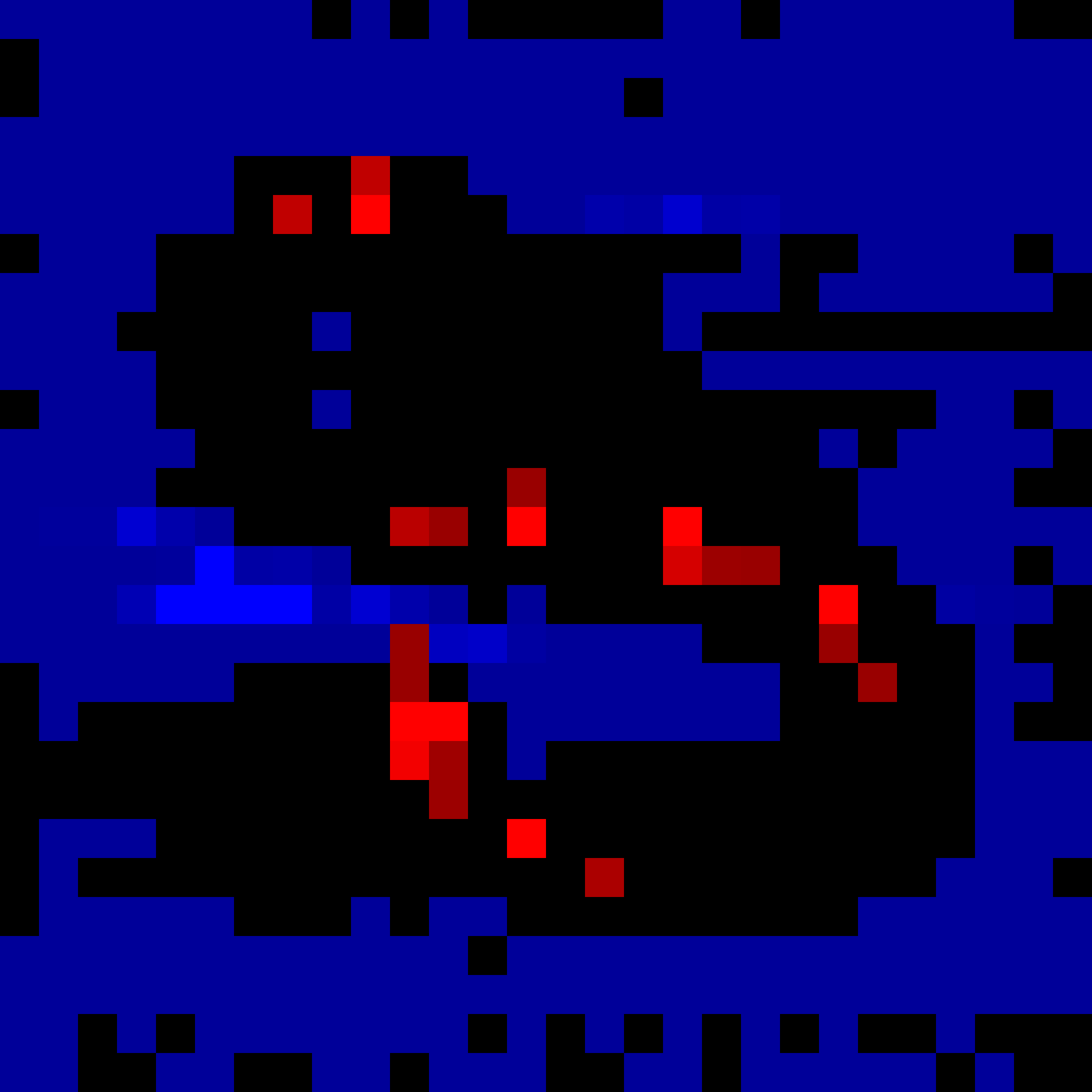}\\[-2pt]{\scriptsize Logit 6}
  \end{minipage}%
  \begin{minipage}[b]{0.19\linewidth}\centering
    \includegraphics[width=\linewidth]{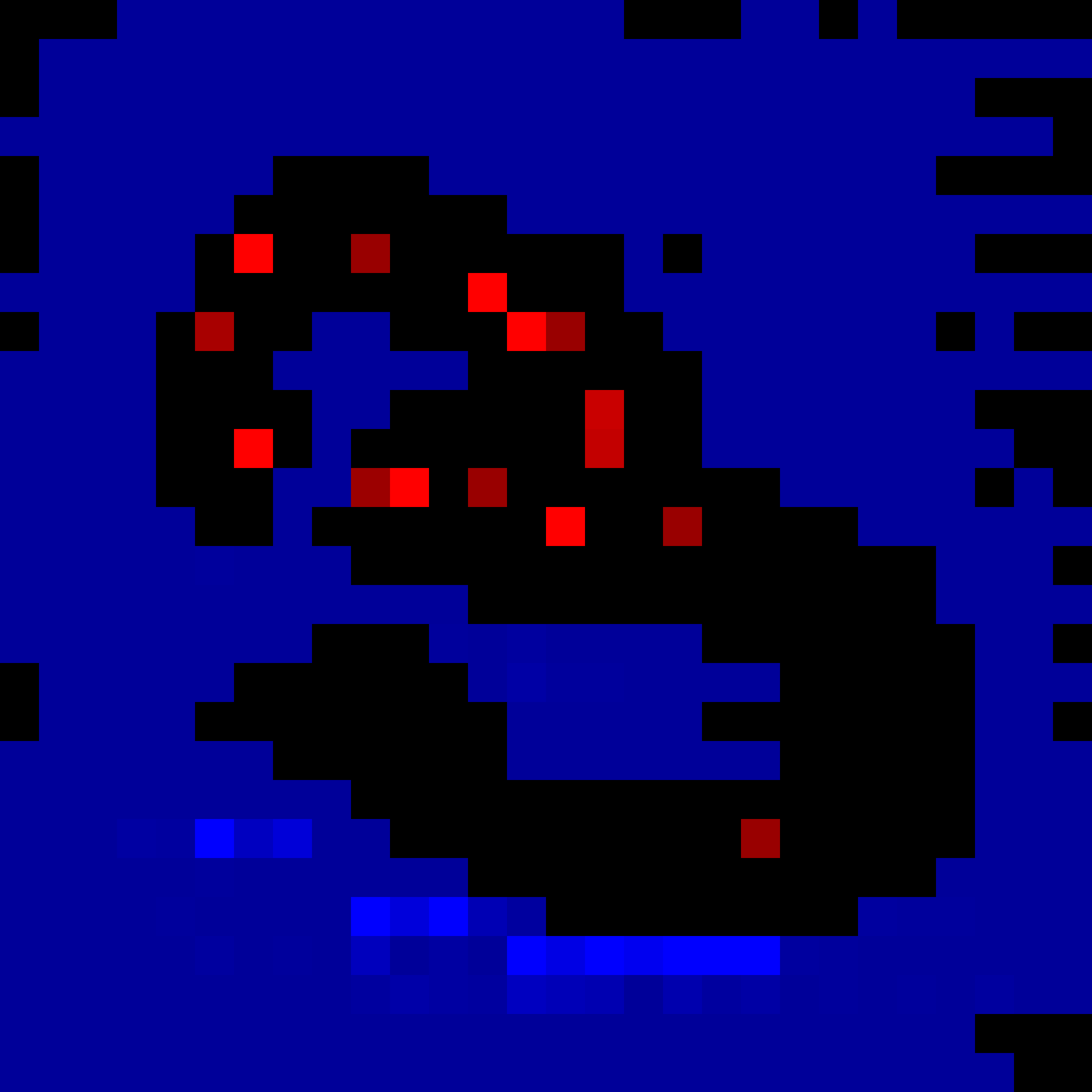}\\[-2pt]{\scriptsize Logit 7}
  \end{minipage}%
  \begin{minipage}[b]{0.19\linewidth}\centering
    \includegraphics[width=\linewidth]{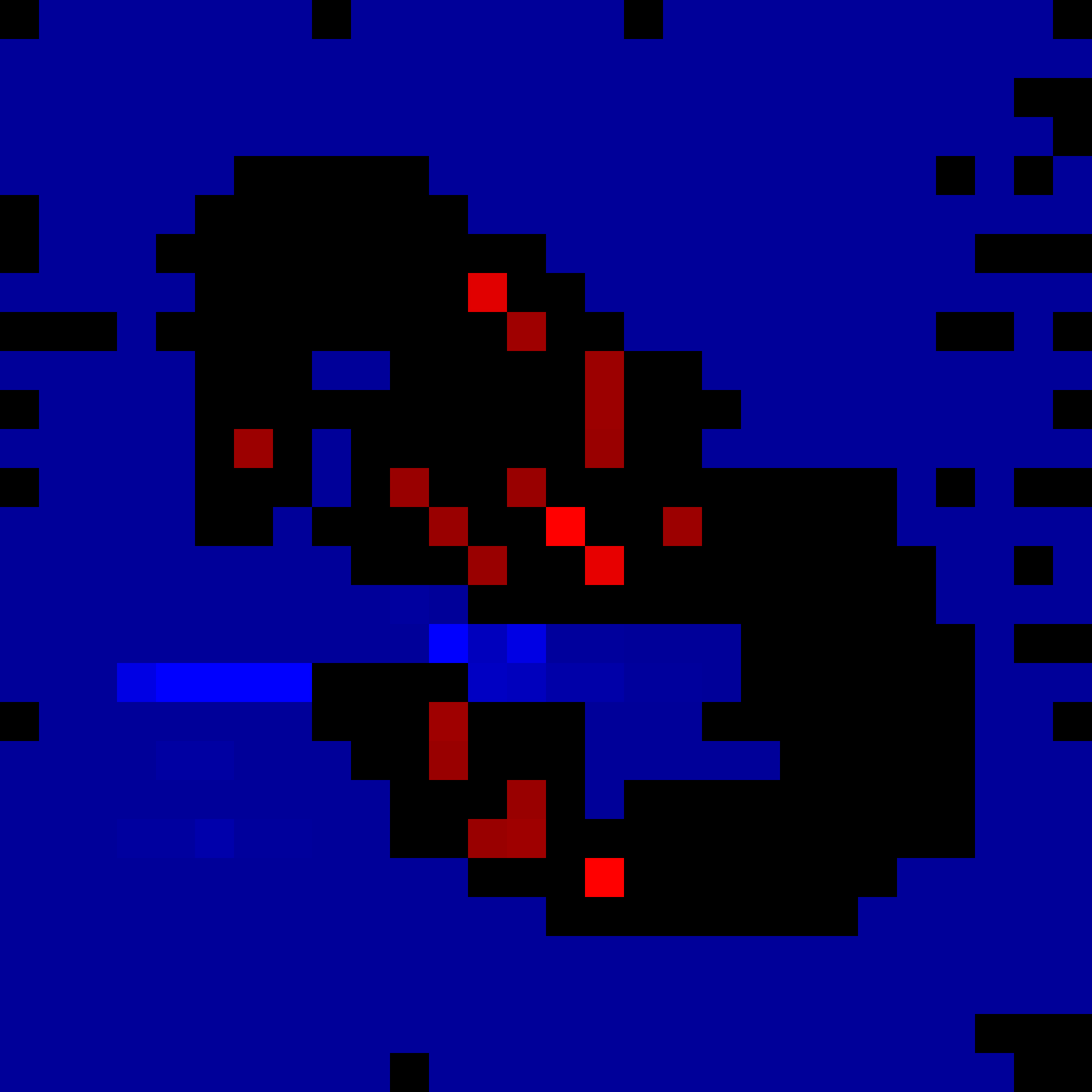}\\[-2pt]{\scriptsize Logit 8}
  \end{minipage}%
  \begin{minipage}[b]{0.19\linewidth}\centering
    \includegraphics[width=\linewidth]{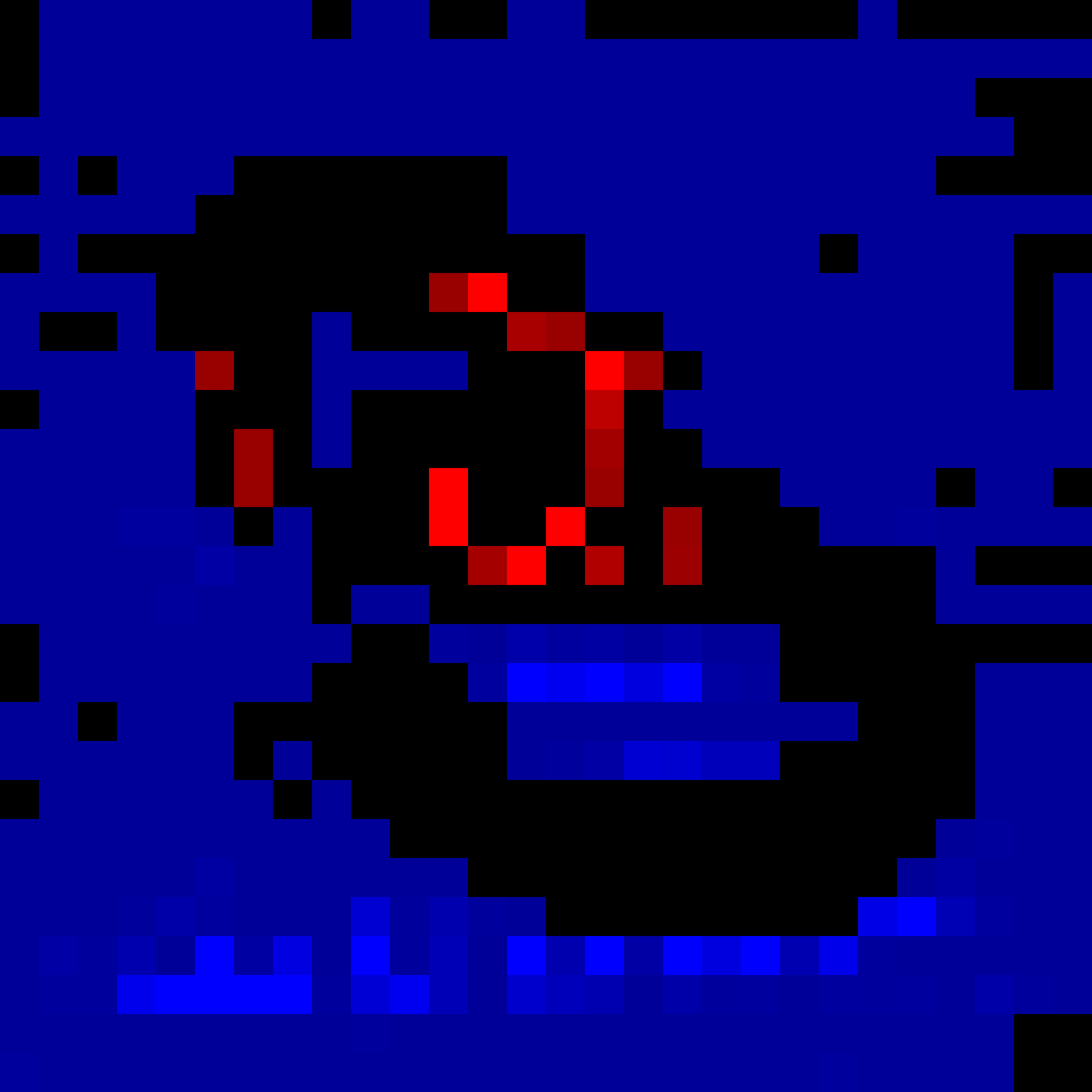}\\[-2pt]{\scriptsize Logit 9}
  \end{minipage}
\end{minipage}

\caption{Combined visualization of the inverted-input DIC gradients. The $g$-model gradient (red) focuses on present image parts, while the $h$-model gradient (blue) focuses on counterfactual features. On the left: the original digit.}
\label{fig:mnist_split_ic_saliency_main}
\end{figure}

%% file: sections/table_main_generated_vgg.tex
\begin{table*}[htb!]
    \centering
    \small
    \caption{Comprehensive XAI Evaluation Results (VGG16). The \emph{wta} abbreviation indicates the use of the ``winner-takes-it-all'' Maxpooling implementation. The \emph{sc} abbreviation emphasizes the scaling procedure in the forward pass. We highlight the best values for each metric in bold and red. The bracket notation in the layer column follows the upper indices of the methods as we introduced them in \cref{sec:method} and \cref{sec:methods_xai}.}
    \label{tab:eval_vgg_main}
    \begin{tabular}{p{4cm}|cc|cccccc}
        \toprule
        \textbf{Method} & \textbf{Layer} & \textbf{Abs} & \textbf{Select.} & \textbf{Attr. Loc.} & \textbf{Point.} & \textbf{Pixel Fl. (AUC@5)} & \textbf{Pixel Fl. (AUC@20)} & \textbf{Max Sens.} \\
        & & & \textbf{$\downarrow$} & \textbf{$\uparrow$} & \textbf{$\uparrow$} & \textbf{$\uparrow$} & \textbf{$\uparrow$} & \textbf{$\downarrow$} \\
        
        \midrule
        \multicolumn{9}{c}{\textit{Split Methods}} \\
        \midrule
        SplitCAM (sc, $\alpha = 0.4$) & (12, g) & n & 4.720 & \fboxsep=0.5pt\colorbox{red!20}{\textbf{0.651}} & 0.912 & 0.213 & 1.794 & 0.389 \\
        SplitCAM (sc, $\alpha = 0.4$) & (14, g) & n & \fboxsep=0.5pt\colorbox{red!20}{\textbf{2.727}} & 0.563 & 0.836 & 0.300 & 2.404 & 1.012 \\
        SplitCAM (sc, $\alpha = 0.4$, wta) & (26, +, g) & y & 4.711 & 0.635 & \fboxsep=0.5pt\colorbox{red!20}{\textbf{0.938}} & 0.165 & 1.590 & 0.456 \\
        SplitGrad ($\alpha = 0.4$, wta) & (12, +, g) & n & 4.735 & 0.635 & 0.883 & 0.186 & 1.756 & 0.651 \\
        SplitGrad ($\alpha = 0.4$, wta) & (14, +, g) & y & 3.287 & 0.540 & 0.818 & 0.249 & 2.186 & 0.857 \\
        SplitLRP (sc, wta) & (26, pos) & y & 5.168 & 0.588 & 0.871 & 0.151 & 1.473 & \fboxsep=0.5pt\colorbox{red!20}{\textbf{0.282}} \\
        \midrule
        \multicolumn{9}{c}{\textit{Baselines (With Layer Optimization)}} \\
        \midrule
        Guided Backprop & 2 & n & 3.276 & 0.632 & 0.883 & 0.281 & 2.302 & 0.571 \\
        Guided Backprop & 5 & n & 3.448 & 0.614 & 0.887 & 0.240 & 2.091 & 0.375 \\
        LayerCAM & 14 & - & 2.792 & 0.576 & 0.798 & \fboxsep=0.5pt\colorbox{red!20}{\textbf{0.317}} & \fboxsep=0.5pt\colorbox{red!20}{\textbf{2.474}} & 2.656 \\
        LayerCAM & 2 & - & 3.128 & 0.580 & 0.855 & 0.300 & 2.191 & 5.508 \\
        LRP $\gamma=0.25$ & 0 & n & 3.189 & 0.615 & 0.680 & 0.218 & 1.918 & 0.519 \\
        LRP $\gamma=0.25$ & 14 & n & 3.593 & 0.577 & 0.611 & 0.279 & 2.255 & 0.491 \\
        \midrule
        \multicolumn{9}{c}{\textit{Baselines}} \\
        \midrule
        DeepLift & 0 & y & 4.447 & 0.551 & 0.726 & 0.281 & 1.945 & 1.053 \\
        Feature Ablation & 0 & y & 6.029 & 0.479 & 0.624 & 0.123 & 1.250 & 0.859 \\
        Gradient SHAP & 0 & y & 4.808 & 0.552 & 0.783 & 0.261 & 1.862 & 1.142 \\
        GradCAM++ & 30 & n & 5.955 & 0.564 & 0.827 & 0.130 & 1.245 & 0.401 \\
        Integrated Gradients & 0 & y & 4.838 & 0.554 & 0.797 & 0.262 & 1.851 & 1.069 \\
        Occlusion & 0 & y & 5.214 & 0.499 & 0.716 & 0.130 & 1.217 & 0.699 \\
        Random Baseline & 0 & y & 3.570 & 0.425 & 0.458 & 0.203 & 1.585 & -- \\
        \bottomrule
    \end{tabular}

\end{table*}

%% file: sections/methods_xai.tex
\section{XAI Methods}\label{sec:methods_xai}

In this section, we leverage the neural network decomposition framework introduced in the previous sections to boost two common XAI attribution methods: LayerCAM and LRP.

\subsection{SplitCAM}\label{subsec:split_cam}

Class Activation Mapping (CAM) \citep{Zhou2016} and its variants (Grad-CAM \citep{Selvaraju2017}, Grad-CAM++ \citep{Chattopadhay2018}, LayerCAM \citep{Jiang2021LayerCAM}) generate visual explanations by combining layer activation maps with class specific weights or layer gradients.

For layer $l$, LayerCAM computes the saliency map as
\begin{equation} \label{eq:layercam}
    \text{LayerCAM}^{(l)} = \text{ReLU}\left(\sum_{c} \frac{\partial y}{\partial a^{(l)}_c} \odot a^{(l)}_c\right)
\end{equation}
where $a^{(l)}_c$ is the activation map of channel $c$ at layer $l$, $y$ is the class score, and $\odot$ denotes element-wise multiplication.

\emph{SplitCAM} extends the LayerCAM framework to our split-pair architecture.
For a given layer~$l$ in the decomposed model we combine the two split-stream activations with its gradients, namely $\delta_{\text{shift}}^{(l,+,g)}$ (one of the shifted local sensitivities) and $\delta_{\text{shift}}^{(l,g)}$ (a combination of such) defined in \cref{sec:backward}. Different from the original LayerCAM formula \cref{eq:layercam}, SplitCAM omits the positive filtering by the ReLU function.

\begin{align}
    \text{SplitCAM}^{(l, +, g)} &\coloneqq \sum_{c} \delta_{\text{shift}}^{(l,+,g)} \odot a^{(l,+)}_c, \\
    \text{SplitCAM}^{(l, g)} &\coloneqq \sum_{c} \delta_{\text{shift}}^{(l,g)} \odot a^{(l)}_c,
\end{align}
where $a^{(l,+)}_c$ denotes the positive activation map of channel~$c$.
$\text{SplitCAM}^{(l, -, g)}$ and the variants for $h$ are defined analogously.

In the forward pass, we scale the entire activation tensor by a fixed \emph{minimize factor} $\theta$ whenever the absolute value of any activation entry exceeds a predefined \emph{minimize threshold} $\Theta$.
In all our experiments we use $\theta = 0.1$ and $\Theta = 10$.
We apply corrective shifts to the split-pair activations such that their difference matches the original network's activation.

In the backward pass we apply the shifting procedure described in \cref{sec:backward} and slightly correct the four sensitivity maps in each layer to ensure the backward invariant \cref{eq:backward_invariant}.

\paragraph{SplitGrad} As a simple attribution method we also propose SplitGrad, which resembles the channel mean of the shifted split-pair gradients in the previous setup.

\subsection{SplitLRP}\label{subsec:split_lrp}

Layer-wise Relevance Propagation \citep{Bach2015} redistributes the output score iteratively from the
output layer to the input layer depending on the weights and activations. There are several extensions to the classical LRP rule. 
Here we present the $\gamma$-rule, introduced by Montavon et al.\ in \citep{Montavon_2022_gammaLRP}, that tries to emphasize positive influences to the network's prediction by increasing the influence of positive weights.

\paragraph{LRP-\(\gamma\) rule (linear / convolution layers)}
For an affine mapping \(z^{(l+1)}_i=\sum_j W^{(l+1)}_{ij} a^{(l)}_j + b^{(l+1)}_i\) we define the $\gamma$-adapted positive forward contribution
\begin{align}
    \tilde{z}^{(l+1)}_{ij} \coloneqq a^{(l)}_j \left( W^{(l+1)}_{ij} + \gamma \max\left\{W^{(l+1)}_{ij},0\right\} \right),
\end{align}
with \(\gamma \ge 0\) accentuating positive contributions.
Given relevances \(R^{(l+1)}_i\) at the outputs, the \(\gamma\)-rule redistributes them to inputs \(j\) as
\begin{equation}
\label{eq:lrp-gamma}
R^{(l)}_j = \sum_i \frac{\tilde{z}^{(l+1)}_{ij}}{\sum_{j'} \tilde{z}^{(l+1)}_{ij'} + \varepsilon\,\mathrm{sign}\!\left(\sum_{j'} \tilde{z}^{(l+1)}_{ij'}\right)}\, R^{(l+1)}_i,
\end{equation}
with a small \(\varepsilon>0\) for numerical stability.

\paragraph{SplitLRP implementation details}
In the DC Decomposition, each layer consists of a positive and a negative split-stream, which we treat as two sets of neurons within a unified model. Accordingly, LRP on the split-pair produces two relevance tensors in each layer, $R^{(l,+)}$ and $R^{(l,-)}$.
Since the LRP rule ignores any activation functions in the backward pass no special care is needed for the Maxout activation functions of the split-pair.
We propagate relevance from the class-specific positive output $a^{(L,+)}$. Since all weights in the decomposed model are non-negative, we use the standard LRP-$\gamma$ rule with $\gamma=0$ and $\varepsilon = 10^{-6}$, essentially forming the so-called $\epsilon$-LRP rule. As common practice, layer-normalization modules are merged into adjacent linear or convolutional layers, and relevance is split evenly across residual addition branches.
We scale and correct the activations in the forward pass just as we do in SplitCAM.
Because LRP normalizes relevance at each layer, see \cref{eq:lrp-gamma}, the total relevance in $R^{(l,+)}$ and $R^{(l,-)}$ always sums to $a^{(L,+)}$, so no additional stabilization is required during the backward pass.

We additionally report \emph{combined} LRP maps, defined as
\begin{align}
    R^{(l,\text{comb})} \coloneqq R^{(l,+)} - R^{(l,-)}.
\end{align}
This quantity aggregates the positive and negative streams into a single saliency map motivated by the following intuition: 
$R^{(l,+)}$ highlights evidence supporting the prediction, while $R^{(l,-)}$ captures evidence against it.

\paragraph{Maxpooling Nonlinearities}
We propagate relevance backwards through Maxpooling layers using the ``winner-takes-all'' scheme, assigning all relevance to the location that achieved the maximum activation.

%% file: sections/experiments.tex
\section{Experiments}\label{sec:experiments}

\subsection{Evaluation Setup}

\paragraph{Implementation}
All models and experiments are implemented using PyTorch~\citep{Paszke2019PyTorch}.

\paragraph{Model and Dataset}
We employ a pretrained VGG16 model on the ImageNet \cite{Deng2009} classification task.
Because localization metrics require ground-truth object segmentation masks, we evaluate on ImageNet-S-50~\citep{Gao2022imagenets}, which provides pixel-level semantic segmentation annotations for 50 ImageNet classes, excluding unsegmentable classes (\eg \emph{bookshop}).
The dataset is partitioned into a validation set (50 images, one randomly selected for each class) and a test set (the remaining 566 images) for final evaluation.

\paragraph{Baseline Methods}
We compare our Split methods against a range of established attribution approaches. Gradient-based methods include Saliency~\citep{Simonyan2014}, Input~$\times$~Gradient~\citep{Shrikumar2017}, Integrated Gradients~\citep{Sundararajan2017} and GradientSHAP~\citep{LundbergLee2017}. CAM-based methods comprise Layer-CAM~\citep{Jiang2021LayerCAM} and Grad-CAM++~\citep{Chattopadhay2018}. Among propagation-based techniques, we evaluate DeepLift~\citep{Shrikumar2017} and standard LRP with $\gamma=0.25$~\citep{Bach2015}. Finally, perturbation-based baselines include Occlusion~\citep{ZeilerFergus2014} and Feature Ablation.
For baseline $\gamma$-LRP comparisons, we use the Zennit library in the default setting~\citep{Anders2021Zennit}.
For all other baselines we use the Captum \citep{kokhlikyan2020captum} implementations in their default settings, except for GradCAM++ and LayerCAM, where we use own implementations.

\input{sections/vgg_saliency_comparison_main}

\subsection{Evaluation Framework and Metrics} \label{subsec:eval_framework}

We evaluate explanations using the Quantus library \citep{Hedstrom2023}, which provides standardized implementations of XAI metrics across key desiderata including \emph{faithfulness}, \emph{localization} and \emph{robustness}.

We adopt Quantus recommended default settings with several modifications for computational efficiency and to match common XAI evaluation practices. Full configurations and seeds are included in the supplementary code.              

Faithfulness metrics, such as \emph{Pixel Flipping} \citep{Bach2015} and \emph{Selectivity} \citep{Montavon2018}, assess whether explanations reflect the model's true decision process. Pixel Flipping measures the drop of           
the respective class logit when pixels are ``removed'' in decreasing attribution order. We perturb 112 pixels per step and replace them with the image mean, following. We report the area under the curve (AUC) of this drop at $5\%$ and $20\%$, where a faster drop means higher area under the drop curve. Selectivity perturbs pixels smoothly instead of masking.
Here, by the Quantus standard implementation a smaller value indicates a faster drop.

Localization metrics evaluate whether explanations highlight the correct spatial regions. \emph{Attribution Localization} \citep{Kohlbrenner2020} computes the ratio of attribution mass within the ground-truth segmentation          
mask and the \emph{Pointing Game} \citep{Zhang2018pointinggame} checks whether the maximum-attribution pixel lies inside the segmentation mask.
We use the segmentation masks provided by ImageNet-S for both metrics.

Robustness metrics, such as \emph{Maximum Sensitivity} \citep{Yeh2019}, quantify the stability of explanations under perturbations, where lower values indicate more stable explanations. We compute this metric with 25 random             
samples per image.
For all metrics we disable Quantus to take the absolute value of the attribution maps 
except for Maximum Sensitivity.

\FloatBarrier

\paragraph{Automated Layer Selection}
For each split based attribution method and the baselines $\gamma$-LRP, standard LayerCAM, standard GradCAM, Guided Backpropagation and Guided LayerCAM we evaluate all layer–formula combinations on the validation set across all Quantus metrics.
This selection includes the layer-wise decision if to apply the absolute value to the attribution maps and layer-wise configurations for the split methods.
We select the configuration that (i) have the best average rank in the two considered localization metrics as well as (ii) the best average rank across the Pixel Flipping metrics.
All metrics are tuned on the custom validation subset of ImageNet-S and evaluated on the custom test subset to ensure robustness and prevent overfitting.

\subsection{Results}

\subsubsection{Qualitative Analysis}

\Cref{fig:vgg_saliency_comparison_main} qualitatively compares saliency maps from our split methods with their baseline counterparts at the best-performing layers.
Mid-layer SplitCAM shares key visual characteristics with classical $\gamma$-LRP from earlier layers.
Compared to standard LayerCAM on the same layer it focuses more weight to all object pixels, while still marking the key object features as especially important.
For $\alpha \ll 0.5$, SplitGrad yields less focused yet finely aligned gradients that closely follow image details. We attribute this to the Maxout activation retaining preactivation information, unlike ReLU, which discards it in inactive neurons. 
Overall, mid-layer SplitCAM offers an effective balance between interpretability and focus (low entropy).

\subsubsection{Quantitative Analysis}

\input{sections/table_main_generated_res}

\Cref{tab:eval_vgg_main} and \cref{tab:eval_res_main} report a comprehensive comparison of attribution methods on VGG16 and ResNet18 across five key Quantus metrics: Selectivity, Attribution Localization, Pointing Game accuracy, Pixel Flipping (AUC@5 and AUC@20) and Maximum Sensitivity.
The \emph{Abs} column specifies whether absolute saliency values were used prior to evaluation in Quantus.
A full table and explanation of both VGG16 and ResNet18 evaluation results with all selected layers and other parameter configurations is found in \cref{app:saliency}.

Overall, the considered backpropagation-based attribution methods heavily depend on the selected network layer, also see \cref{app:ablation} for extensive ablation.

On VGG, the Split methods demonstrate superior performance compared to all considered baseline methods across the Quantus metric categories.
For example, SplitCAM achieves the highest Pointing Game score \textbf{0.938} with $\alpha = 0.4$ overshadowing the best baseline Guided Backprop in layer 5 with $0.887$ and LayerCAM in layer 2 with $0.855$.
SplitCAM outperforms classical methods on Selectivity, with comparable Pixel Flipping performance as the most faithful baseline LayerCAM.
SplitLRP slightly outperforms classical $\gamma$-LRP in localization while staying competitive in Faithfulness.

On ResNet18, the Split method slightly improve the Selectivity and Pointing Game metrics of the strongest baseline LayerCAM, while staying competitive with the other baselines across the considered metrics.

\subsubsection{Ablation studies.} \label{subsec:ablation}

\input{sections/alpha_ablation_main}

Table \ref{tab:alpha_ablation_main} presents the performance of two selected SplitGrad and SplitCAM configurations across different shifting parameters $\alpha$, evaluated on our custom ImageNet-S test split.
As explained in \cref{sec:backward} the $\alpha$ value determines how much we stablize the backward pass, where $\alpha=0$ results in no stabilization and $\alpha = 0.5$ results in local sensitivities resembling scaled versions of the original network gradient.
It is evident that performance for all considered metrics increases with $\alpha$ smaller than $0.5$ but diminishes for very small $\alpha$ values, except Maximum Sensitivity is best for $\alpha=0$. This demonstrates that both the splitting and the backward pass stabilization boost our explanations.
We provide further ablation studies in \cref{app:ablation}.

%% file: sections/vgg_saliency_comparison_main.tex
\newcommand{\colA}{0.10}
\newcommand{\colB}{0.075}
\newcommand{\colSpace}{0.0015\linewidth}

\begin{figure*}[!t]
    \centering

    \begin{minipage}[c]{\colA\linewidth}\centering\scriptsize\textbf{Class}\end{minipage}\hspace{\colSpace}%
    \begin{minipage}[c]{\colB\linewidth}\centering\scriptsize\textbf{Original}\end{minipage}\hspace{\colSpace}%
    \begin{minipage}[c]{\colB\linewidth}\centering\scriptsize\textbf{SplitLRP}\\\scriptsize\textbf{$R^{(10,\text{pos})}$ abs}\end{minipage}\hspace{\colSpace}%
    \begin{minipage}[c]{\colB\linewidth}\centering\scriptsize\textbf{LRP $\gamma$=0.25}\\\scriptsize\textbf{layer 5 abs}\end{minipage}\hspace{\colSpace}%
    \begin{minipage}[c]{\colB\linewidth}\centering\scriptsize\textbf{SplitCAM}\\\scriptsize\textbf{$\alpha=0.4$}\\\scriptsize\textbf{$\delta^{(14,g)}$}\end{minipage}\hspace{\colSpace}%
    \begin{minipage}[c]{\colB\linewidth}\centering\scriptsize\textbf{LayerCAM}\\\scriptsize\textbf{layer 14}\end{minipage}\hspace{\colSpace}%
    \begin{minipage}[c]{\colB\linewidth}\centering\scriptsize\textbf{SplitCAM}\\\scriptsize\textbf{$\alpha=0.3$}\\\scriptsize\textbf{$\delta^{(26,g)}$ abs}\end{minipage}\hspace{\colSpace}%
    \begin{minipage}[c]{\colB\linewidth}\centering\scriptsize\textbf{LayerCAM}\\\scriptsize\textbf{layer 26}\end{minipage}\hspace{\colSpace}%
    \begin{minipage}[c]{\colB\linewidth}\centering\scriptsize\textbf{SplitGrad}\\\scriptsize\textbf{$\alpha=0.3$}\\\scriptsize\textbf{$\delta^{(0,+,g)}$}\end{minipage}

    \vspace{0.25cm}

    \begin{minipage}[c]{\colA\linewidth}
        \centering
        \small goldfinch\\[1pt]
        {\scriptsize (0.9996)}
    \end{minipage}\hspace{\colSpace}%
    \begin{minipage}[c]{\colB\linewidth}
        \centering
        \includegraphics[width=\linewidth]{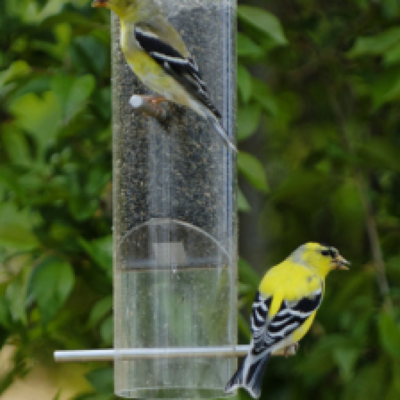}
    \end{minipage}\hspace{\colSpace}%
    \begin{minipage}[c]{\colB\linewidth}
        \centering
        \includegraphics[width=\linewidth]{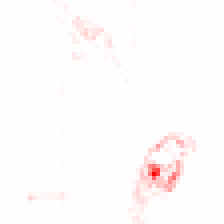}
    \end{minipage}\hspace{\colSpace}%
    \begin{minipage}[c]{\colB\linewidth}
        \centering
        \includegraphics[width=\linewidth]{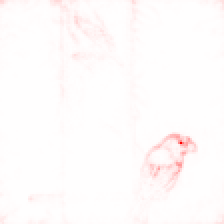}
    \end{minipage}\hspace{\colSpace}%
    \begin{minipage}[c]{\colB\linewidth}
        \centering
        \includegraphics[width=\linewidth]{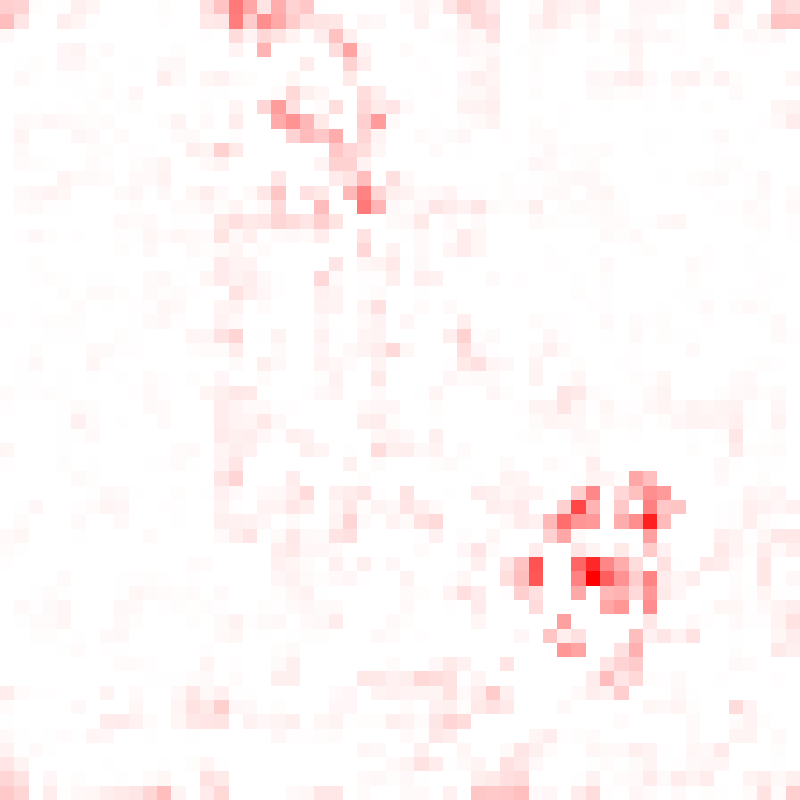}
    \end{minipage}\hspace{\colSpace}%
    \begin{minipage}[c]{\colB\linewidth}
        \centering
        \includegraphics[width=\linewidth]{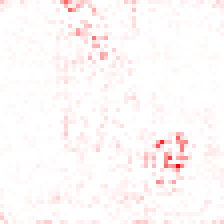}
    \end{minipage}\hspace{\colSpace}%
    \begin{minipage}[c]{\colB\linewidth}
        \centering
        \includegraphics[width=\linewidth]{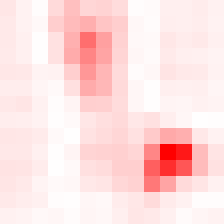}
    \end{minipage}\hspace{\colSpace}%
    \begin{minipage}[c]{\colB\linewidth}
        \centering
        \includegraphics[width=\linewidth]{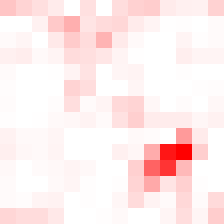}
    \end{minipage}\hspace{\colSpace}%
    \begin{minipage}[c]{\colB\linewidth}
        \centering
        \includegraphics[width=\linewidth]{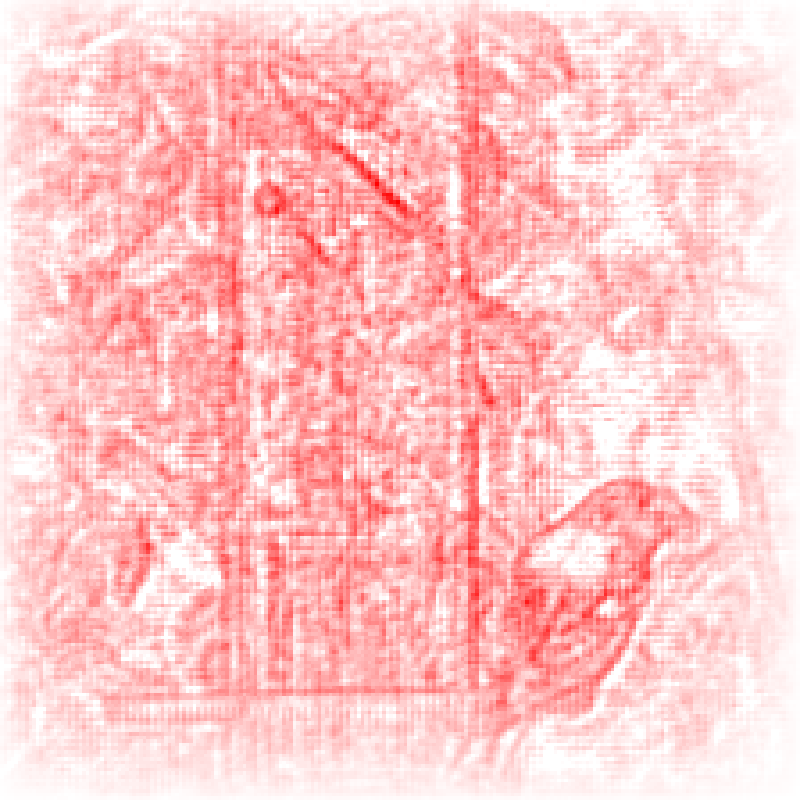}
    \end{minipage}%

    \vspace{0.25cm}

    \begin{minipage}[c]{\colA\linewidth}
        \centering
        \small sulphur butterfly\\[1pt]
        {\scriptsize (0.9999)}
    \end{minipage}\hspace{\colSpace}%
    \begin{minipage}[c]{\colB\linewidth}
        \centering
        \includegraphics[width=\linewidth]{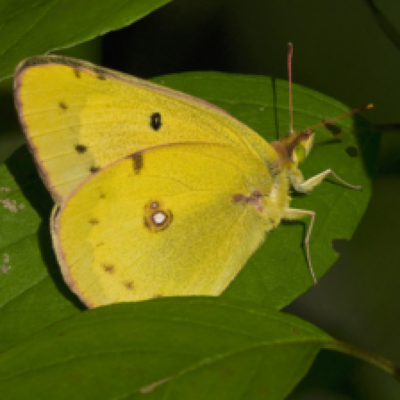}
    \end{minipage}\hspace{\colSpace}%
    \begin{minipage}[c]{\colB\linewidth}
        \centering
        \includegraphics[width=\linewidth]{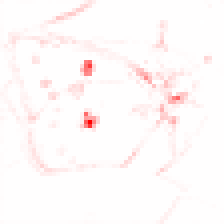}
    \end{minipage}\hspace{\colSpace}%
    \begin{minipage}[c]{\colB\linewidth}
        \centering
        \includegraphics[width=\linewidth]{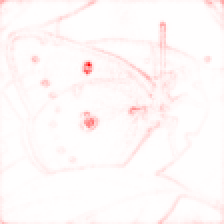}
    \end{minipage}\hspace{\colSpace}%
    \begin{minipage}[c]{\colB\linewidth}
        \centering
        \includegraphics[width=\linewidth]{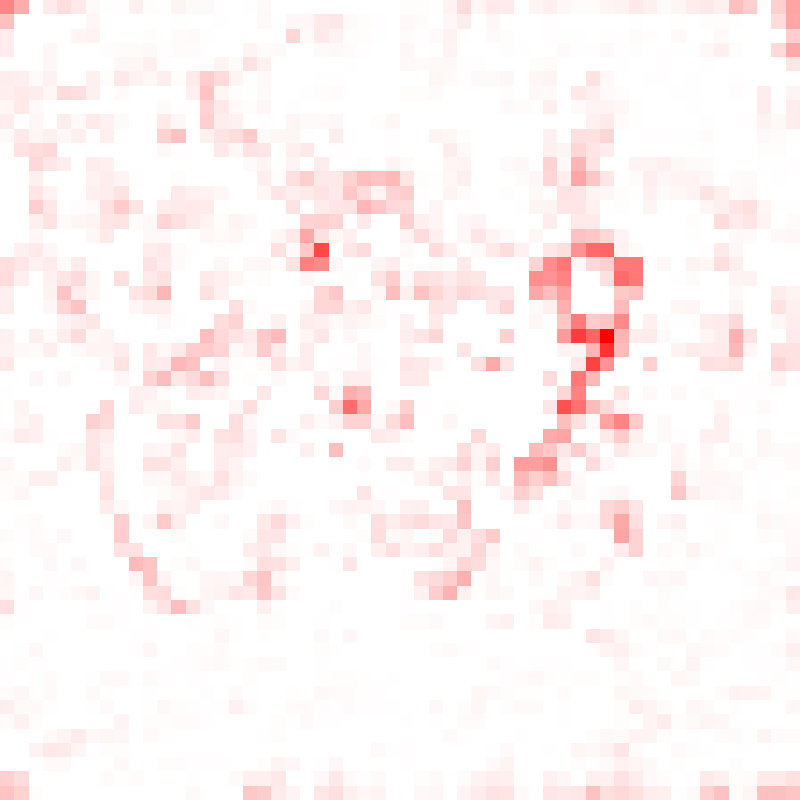}
    \end{minipage}\hspace{\colSpace}%
    \begin{minipage}[c]{\colB\linewidth}
        \centering
        \includegraphics[width=\linewidth]{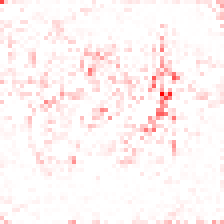}
    \end{minipage}\hspace{\colSpace}%
    \begin{minipage}[c]{\colB\linewidth}
        \centering
        \includegraphics[width=\linewidth]{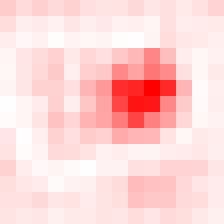}
    \end{minipage}\hspace{\colSpace}%
    \begin{minipage}[c]{\colB\linewidth}
        \centering
        \includegraphics[width=\linewidth]{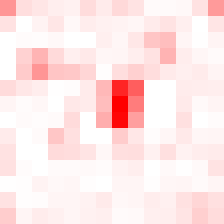}
    \end{minipage}\hspace{\colSpace}%
    \begin{minipage}[c]{\colB\linewidth}
        \centering
        \includegraphics[width=\linewidth]{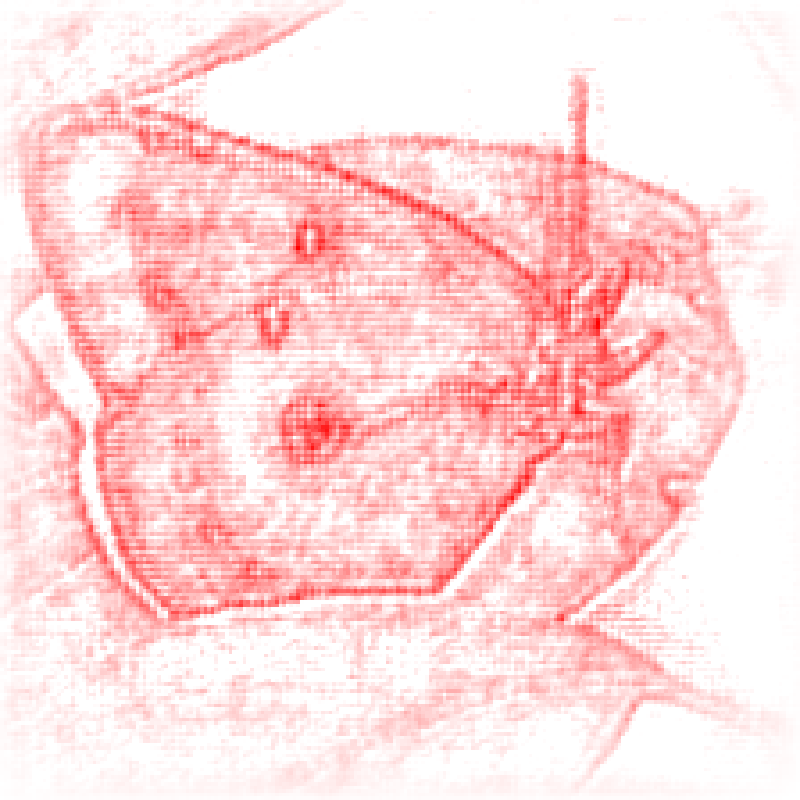}
    \end{minipage}%

    \vspace{0.25cm}

    \begin{minipage}[c]{\colA\linewidth}
        \centering
        \small goldfish\\[1pt]
        {\scriptsize (1.0000)}
    \end{minipage}\hspace{\colSpace}%
    \begin{minipage}[c]{\colB\linewidth}
        \centering
        \includegraphics[width=\linewidth]{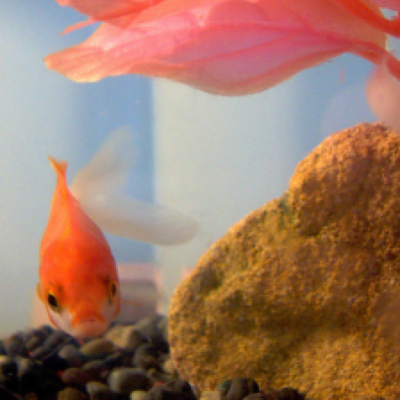}
    \end{minipage}\hspace{\colSpace}%
    \begin{minipage}[c]{\colB\linewidth}
        \centering
        \includegraphics[width=\linewidth]{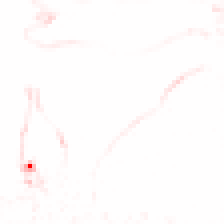}
    \end{minipage}\hspace{\colSpace}%
    \begin{minipage}[c]{\colB\linewidth}
        \centering
        \includegraphics[width=\linewidth]{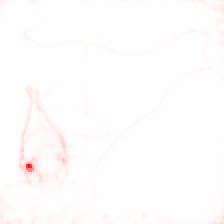}
    \end{minipage}\hspace{\colSpace}%
    \begin{minipage}[c]{\colB\linewidth}
        \centering
        \includegraphics[width=\linewidth]{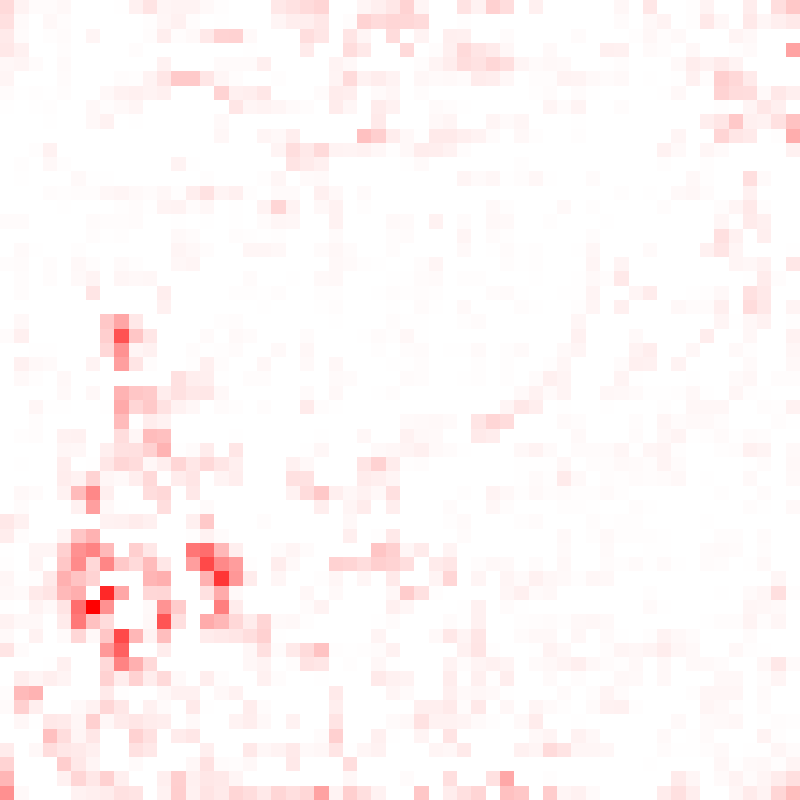}
    \end{minipage}\hspace{\colSpace}%
    \begin{minipage}[c]{\colB\linewidth}
        \centering
        \includegraphics[width=\linewidth]{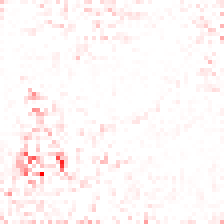}
    \end{minipage}\hspace{\colSpace}%
    \begin{minipage}[c]{\colB\linewidth}
        \centering
        \includegraphics[width=\linewidth]{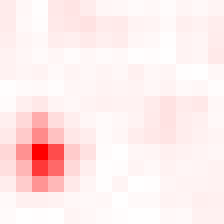}
    \end{minipage}\hspace{\colSpace}%
    \begin{minipage}[c]{\colB\linewidth}
        \centering
        \includegraphics[width=\linewidth]{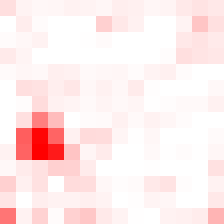}
    \end{minipage}\hspace{\colSpace}%
    \begin{minipage}[c]{\colB\linewidth}
        \centering
        \includegraphics[width=\linewidth]{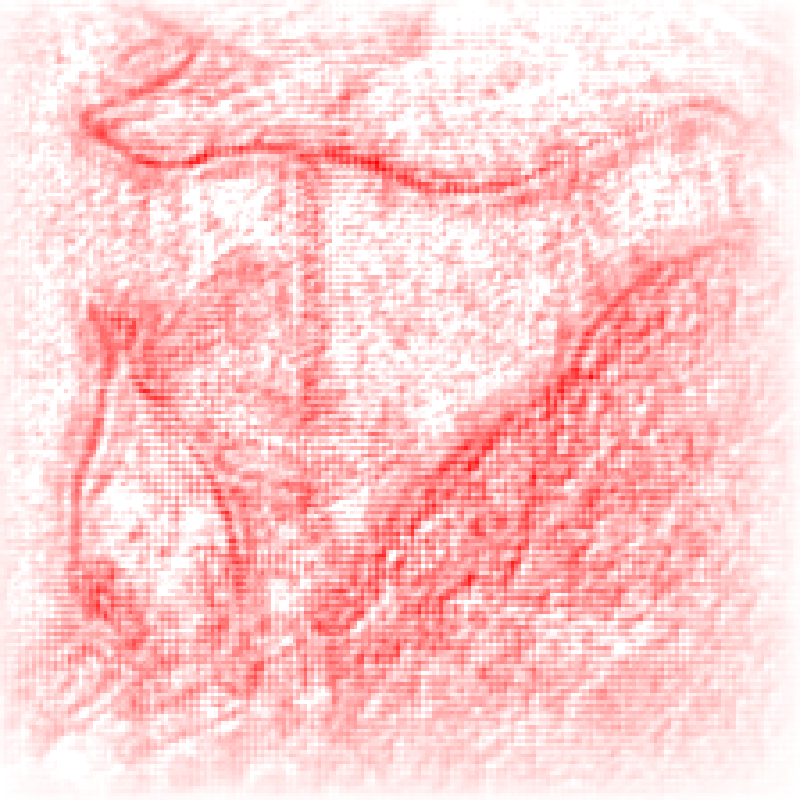}
    \end{minipage}%

    \caption{Saliency map comparison for VGG16 on our custom ImageNet-S test split. Each row shows the predicted class (with confidence), original image, and saliency maps from different methods.}
    \label{fig:vgg_saliency_comparison_main}
\end{figure*}

%% file: sections/table_main_generated_res.tex
\begin{table*}[htb!]
    \centering
    \small
    \caption{Comprehensive XAI Evaluation Results (ResNet18). The triples $x.y.z$ are a ResNet18 internal indexing of convolutional layer, at which the attribution maps are calculated. See the caption of \cref{tab:eval_res_main} for further abbreviations and table explanaition.}
    \label{tab:eval_res_main}
    \begin{tabular}{p{4cm}|cc|cccccc}
        \toprule
        \textbf{Method} & \textbf{Layer} & \textbf{Abs} & \textbf{Select.} & \textbf{Attr. Loc.} & \textbf{Point.} & \textbf{Pixel Fl. (AUC@5)} & \textbf{Pixel Fl. (AUC@20)} & \textbf{Max Sens.} \\
         & & & \textbf{$\downarrow$} & \textbf{$\uparrow$} & \textbf{$\uparrow$} & \textbf{$\uparrow$} & \textbf{$\uparrow$} & \textbf{$\downarrow$} \\
        
         \midrule
        \multicolumn{9}{c}{\textit{SplitCAM Methods}} \\
        \midrule
        SplitCAM (sc, $\alpha = 0.3$, wta) & (1.0.2, g) & n & \fboxsep=0.5pt\colorbox{red!20}{\textbf{3.503}} & 0.487 & 0.747 & 0.168 & 1.395 & 1.277 \\
        SplitCAM (sc, $\alpha = 0.3$, wta) & (4.1.2, g) & n & 5.654 & \fboxsep=0.5pt\colorbox{red!20}{\textbf{0.673}} & \fboxsep=0.5pt\colorbox{red!20}{\textbf{0.972}} & 0.081 & 0.880 & 0.233 \\
        SplitGrad ($\alpha = 0.3$, wta) & (1.0.2, g) & y & 4.449 & 0.513 & 0.777 & 0.149 & 1.266 & 0.813 \\
        SplitGrad ($\alpha = 0.3$, wta) & (4.1.2, g) & y & 5.663 & 0.649 & 0.972 & 0.082 & 0.882 & 0.219 \\
        SplitLRP (sc, wta) & (conv1, pos) & y & 6.313 & 0.506 & 0.587 & 0.056 & 0.697 & 0.755 \\
        \midrule
        \multicolumn{9}{c}{\textit{Baselines (With Layer Optimization)}} \\
        \midrule
        GradCAM & 4.1.2 & n & 6.070 & 0.554 & 0.945 & 0.076 & 0.829 & \fboxsep=0.5pt\colorbox{red!20}{\textbf{0.131}} \\
        LayerCAM & 2.0.2 & n & 3.690 & 0.549 & 0.865 & 0.144 & 1.428 & 0.913 \\
        LayerCAM & 4.1.1 & n & 5.661 & 0.664 & 0.971 & 0.083 & 0.890 & 0.227 \\
        LRP $\gamma=0.25$ & conv1 & y & 3.550 & 0.656 & 0.852 & \fboxsep=0.5pt\colorbox{red!20}{\textbf{0.213}} & 1.737 & 0.521 \\
        \midrule
        \multicolumn{9}{c}{\textit{Baselines}} \\
        \midrule
        Guided Backprop & conv1 & n & 3.517 & 0.621 & 0.867 & 0.212 & \fboxsep=0.5pt\colorbox{red!20}{\textbf{1.757}} & 0.459 \\
        Gradient SHAP & conv1 & y & 5.269 & 0.527 & 0.779 & 0.166 & 1.227 & 1.074 \\
        Integrated Gradients & conv1 & y & 5.265 & 0.528 & 0.777 & 0.167 & 1.226 & 0.994 \\
        Feature Ablation & conv1 & n & 4.655 & 0.136 & 0.684 & 0.188 & 1.388 & 1.077 \\
        Occlusion & conv1 & n & 5.300 & 0.179 & 0.726 & 0.130 & 1.030 & 0.785 \\
        Deconvolution & conv1 & n & 3.934 & 0.484 & 0.666 & 0.147 & 1.203 & 0.751 \\
        Random Baseline & - & y & 4.071 & 0.425 & 0.440 & 0.147 & 1.067 & -- \\
        \bottomrule
    \end{tabular}
\end{table*}

%% file: sections/alpha_ablation_main.tex
\begin{table}[h]
    \centering
    \footnotesize

    \resizebox{1.0\columnwidth}{!}{%
    \begin{tabular}{|l|l|c|c|c|c|c|c|}
    \hhline{|-|-|------|}
    \textbf{Method} & \textbf{Metric} & $\alpha \ .50$ & $\alpha \ .45$ & $\alpha \ .40$ & $\alpha \ .35$ & $\alpha \ .30$ & $\alpha \ .00$ \\
    \hhline{|-|-|------|}
    \multirow{4}{*}{\makecell{\textbf{SplitGrad}\\(26, g)}} & Point. $\uparrow$ & 0.898 & 0.906 & 0.947 & \textbf{0.956} & 0.889 & 0.401 \\
    \hhline{|~|-|------|}
     & Attr. Loc. $\uparrow$ & 0.543 & 0.548 & 0.605 & \textbf{0.617} & 0.535 & 0.449 \\
    \hhline{|~|-|------|}
     & Max Sens. $\downarrow$ & 0.616 & 0.607 & 0.440 & 0.238 & 0.334 & \textbf{0.028} \\
    \hhline{|~|-|------|}
     & Pixel Fl. @20 $\uparrow$ & 1.527 & 1.534 & \textbf{1.581} & 1.448 & 1.317 & 1.247 \\
    \hhline{|-|-|------|}
    \multirow{4}{*}{\makecell{\textbf{SplitCAM}\\(26, +, g)}} & Point. $\uparrow$ & 0.868 & 0.868 & \textbf{0.938} & 0.933 & 0.613 & 0.855 \\
    \hhline{|~|-|------|}
     & Attr. Loc. $\uparrow$ & 0.550 & 0.554 & 0.635 & \textbf{0.663} & 0.403 & 0.513 \\
    \hhline{|~|-|------|}
     & Max Sens. $\downarrow$ & 0.691 & 0.682 & 0.457 & 0.265 & 0.251 & \textbf{0.120} \\
    \hhline{|~|-|------|}
     & Pixel Fl. @20 $\uparrow$ & 1.607 & \textbf{1.609} & 1.585 & 1.499 & 0.853 & 1.507 \\
    \hhline{|-|-|------|}
    \end{tabular}
    }%
    
    \caption{Comparison of different $\alpha$ parameters with two SplitGrad and SplitCAM base configurations, both in \emph{wta} and \emph{abs} mode.}
    \label{tab:alpha_ablation_main}
\end{table}

%% file: sections/conclusion.tex
\section{Conclusion}\label{sec:conclusion}

We have presented a numerically stable framework for decomposing pretrained ReLU networks into split-pairs of monotone and convex split-streams, enabling improved explainability without retraining or altering model predictions.
While there is no single saliency method excellent in \emph{all} metrics at the same time, our experiments establish splitting as a tool to boost various of them on trained networks. 
By directly training DC-models, we initiate their application as a further approach to increase explainability of a network. 
Especially the separation between factual and counterfactual attributes observed in the MNIST experiments is remarkable and may be understood further theoretically and experimentally.

While we provided some details on the splitting of feedforward networks, with similar ideas, this can be extended to further architectures. 
We provide a novel mathematical framework to decompose multiplicative networks as the transformer \citep{Vaswani2017} into the difference of monotone and convex parts in \cref{app:splitting_multiplicative_maxout_networks}.
It is future work to analyze the properties of the proposed decomposition approach for such multiplicative networks and to adapt SplitCAM and SplitLRP to the transformer architecture.

The split network need not duplicate parameters: instead of materializing a separate model, one can traverse the original \texttt{nn.Module} graph and realize the split using masked views of the same weights. It is future work to implement the network decomposition framework more memory efficient without duplicating model weights.

Our experiments show that attribution quality heavily depends on layer selection, with different layers excelling at different metrics. 
This motivates to develop principled methods for automatic layer selection or combining attributions across multiple layers for more robust explanations.

%% file: appendix/mathematical_perspective.tex
\section{Mathematical Perspective: Support Functions and Imaginative Polytopes}\label{app:imaginative_polytopes}

\subsection{Preliminaries}\label{subsec:support_app}
A \emph{polytope} $P\subseteq\RR^d$ is the convex hull of finitely many vertices in $\RR^d$, or by the Minkowski–Weyl Theorem any finite intersection of finitely many closed halfspaces \citeapp{AppZiegler1995}.
For a polytope $P\subseteq\RR^d$, the \emph{support function} $f_P:\RR^d\to\RR$ is
\begin{align}
    f_P(x) \coloneqq \max_{p\in P} \langle p, x \rangle_{L^2}.
\end{align}
We call the set of points in $P$ that cannot be represented as a nontrivial convex combination of two distinct points in $P$ the vertices of $P$ and denote them by $\text{vert}(P)$.
It is easy to see that
\begin{align}
    f_P(x) \coloneqq \max_{p\in \text{vert}(P)} \langle p, x \rangle_{L^2}. \label{eq:support_function}
\end{align}

A function $f:\mathbb{R}^n \to \mathbb{R}$ is \emph{piecewise linear} (PWL) if there exists a finite polyhedral partition of $\mathbb{R}^n$ such that $f$ is affine on each piece.
It is \emph{continuous piecewise linear} (CPWL) if it is also continuous. 
Note that a function is CPWL exactly if it can be represented by a ReLU network \citeapp{AppArora2018}.

Given the representation \cref{eq:support_function} it is easy to see that the vertices of $P$ correspond one by one to all attainable gradients of $f_P$.
Further, we observe from \cref{eq:support_function} that support functions are convex, continuous, positively homogeneous ($f_P(\lambda c)=\lambda f_P(c)$ for $\lambda\ge 0$), and CPWL. Indeed, the mapping
\begin{align}
    \Phi:\ \mathsf{Polytopes}_d &\longrightarrow \mathsf{CPWL}^+_d, \notag \\
    P &\longmapsto f_P, \label{eq:bijection_cpwl}
\end{align}
is a bijection between $d$-dimensional polytopes and convex, positively homogeneous CPWL functions.

Tropical geometry refers to $P$ as the \emph{Newton polytope} of $f_P$; the correspondence and its learning-theoretic consequences have been exploited in analyses of neural networks and their expressivity \citepapp{AppZhang2018}, \citepapp{AppHaase2023}, \citepapp{AppBrandenburg2024b}.

We call a function $f:\mathbb{R}^n \to \mathbb{R}$ \emph{monotone} in case for any $x_1, x_2 \in \mathbb{R}^d$ with $x_1 \leq x_2$ (understood elementwise) one has $f(x_1) \leq f(x_2)$.

As shown in \cref{app:fundamental}, the split-streams $g$ and $h$ defined in \cref{sec:method} are positively homogeneous monotone convex CPWL functions and therefore possess corresponding Newton polytopes $P_g$ and $P_h$ by the bijection in \cref{eq:bijection_cpwl}.
It is easy to see that by the monotonicity of $g$ and $h$ all points in $P_g$ and $P_h$ have non-negative coordinates as the gradients of $g$ and $h$ are non-negative.
\TODO{GL: Ist die folgende Aussage nicht auch schon im Abschnitt A.1 implizit?}
\TODO{JZ: Ich glaube, ich habe den remark zu positiven vertices wieder gelöscht gehabt. Ich habe das Theorem hinzugefügt um klarzumachen, dass unsere Intuition zu positiven Gradienten nicht nur NNs mit positiven Gewichten betrifft. Das Argument ist natürlich sehr einfach und ich finde es auch implizit in der maximum formulierung der support funktion.}
\begin{theorem} \label{thm:monotonicity_positive_gradients}
    Let $g: \ \mathbb{R}^{d} \longrightarrow \mathbb{R}$ be a monotone and CPWL function. Then $\nabla g \geq 0$, \ie all its attainable gradients are non-negative.
\end{theorem}

\begin{proofref}{Theorem}{thm:monotonicity_positive_gradients}
    Assume for a contradiction that there is $x \in \mathbb{R}^d$ and $j \in [d]$ such that the $j$th component of the gradient at $x$ is negative, \ie $\left( \frac{\partial g}{\partial x} \right)_j < 0$.
    As $g$ is piecewise linear, there is $\epsilon > 0$ such that $g$ is linear in the $\epsilon$-ball around $x$. However for $\tilde{x} \coloneqq x - \frac{\epsilon}{2} e_j$ one finds $\tilde{x} \leq x$ but $f\left( \tilde{x} \right) > f(x)$. This contradiction concludes the proof of \cref{thm:monotonicity_positive_gradients}.
\end{proofref}

\subsection{Imaginative Polytopes}\label{subsec:imaginative_polytopes}
In the ablation study \cref{app:ablation} we observe that the gradients of the split-pair show greater robustness than the gradients of the original model, \ie small input perturbation results in less perturbed gradients.
We want to argue that indeed the non-negativity of the split-stream gradient could be one important reason for this behavior.
Indeed, as the split-stream outputs are just the maximum scalar products of the input with all attainable gradients respectively, see \cref{eq:support_function}, the non-negativity of the gradients results in more stable gradients due to no cancellation in the scalar product. 
Further, any jump of the gradients in the original network does not force both gradients of $g$ and $h$ to jump, but only one of them.

Secondly, we observe dramatically better localization properties of the split-pair gradients compared to the original gradients. Again, we find it intuitive that positive gradients show better focus on salient image regions as no cancellation can happen.

\begin{figure}[htbp]
    \centering
    \includegraphics[width=0.6\textwidth]{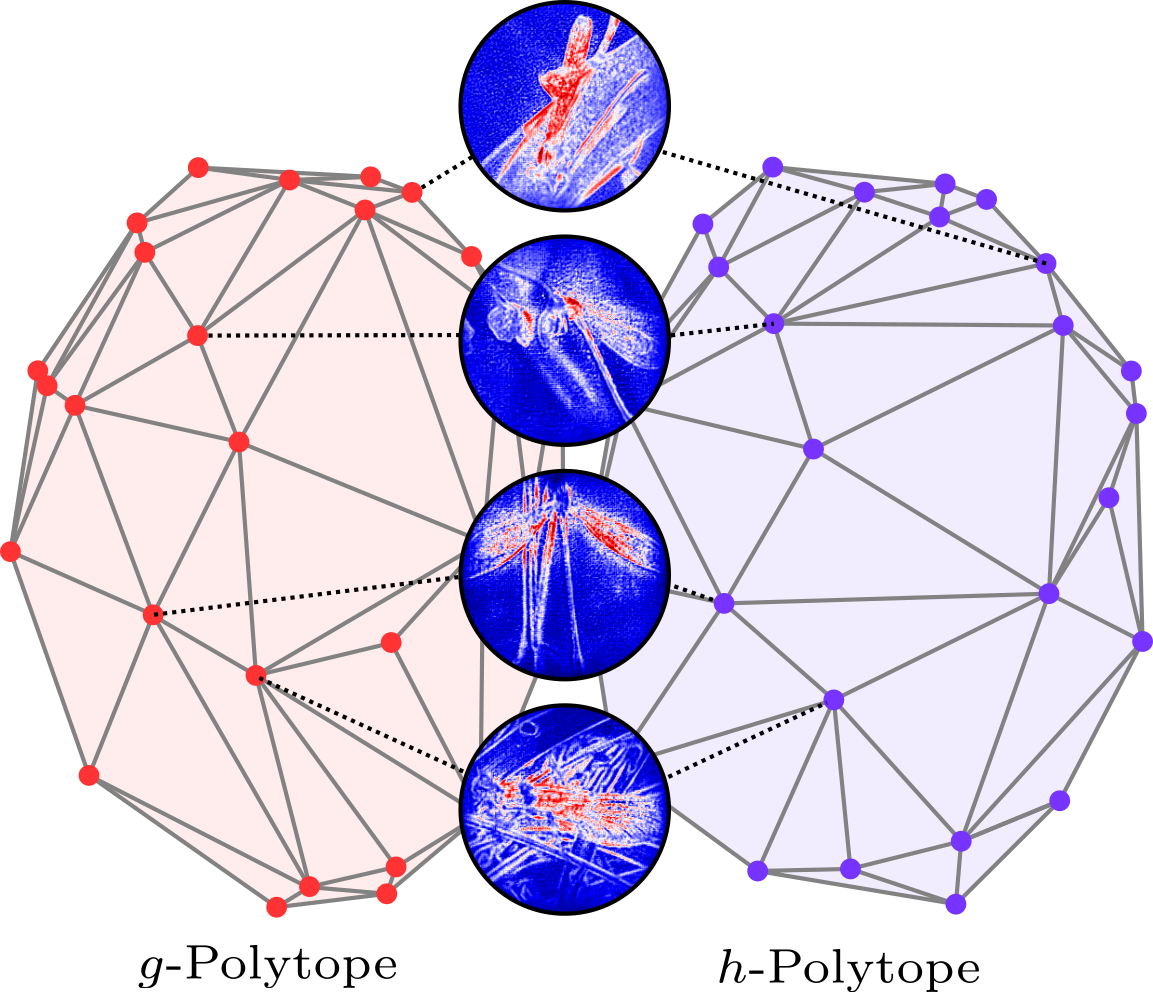}
    \caption{Concept visualization of the imaginative polytopes corresponding to VGG's dragonfly class output neuron. The polytope's vertices represent the $g$-stream and $h$-stream gradients respectively, revealing learned visual prototypes.}
    \label{fig:imaginative_polytope}
\end{figure}

Because the vertices of $P_g$ and $P_h$ empirically align closely with salient image structures we term them \emph{Imaginative Polytopes}: intrinsic geometric objects induced by ReLU networks that summarize the model's learned visual features and thereby help to explain how common vision models envision and ``imagine'' the world. Their vertices correspond to distinct activation patterns and represent the fundamental ``prototypes'' that the network has learned to distinguish different input features.
We visualize the $g$-stream gradient of VGG16 with respect to the ``dragonfly'' class output logit for four ImageNet test images in \cref{fig:imaginative_polytope}, where we applied backward pass shifting with $\alpha=0.25$. We remark that the $h$-stream gradient looks very similar, as it only differs to the $g$-stream gradient in the gradient of the original VGG model -- for this reason we linked the $g$-stream gradient visualizations also with corresponding $h$-stream gradients.

%% file: appendix/saliency_analysis.tex
\newpage

\section{Saliency Analysis} \label{app:saliency}

This appendix provides comprehensive evaluation results for all XAI methods tested in the experimental setup described in \cref{subsec:eval_framework}.

\subsection{VGG16 Qualitative Saliency Map Comparison} \label{app:vgg_qualitative}

Complementing \cref{fig:vgg_saliency_comparison_main}, we give more saliency examples in \cref{fig:vgg_saliency_comparison}. Again, we observe more focus on the whole classified objects by SplitCAM in layer 26 compared to LayerCAM in the same layer.
SplitGrad in layer 0 detects too many edges of the original image to be a useful explanation of the model's decision. Especially in images with many image details in the background, \eg the rubble in the wood rabbit image, it fails to focus on the important image parts. 
This aligns with the observed performance peaks of SplitGrad and SplitCAM in deeper layers, presented in \cref{fig:layer_dependence} in the layer ablation studies.

\input{appendix/figures/vgg_saliency_comparison}

\newpage

\subsection{Quantitative Analysis on VGG16} \label{app:vgg_split}

\Cref{tab:xai_eval_comprehensive_vgg} presents detailed quantitative metrics analysis across all split configurations, baselines, and classical methods for VGG16. 
We observe a slight boost in performance between SplitCAM with $\alpha = 0.4$ compared to $\alpha = 0.5$ and compared to the classical LayerCAM respectively, indicating that the LayerCAM formulas indeed yield more powerful explanations on the split-pair compared to the original model. This can be seen even more drastically in ablation studies \cref{app:ablation}.

\input{appendix/table_appendix_generated_vgg_full_eval_sorted}

\subsection{Quantitative Analysis on ResNet18} \label{app:resnet_split}

\Cref{tab:xai_eval_comprehensive_res} presents detailed quantitative metrics across all split configurations, baselines, and classical methods for ResNet18.

\input{appendix/table_appendix_generated_res_full_eval_sorted}

\subsection{Randomization Sanity Check} \label{app:sanity}

In this subsection, we apply the model randomization sanity check introduced by \citetapp{AppAdebayo2018}. 
The test evaluates whether a saliency method truly depends on the model’s learned parameters 
or whether it merely reflects input-dependent patterns that remain even when the model is corrupted.
For each attribution method, we compute saliency maps on the trained VGG16 model and compare them 
to saliency maps obtained after randomly reinitializing the weights of the last three feed-forward 
layers using Gaussian noise. A method that is genuinely sensitive to the model's trained parameters should show a substantial
drop in similarity between the two saliency maps, as proposed by the original paper.
While \citetapp{AppBinder2023} discuss important shortcomings of top-down randomization-based sanity checks, we apply this test as a complementary diagnostic tool to assess the relative sensitivity of split-based methods to learned model parameters.

We quantify similarity using three metrics:
\textbf{(i) HOG similarity}, which measures agreement of gradient-orientation structure via the Pearson correlation of the two Histogram of Oriented Gradients; 
\textbf{(ii) Spearman rank correlation (signed)}, which captures preservation of the pixel-wise 
ordering and direction of attributions; and 
\textbf{(iii) SSIM}, the Structural Similarity Index, assessing perceptual similarity based on 
luminance, contrast, and structural information. 
Lower similarity values indicate stronger dependence on learned model parameters.

\input{appendix/randomization_sanity_check_table}

\paragraph{Analysis}
The results in \cref{tab:randomization_sanity_check} highlight that the backward-pass
stabilization used in SplitCAM and SplitGrad effectively mitigates the gradient
explosion issues caused by the non-negative split-stream weights. 
In contrast, SplitLRP remains fairly insensitive to model randomization: across HOG and Spearman similarity metrics it shows values close to one (HOG~$\approx 0.99$,
Spearman~$\approx 1.00$). The change of attribution maps when the final VGG16 layers are reinitialized is only measured by the SSIM metric (SSIM~$\approx 0.45$--$0.56$).
By contrast, our SplitCAM and SplitGrad variants show substantially lower similarities than all of the considered baselines, 
particularly in Spearman and SSIM, demonstrating their exceptionally strong dependence on the learned weights and the model's actual decision process.

We attribute the insensitivity of SplitLRP to the last three VGG16 layer weights to the gradient explosion inherent to the split-stream due to its non-negative weight matrices. 
Since we apply the $\epsilon$-LRP rule, which is closely related to the vanilla LRP rule and effectively reduces to a scaled version of $\text{Input}\times\text{Gradient}$ on the split-stream, the resulting SplitLRP attributions suffer from a diminishing-gradient effect. 
This makes changes in the final-layer weights largely insignificant for the propagated relevance. Consequently, the deeper the considered split layer, the more stable the SplitLRP maps become under randomization. Indeed, the layer-26 SplitLRP configuration performs `best' (i.e., shows the lowest similarity to the original maps) across all three metrics and SplitLRP configurations, confirming its stronger sensitivity to the learned parameters compared to early-layer SplitLRP attribution maps. Here, the scaling in the backward pass of SplitGrad and SplitCAM described and analyzed in \cref{app:closed_formulas} proves powerful.

\subsection{Computational Efficiency Analysis} \label{app:timing}

To assess the computational overhead of our split-based methods, we measure mean computation times across the custom 50 image validation subset of ImageNet-S for both ResNet18 and VGG16 architectures.
The model splitting operation itself is performed once per model. However, we averaged the running time over 20 split procedures for each of the models. It requires only 0.033 seconds for ResNet18 and 0.030 seconds for VGG16, representing negligible overhead.
\Cref{tab:timing_comparison} presents the mean saliency computation times, revealing that SplitCAM and SplitGrad achieve competitive performance (0.33--0.44 seconds per image) compared to classical CAM-based methods, while being significantly faster than perturbation-based approaches such as Feature Ablation and Occlusion.
Notably, the shifting parameter $\alpha$ used in SplitCAM and SplitGrad does not affect runtime, as it only modifies the forward pass computations without introducing additional computational complexity. For comparability in this test we compute the attribution maps with respect to the first model layer for all methods.

We note that we manually implemented the split-pair forward and backward passes in PyTorch without using its hooking system. 
Further, we build the split-pair as an independent module and duplicate all weight matrices when splitting into their positive and negative part. 
As a result, our current implementation is not fully optimized; both running time and memory efficiency can be improved substantially. In principle, the split stream can operate directly on the original model weights, making weight duplication unnecessary. Moreover, the forward and backward computations for each layer can be parallelized, so the overall overhead relative to the standard forward and backward pass of the model is expected to be modest.

\input{appendix/table_timing_comparison}

%% file: appendix/figures/vgg_saliency_comparison.tex
\begin{figure}[H]
    \centering

    \begin{minipage}[c]{0.08\linewidth}
        \centering
        \scriptsize \textbf{Class}
    \end{minipage}\hspace{0.002\linewidth}%
    \begin{minipage}[c]{0.10\linewidth}
        \centering
        \scriptsize \textbf{Original}
    \end{minipage}\hspace{0.002\linewidth}%
    \begin{minipage}[c]{0.10\linewidth}
        \centering
        \scriptsize \textbf{SplitLRP}\\
        \scriptsize \textbf{(10, pos) abs}
    \end{minipage}\hspace{0.002\linewidth}%
    \begin{minipage}[c]{0.10\linewidth}
        \centering
        \scriptsize \textbf{LRP $\gamma$=0.25}\\
        \scriptsize \textbf{layer 5 abs}
    \end{minipage}\hspace{0.002\linewidth}%
    \begin{minipage}[c]{0.10\linewidth}
        \centering
        \scriptsize \textbf{SplitCAM}\\
        \scriptsize \textbf{(14, g) $\alpha=0.4$}
    \end{minipage}\hspace{0.002\linewidth}%
    \begin{minipage}[c]{0.10\linewidth}
        \centering
        \scriptsize \textbf{LayerCAM}\\
        \scriptsize \textbf{layer 14}
    \end{minipage}\hspace{0.002\linewidth}%
    \begin{minipage}[c]{0.10\linewidth}
        \centering
        \scriptsize \textbf{SplitCAM}\\
        \scriptsize \textbf{(26, g) $\alpha=0.3$ abs}
    \end{minipage}\hspace{0.002\linewidth}%
    \begin{minipage}[c]{0.10\linewidth}
        \centering
        \scriptsize \textbf{LayerCAM}\\
        \scriptsize \textbf{layer 26}
    \end{minipage}\hspace{0.002\linewidth}%
    \begin{minipage}[c]{0.10\linewidth}
        \centering
        \scriptsize \textbf{SplitGrad}\\
        \scriptsize \textbf{(0, +, g)}
    \end{minipage}

    \vspace{0.3cm}

    \begin{minipage}[c]{0.08\linewidth}
        \centering
        \small tiger shark\\[1pt]
        {\scriptsize (0.9979)}
    \end{minipage}\hspace{0.002\linewidth}%
    \begin{minipage}[c]{0.10\linewidth}
        \centering
        \includegraphics[width=\linewidth]{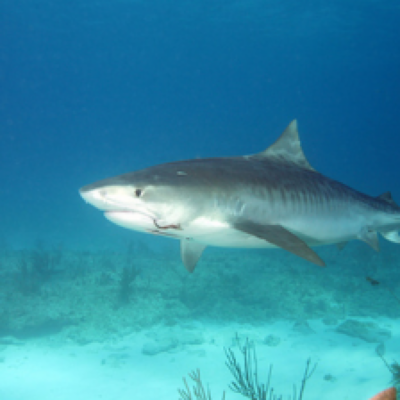}
    \end{minipage}\hspace{0.002\linewidth}%
    \begin{minipage}[c]{0.10\linewidth}
        \centering
        \includegraphics[width=\linewidth]{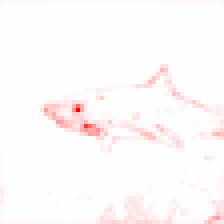}
    \end{minipage}\hspace{0.002\linewidth}%
    \begin{minipage}[c]{0.10\linewidth}
        \centering
        \includegraphics[width=\linewidth]{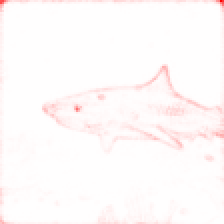}
    \end{minipage}\hspace{0.002\linewidth}%
    \begin{minipage}[c]{0.10\linewidth}
        \centering
        \includegraphics[width=\linewidth]{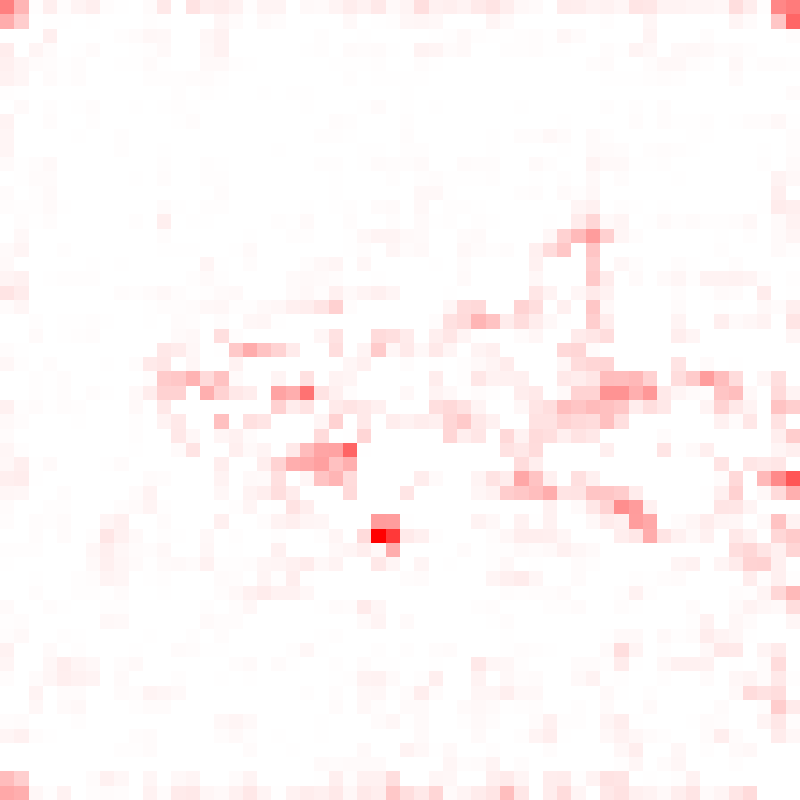}
    \end{minipage}\hspace{0.002\linewidth}%
    \begin{minipage}[c]{0.10\linewidth}
        \centering
        \includegraphics[width=\linewidth]{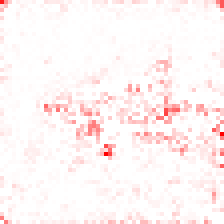}
    \end{minipage}\hspace{0.002\linewidth}%
    \begin{minipage}[c]{0.10\linewidth}
        \centering
        \includegraphics[width=\linewidth]{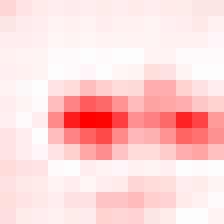}
    \end{minipage}\hspace{0.002\linewidth}%
    \begin{minipage}[c]{0.10\linewidth}
        \centering
        \includegraphics[width=\linewidth]{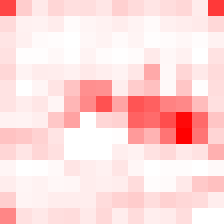}
    \end{minipage}\hspace{0.002\linewidth}%
    \begin{minipage}[c]{0.10\linewidth}
        \centering
        \includegraphics[width=\linewidth]{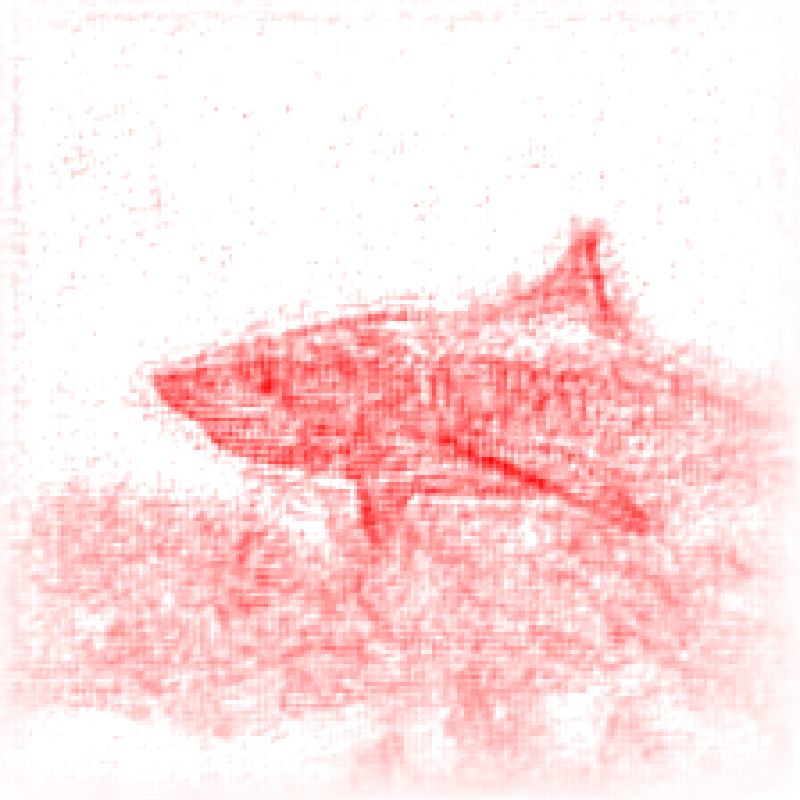}
    \end{minipage}%

    \vspace{0.3cm}

    \begin{minipage}[c]{0.08\linewidth}
        \centering
        \small goldfinch\\[1pt]
        {\scriptsize (1.0000)}
    \end{minipage}\hspace{0.002\linewidth}%
    \begin{minipage}[c]{0.10\linewidth}
        \centering
        \includegraphics[width=\linewidth]{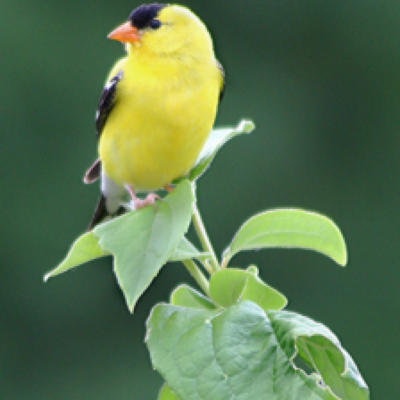}
    \end{minipage}\hspace{0.002\linewidth}%
    \begin{minipage}[c]{0.10\linewidth}
        \centering
        \includegraphics[width=\linewidth]{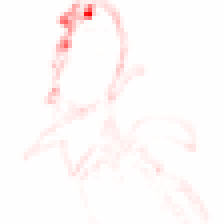}
    \end{minipage}\hspace{0.002\linewidth}%
    \begin{minipage}[c]{0.10\linewidth}
        \centering
        \includegraphics[width=\linewidth]{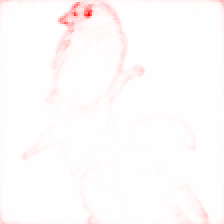}
    \end{minipage}\hspace{0.002\linewidth}%
    \begin{minipage}[c]{0.10\linewidth}
        \centering
        \includegraphics[width=\linewidth]{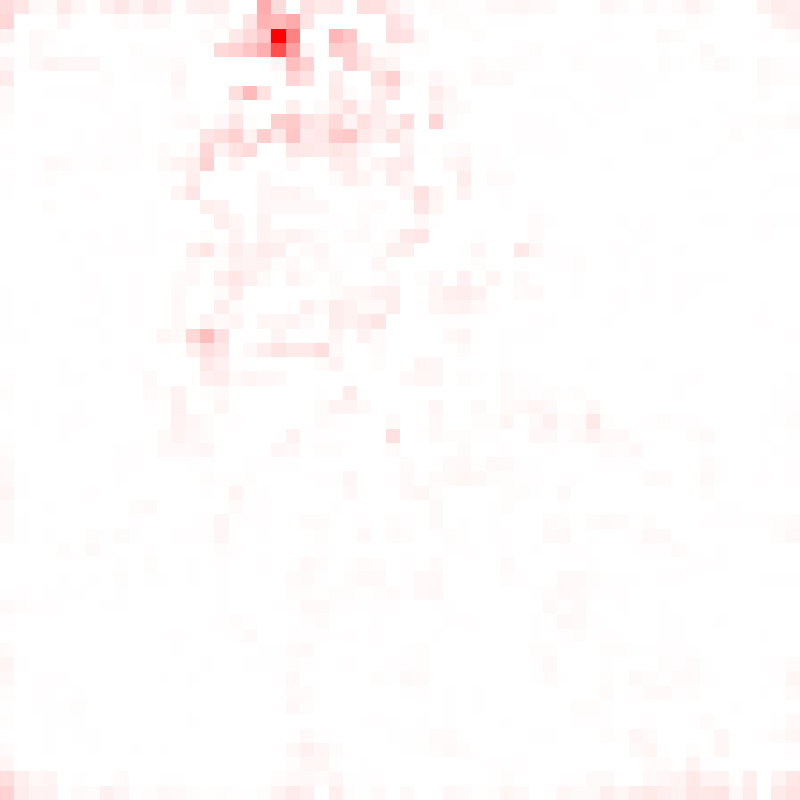}
    \end{minipage}\hspace{0.002\linewidth}%
    \begin{minipage}[c]{0.10\linewidth}
        \centering
        \includegraphics[width=\linewidth]{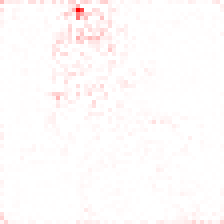}
    \end{minipage}\hspace{0.002\linewidth}%
    \begin{minipage}[c]{0.10\linewidth}
        \centering
        \includegraphics[width=\linewidth]{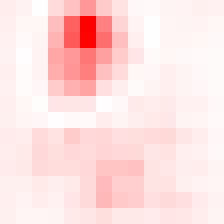}
    \end{minipage}\hspace{0.002\linewidth}%
    \begin{minipage}[c]{0.10\linewidth}
        \centering
        \includegraphics[width=\linewidth]{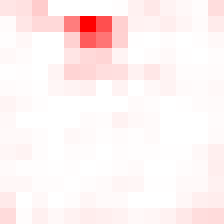}
    \end{minipage}\hspace{0.002\linewidth}%
    \begin{minipage}[c]{0.10\linewidth}
        \centering
        \includegraphics[width=\linewidth]{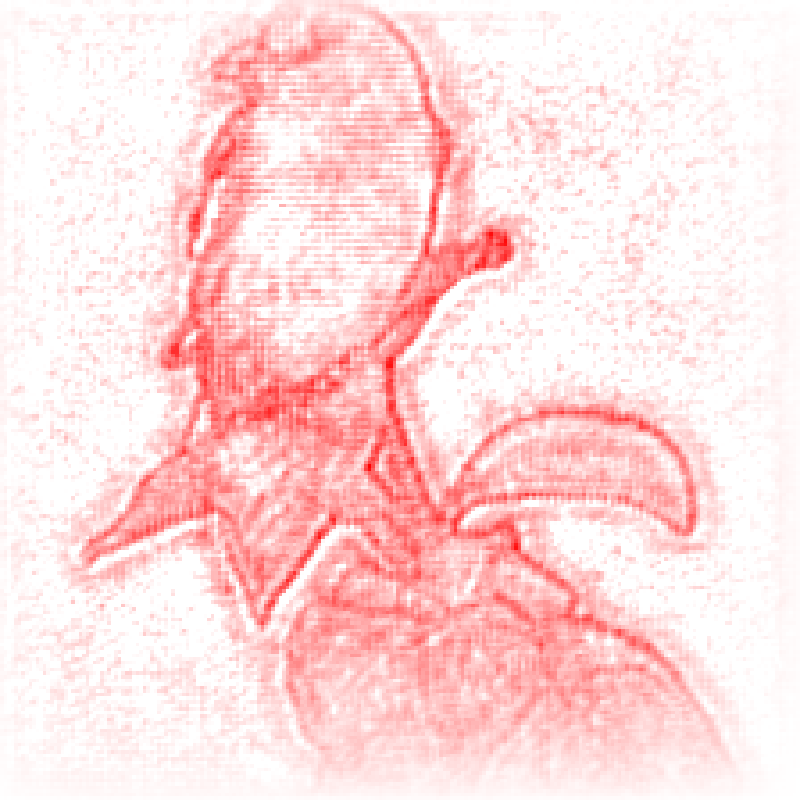}
    \end{minipage}%

    \vspace{0.3cm}

    \begin{minipage}[c]{0.08\linewidth}
        \centering
        \small tree frog\\[1pt]
        {\scriptsize (0.9954)}
    \end{minipage}\hspace{0.002\linewidth}%
    \begin{minipage}[c]{0.10\linewidth}
        \centering
        \includegraphics[width=\linewidth]{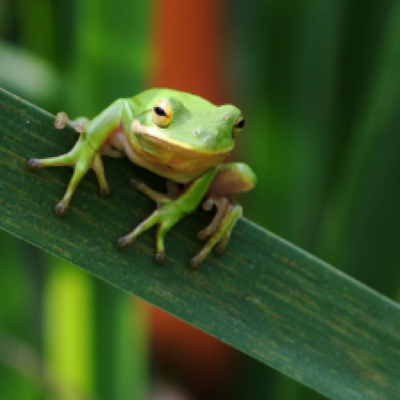}
    \end{minipage}\hspace{0.002\linewidth}%
    \begin{minipage}[c]{0.10\linewidth}
        \centering
        \includegraphics[width=\linewidth]{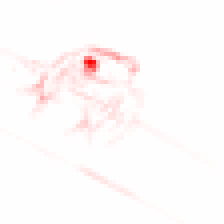}
    \end{minipage}\hspace{0.002\linewidth}%
    \begin{minipage}[c]{0.10\linewidth}
        \centering
        \includegraphics[width=\linewidth]{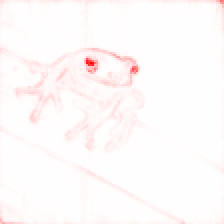}
    \end{minipage}\hspace{0.002\linewidth}%
    \begin{minipage}[c]{0.10\linewidth}
        \centering
        \includegraphics[width=\linewidth]{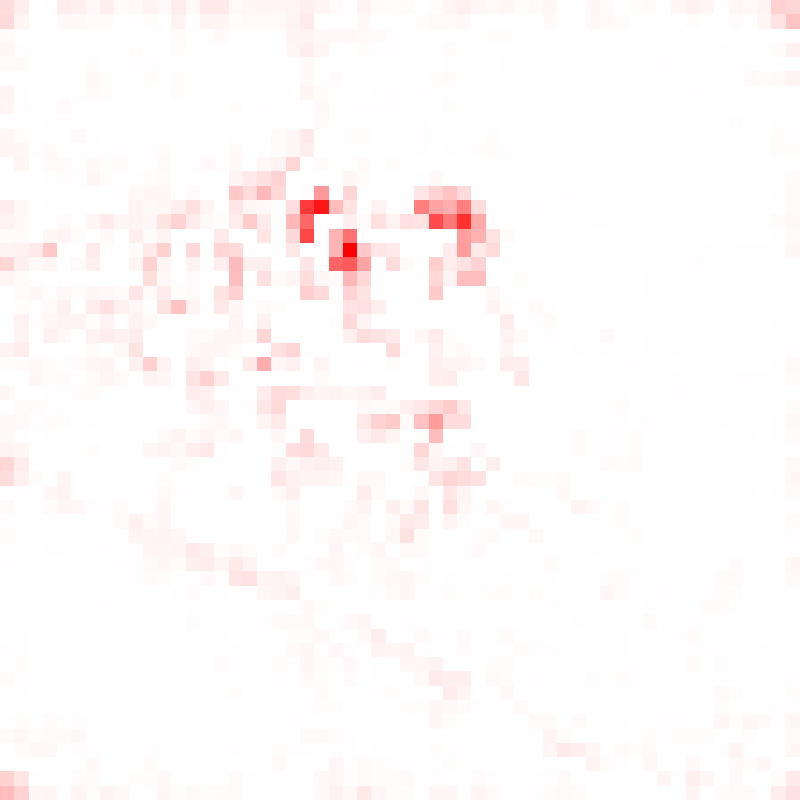}
    \end{minipage}\hspace{0.002\linewidth}%
    \begin{minipage}[c]{0.10\linewidth}
        \centering
        \includegraphics[width=\linewidth]{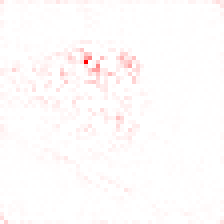}
    \end{minipage}\hspace{0.002\linewidth}%
    \begin{minipage}[c]{0.10\linewidth}
        \centering
        \includegraphics[width=\linewidth]{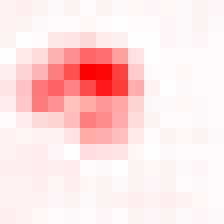}
    \end{minipage}\hspace{0.002\linewidth}%
    \begin{minipage}[c]{0.10\linewidth}
        \centering
        \includegraphics[width=\linewidth]{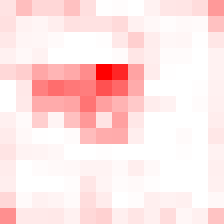}
    \end{minipage}\hspace{0.002\linewidth}%
    \begin{minipage}[c]{0.10\linewidth}
        \centering
        \includegraphics[width=\linewidth]{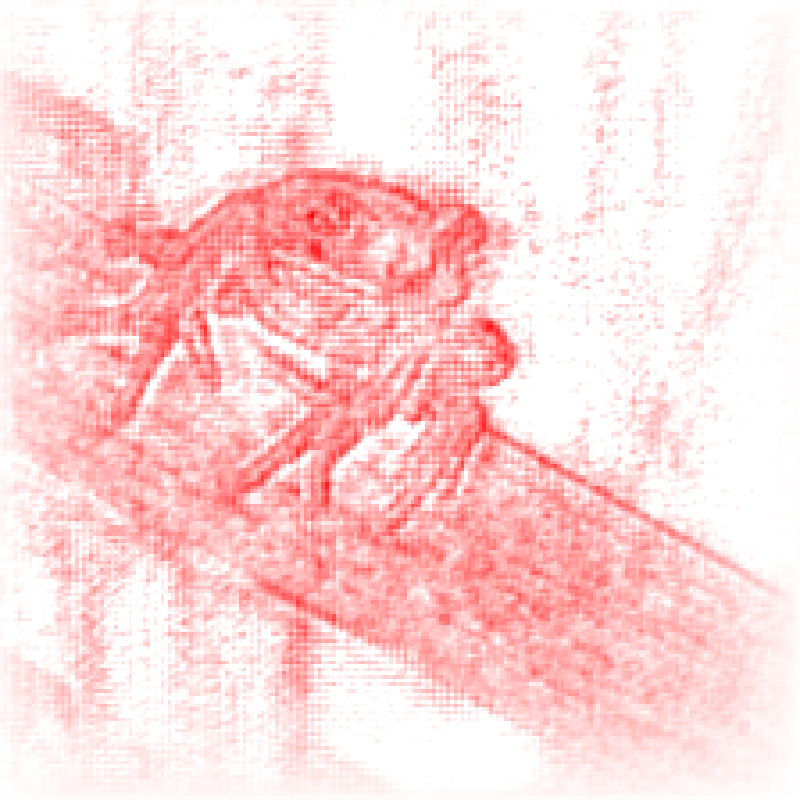}
    \end{minipage}%

    \vspace{0.3cm}

    \begin{minipage}[c]{0.08\linewidth}
        \centering
        \small ladybug\\[1pt]
        {\scriptsize (0.8272)}
    \end{minipage}\hspace{0.002\linewidth}%
    \begin{minipage}[c]{0.10\linewidth}
        \centering
        \includegraphics[width=\linewidth]{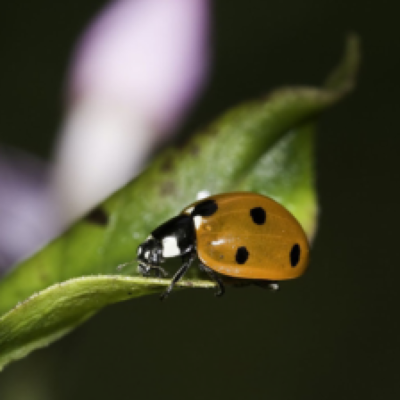}
    \end{minipage}\hspace{0.002\linewidth}%
    \begin{minipage}[c]{0.10\linewidth}
        \centering
        \includegraphics[width=\linewidth]{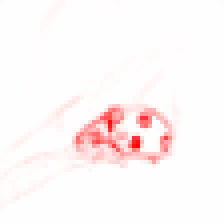}
    \end{minipage}\hspace{0.002\linewidth}%
    \begin{minipage}[c]{0.10\linewidth}
        \centering
        \includegraphics[width=\linewidth]{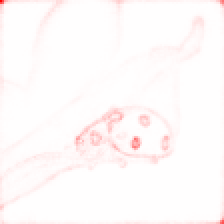}
    \end{minipage}\hspace{0.002\linewidth}%
    \begin{minipage}[c]{0.10\linewidth}
        \centering
        \includegraphics[width=\linewidth]{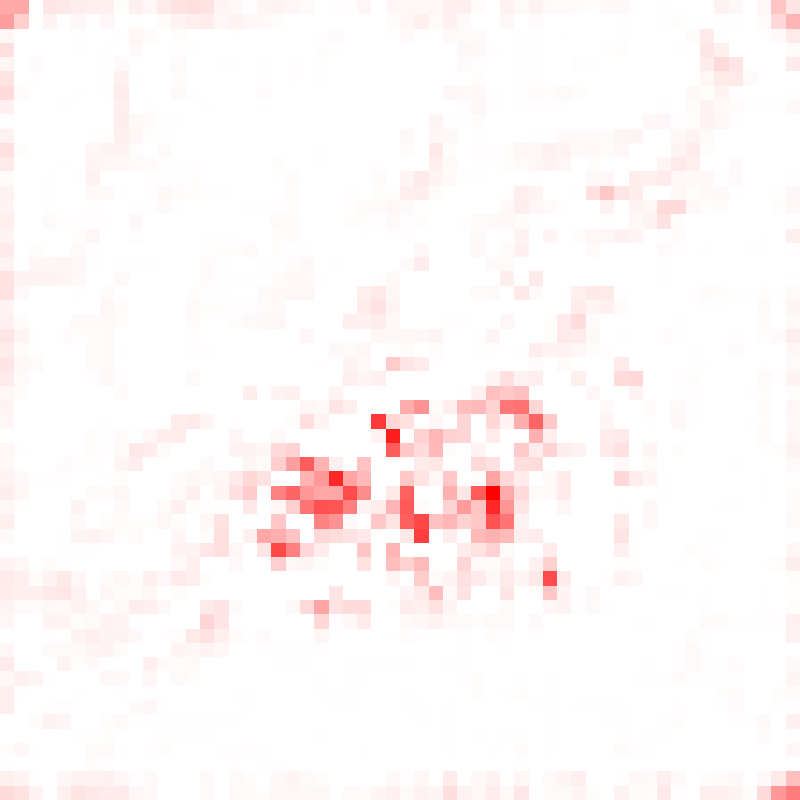}
    \end{minipage}\hspace{0.002\linewidth}%
    \begin{minipage}[c]{0.10\linewidth}
        \centering
        \includegraphics[width=\linewidth]{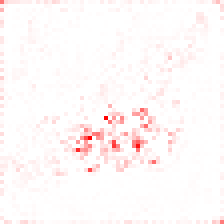}
    \end{minipage}\hspace{0.002\linewidth}%
    \begin{minipage}[c]{0.10\linewidth}
        \centering
        \includegraphics[width=\linewidth]{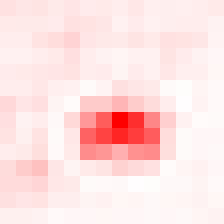}
    \end{minipage}\hspace{0.002\linewidth}%
    \begin{minipage}[c]{0.10\linewidth}
        \centering
        \includegraphics[width=\linewidth]{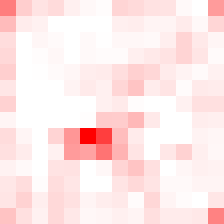}
    \end{minipage}\hspace{0.002\linewidth}%
    \begin{minipage}[c]{0.10\linewidth}
        \centering
        \includegraphics[width=\linewidth]{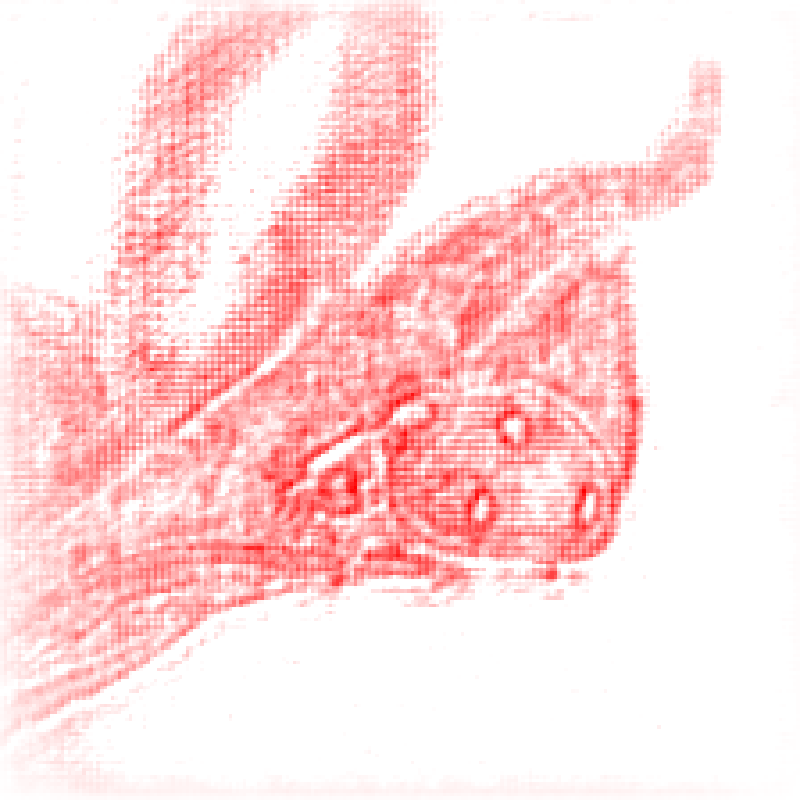}
    \end{minipage}%

    \vspace{0.3cm}

    \begin{minipage}[c]{0.08\linewidth}
        \centering
        \small wood rabbit\\[1pt]
        {\scriptsize (0.9782)}
    \end{minipage}\hspace{0.002\linewidth}%
    \begin{minipage}[c]{0.10\linewidth}
        \centering
        \includegraphics[width=\linewidth]{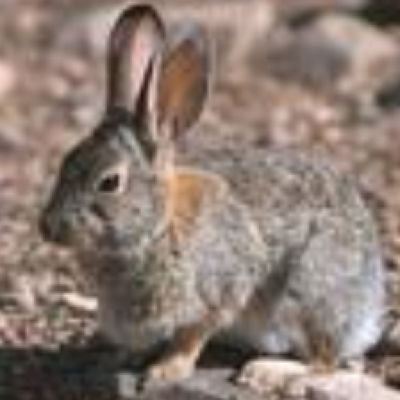}
    \end{minipage}\hspace{0.002\linewidth}%
    \begin{minipage}[c]{0.10\linewidth}
        \centering
        \includegraphics[width=\linewidth]{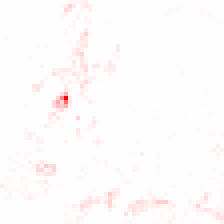}
    \end{minipage}\hspace{0.002\linewidth}%
    \begin{minipage}[c]{0.10\linewidth}
        \centering
        \includegraphics[width=\linewidth]{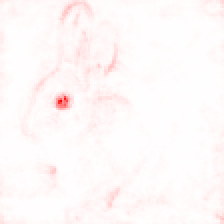}
    \end{minipage}\hspace{0.002\linewidth}%
    \begin{minipage}[c]{0.10\linewidth}
        \centering
        \includegraphics[width=\linewidth]{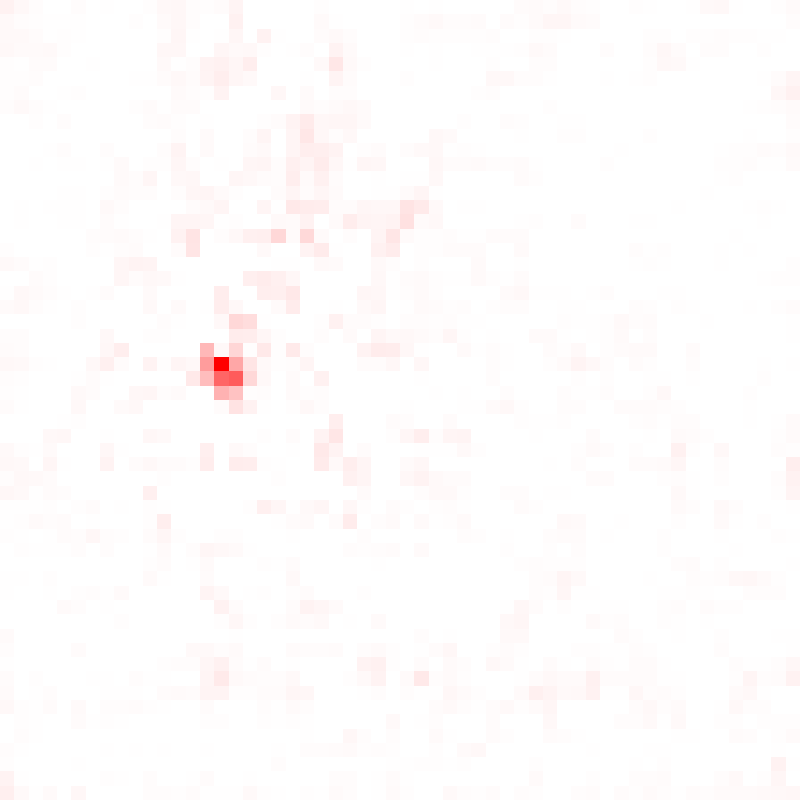}
    \end{minipage}\hspace{0.002\linewidth}%
    \begin{minipage}[c]{0.10\linewidth}
        \centering
        \includegraphics[width=\linewidth]{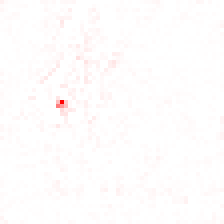}
    \end{minipage}\hspace{0.002\linewidth}%
    \begin{minipage}[c]{0.10\linewidth}
        \centering
        \includegraphics[width=\linewidth]{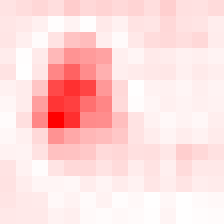}
    \end{minipage}\hspace{0.002\linewidth}%
    \begin{minipage}[c]{0.10\linewidth}
        \centering
        \includegraphics[width=\linewidth]{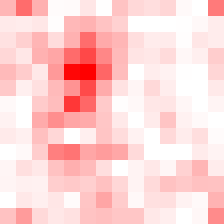}
    \end{minipage}\hspace{0.002\linewidth}%
    \begin{minipage}[c]{0.10\linewidth}
        \centering
        \includegraphics[width=\linewidth]{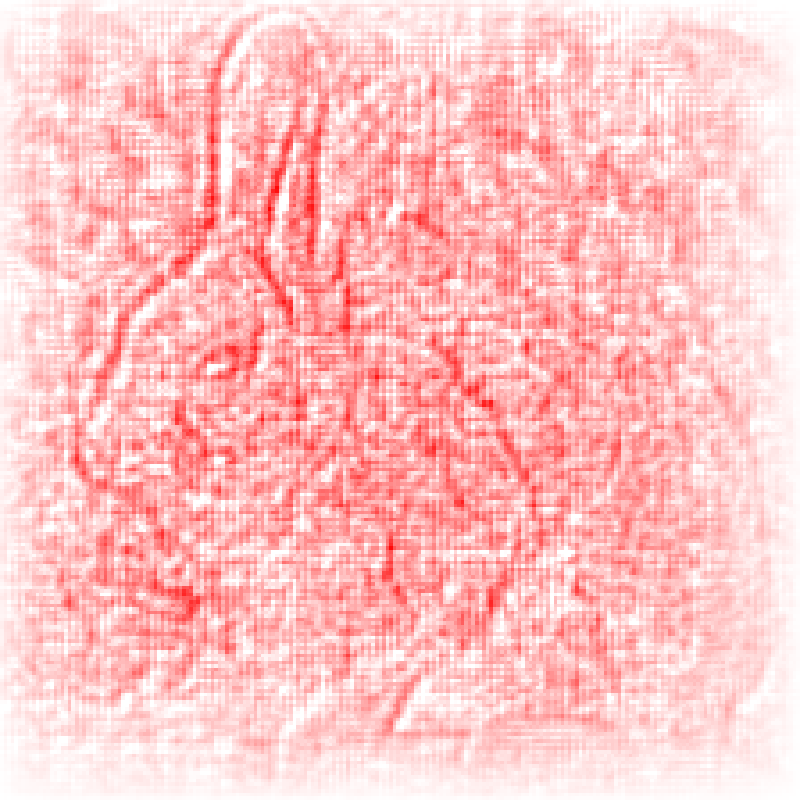}
    \end{minipage}%

    \vspace{0.3cm}

    \begin{minipage}[c]{0.08\linewidth}
        \centering
        \small water tower\\[1pt]
        {\scriptsize (1.0000)}
    \end{minipage}\hspace{0.002\linewidth}%
    \begin{minipage}[c]{0.10\linewidth}
        \centering
        \includegraphics[width=\linewidth]{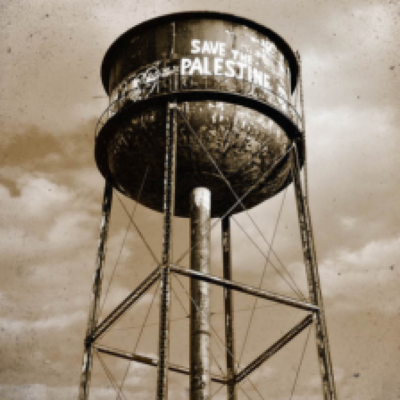}
    \end{minipage}\hspace{0.002\linewidth}%
    \begin{minipage}[c]{0.10\linewidth}
        \centering
        \includegraphics[width=\linewidth]{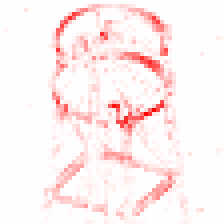}
    \end{minipage}\hspace{0.002\linewidth}%
    \begin{minipage}[c]{0.10\linewidth}
        \centering
        \includegraphics[width=\linewidth]{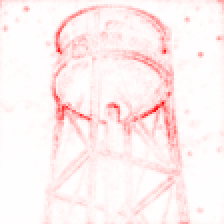}
    \end{minipage}\hspace{0.002\linewidth}%
    \begin{minipage}[c]{0.10\linewidth}
        \centering
        \includegraphics[width=\linewidth]{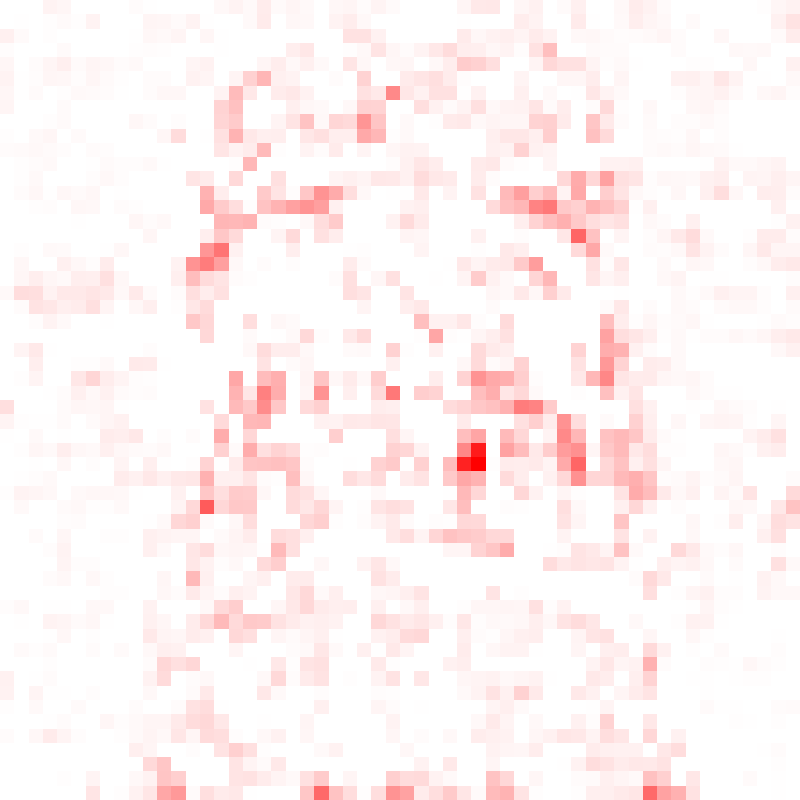}
    \end{minipage}\hspace{0.002\linewidth}%
    \begin{minipage}[c]{0.10\linewidth}
        \centering
        \includegraphics[width=\linewidth]{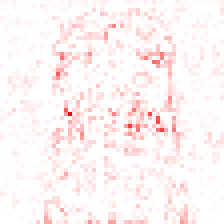}
    \end{minipage}\hspace{0.002\linewidth}%
    \begin{minipage}[c]{0.10\linewidth}
        \centering
        \includegraphics[width=\linewidth]{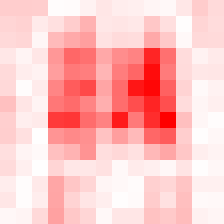}
    \end{minipage}\hspace{0.002\linewidth}%
    \begin{minipage}[c]{0.10\linewidth}
        \centering
        \includegraphics[width=\linewidth]{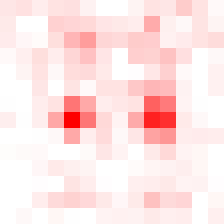}
    \end{minipage}\hspace{0.002\linewidth}%
    \begin{minipage}[c]{0.10\linewidth}
        \centering
        \includegraphics[width=\linewidth]{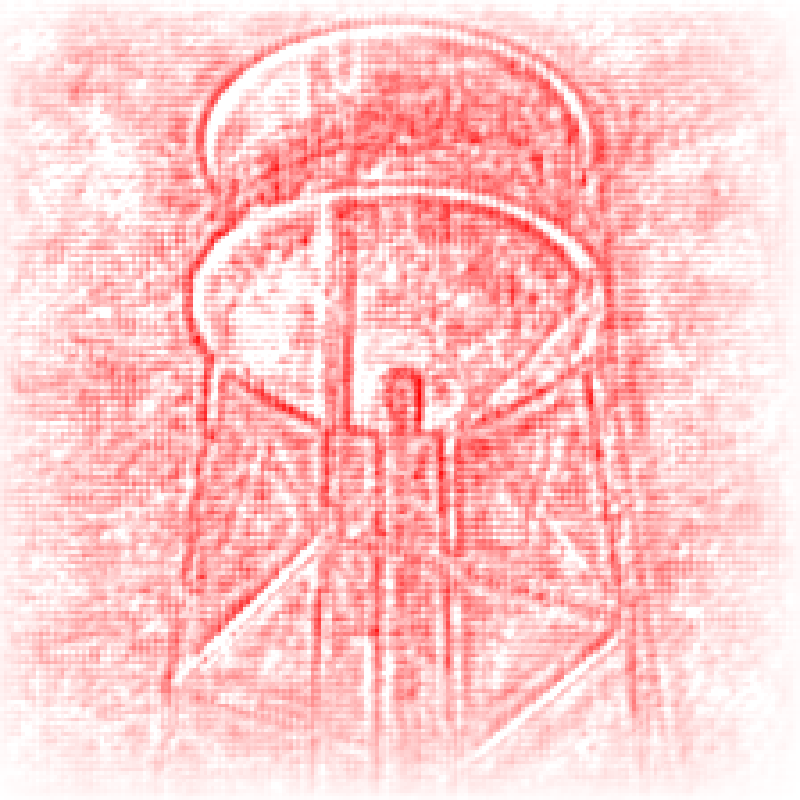}
    \end{minipage}%

    \vspace{0.3cm}

    \begin{minipage}[c]{0.08\linewidth}
        \centering
        \small yawl\\[1pt]
        {\scriptsize (0.9953)}
    \end{minipage}\hspace{0.002\linewidth}%
    \begin{minipage}[c]{0.10\linewidth}
        \centering
        \includegraphics[width=\linewidth]{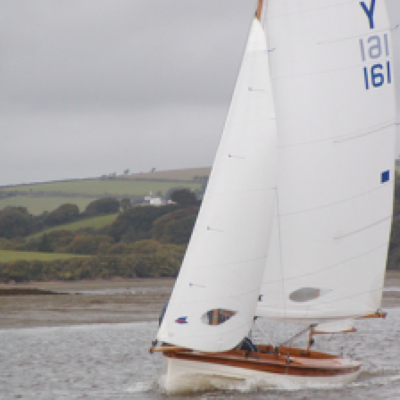}
    \end{minipage}\hspace{0.002\linewidth}%
    \begin{minipage}[c]{0.10\linewidth}
        \centering
        \includegraphics[width=\linewidth]{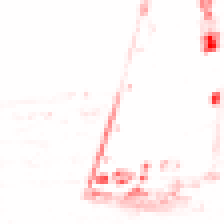}
    \end{minipage}\hspace{0.002\linewidth}%
    \begin{minipage}[c]{0.10\linewidth}
        \centering
        \includegraphics[width=\linewidth]{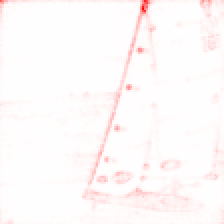}
    \end{minipage}\hspace{0.002\linewidth}%
    \begin{minipage}[c]{0.10\linewidth}
        \centering
        \includegraphics[width=\linewidth]{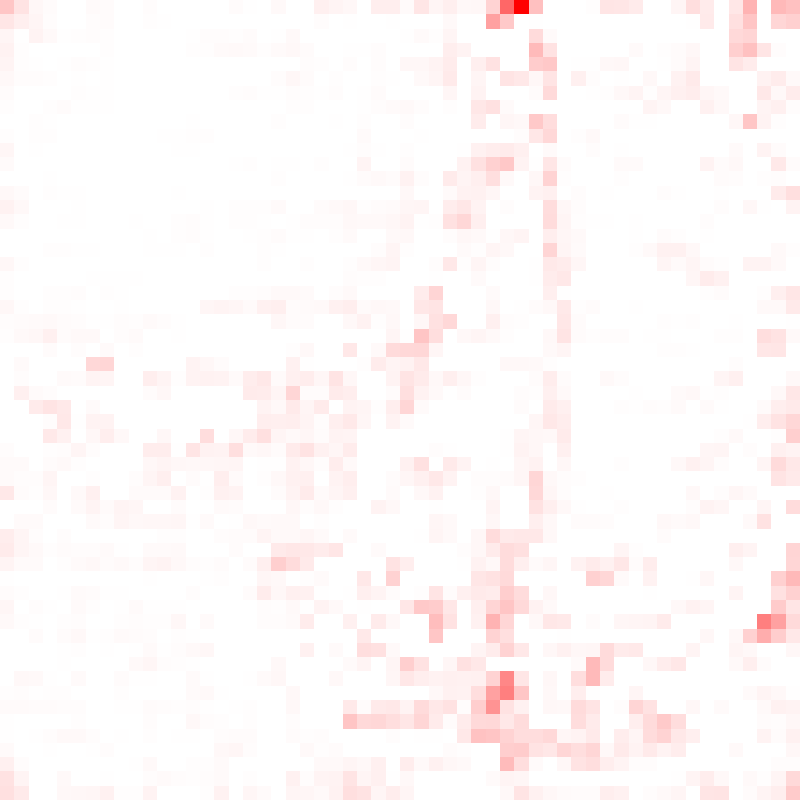}
    \end{minipage}\hspace{0.002\linewidth}%
    \begin{minipage}[c]{0.10\linewidth}
        \centering
        \includegraphics[width=\linewidth]{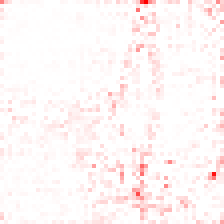}
    \end{minipage}\hspace{0.002\linewidth}%
    \begin{minipage}[c]{0.10\linewidth}
        \centering
        \includegraphics[width=\linewidth]{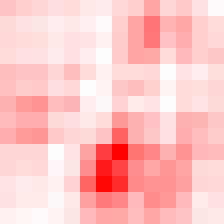}
    \end{minipage}\hspace{0.002\linewidth}%
    \begin{minipage}[c]{0.10\linewidth}
        \centering
        \includegraphics[width=\linewidth]{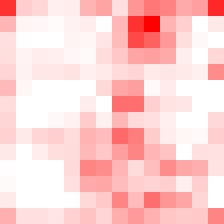}
    \end{minipage}\hspace{0.002\linewidth}%
    \begin{minipage}[c]{0.10\linewidth}
        \centering
        \includegraphics[width=\linewidth]{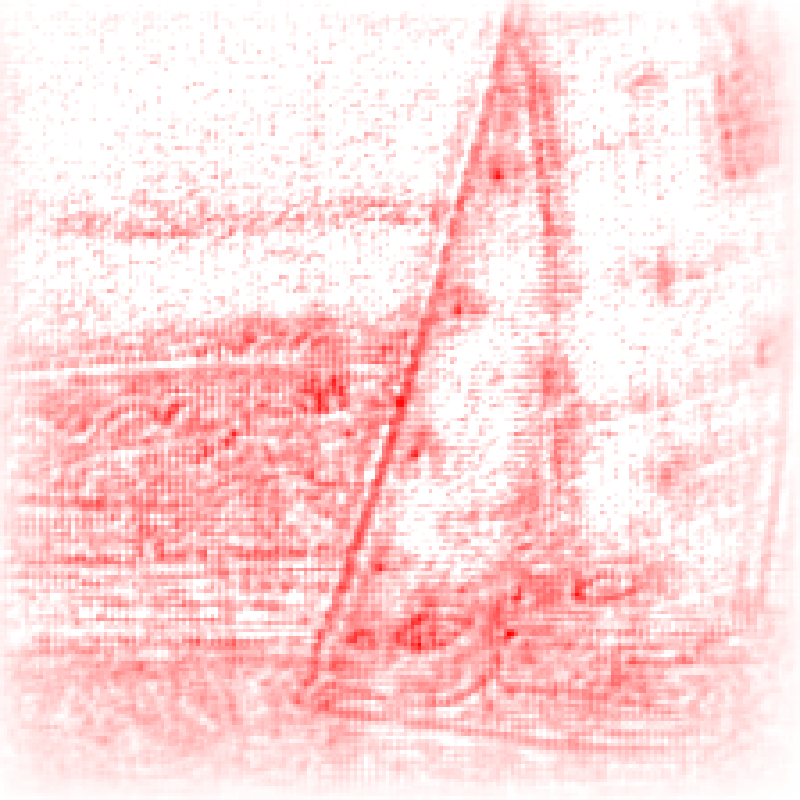}
    \end{minipage}%

    \caption{Saliency map comparison for VGG16 on images from the custom ImageNet-S test subset. Each row shows the predicted class (with confidence), original image, and saliency maps from different methods. Methods include SplitLRP, classical LRP, SplitCAM variants, LayerCAM, and SplitGrad.}
    \label{fig:vgg_saliency_comparison}
\end{figure}

%% file: appendix/table_appendix_generated_vgg_full_eval_sorted.tex
\begin{longtable}{p{4cm}|cc|cccccc}
    \caption{Comprehensive XAI Evaluation Results (VGG16). The abbreviations are similar to \cref{tab:eval_vgg_main}. 
    For each method configuration we highlight the best metrics values across the considered layer configurations in bold.}
    \label{tab:xai_eval_comprehensive_vgg} \\
    \toprule
    \textbf{Method} & \textbf{Layer} & \textbf{Abs} & \textbf{Select.} & \textbf{Attr. Loc.} & \textbf{Point.} & \textbf{Pixel Fl.} & \textbf{Pixel Fl.} & \textbf{Max Sens.} \\
     & & & \textbf{$\downarrow$} & \textbf{$\uparrow$} & \textbf{$\uparrow$} & \textbf{AUC@5 $\uparrow$} & \textbf{AUC@20 $\uparrow$} & \textbf{$\downarrow$} \\
    \midrule
    \endfirsthead
    
    \multicolumn{9}{c}%
    {{\tablename\ \thetable{} -- continued from previous page}} \\
    \toprule
    \textbf{Method} & \textbf{Layer} & \textbf{Abs} & \textbf{Select.} & \textbf{Attr. Loc.} & \textbf{Point.} & \textbf{Pixel Fl.} & \textbf{Pixel Fl.} & \textbf{Max Sens.} \\
     & & & \textbf{$\downarrow$} & \textbf{$\uparrow$} & \textbf{$\uparrow$} & \textbf{AUC@5 $\uparrow$} & \textbf{AUC@20 $\uparrow$} & \textbf{$\downarrow$} \\
    \midrule
    \endhead
    
    \midrule
    \multicolumn{9}{r}{{Continued on next page}} \\
    \midrule
    \endfoot
    
    \bottomrule
    \endlastfoot
    
    \multicolumn{9}{c}{\textit{SplitCAM Methods}} \\
    \midrule
    SplitCAM (sc, $\alpha \ 0.4$) & (12, g) & n & 4.720 & \fboxsep=0.5pt\colorbox{red!20}{\textbf{0.651}} & \textbf{0.912} & 0.213 & 1.794 & \textbf{0.389} \\
    SplitCAM (sc, $\alpha \ 0.4$) & (14, g) & n & \fboxsep=0.5pt\colorbox{red!20}{\textbf{2.727}} & 0.563 & 0.836 & \textbf{0.300} & \textbf{2.404} & 1.012 \\
    \cmidrule(lr){1-9}
    SplitCAM (sc, $\alpha \ 0.4$, wta) & (14, g) & n & \textbf{2.761} & 0.543 & 0.827 & \textbf{0.303} & \textbf{2.414} & 1.056 \\
    SplitCAM (sc, $\alpha \ 0.4$, wta) & (26, +, g) & y & 4.711 & \textbf{0.635} & \fboxsep=0.5pt\colorbox{red!20}{\textbf{0.938}} & 0.165 & 1.590 & \textbf{0.456} \\
    \cmidrule(lr){1-9}
    SplitCAM (sc, $\alpha \ 0.5$, wta) & (14, g) & n & \textbf{2.766} & 0.536 & 0.823 & \textbf{0.303} & \textbf{2.416} & 1.056 \\
    SplitCAM (sc, $\alpha \ 0.5$, wta) & (26, +, g) & n & 4.608 & \textbf{0.607} & \textbf{0.903} & 0.185 & 1.684 & \textbf{0.566} \\
    \midrule
    \multicolumn{9}{c}{\textit{SplitGrad Methods}} \\
    \midrule
    SplitGrad ($\alpha \ 0.4$) & (12, +, g) & n & 5.633 & \textbf{0.613} & \textbf{0.873} & 0.148 & 1.461 & \textbf{0.364} \\
    SplitGrad ($\alpha \ 0.4$) & (14, +, g) & y & \textbf{3.429} & 0.533 & 0.820 & \textbf{0.235} & \textbf{2.119} & 0.851 \\
    \cmidrule(lr){1-9}
    SplitGrad ($\alpha \ 0.4$, wta) & (12, +, g) & n & 4.735 & \textbf{0.635} & \textbf{0.883} & 0.186 & 1.756 & 0.651 \\
    SplitGrad ($\alpha \ 0.4$, wta) & (14, +, g) & y & \textbf{3.287} & 0.540 & 0.818 & \textbf{0.249} & \textbf{2.186} & 0.857 \\
    \cmidrule(lr){1-9}
    SplitGrad ($\alpha \ 0.5$, wta) & (12, +, g) & y & \textbf{3.379} & \textbf{0.551} & 0.859 & \textbf{0.261} & \textbf{2.220} & 0.869 \\
    SplitGrad ($\alpha \ 0.5$, wta) & (26, +, g) & y & 4.948 & 0.543 & \textbf{0.899} & 0.162 & 1.527 & \textbf{0.617} \\
    \midrule
    \multicolumn{9}{c}{\textit{SplitLRP Methods}} \\
    \midrule
    SplitLRP (sc) & (7, pos) & n & \textbf{3.062} & 0.092 & \textbf{0.857} & \textbf{0.296} & \textbf{2.290} & 0.865 \\
    SplitLRP (sc) & (19, pos) & n & 4.393 & \textbf{0.600} & 0.823 & 0.166 & 1.656 & \textbf{0.291} \\
    \cmidrule(lr){1-9}
    SplitLRP (sc, wta) & (7, pos) & n & \textbf{3.235} & -0.036 & 0.836 & \textbf{0.296} & \textbf{2.291} & 0.866 \\
    SplitLRP (sc, wta) & (26, pos) & y & 5.168 & \textbf{0.588} & \textbf{0.871} & 0.151 & 1.473 & \fboxsep=0.5pt\colorbox{red!20}{\textbf{0.282}} \\
    \midrule
    \multicolumn{9}{c}{\textit{Classical Methods (No Layer Optimization)}} \\
    \midrule
    Deconvolution & 0 & y & 5.435 & 0.468 & 0.629 & 0.204 & 1.715 & 0.956 \\
    DeepLift & 0 & y & 4.447 & 0.551 & 0.726 & 0.281 & 1.945 & 1.053 \\
    Feature Ablation & 0 & y & 6.029 & 0.479 & 0.624 & 0.123 & 1.250 & 0.859 \\
    Gradient SHAP & 0 & y & 4.808 & 0.552 & 0.783 & 0.261 & 1.862 & 1.142 \\
    Integrated Gradients & 0 & y & 4.838 & 0.554 & 0.797 & 0.262 & 1.851 & 1.069 \\
    Occlusion & 0 & y & 5.214 & 0.499 & 0.716 & 0.130 & 1.217 & 0.699 \\
    GradCAM++ & 30 & n & 5.955 & 0.564 & 0.827 & 0.130 & 1.245 & 0.401 \\
    \midrule
    \multicolumn{9}{c}{\textit{Classical Methods (Layer Optimization)}} \\
    \midrule
    Guided Backprop & 2 & n & \textbf{3.276} & \textbf{0.632} & 0.883 & \textbf{0.281} & \textbf{2.302} & 0.571 \\
    Guided Backprop & 5 & n & 3.448 & 0.614 & \textbf{0.887} & 0.240 & 2.091 & \textbf{0.375} \\
    \cmidrule(lr){1-9}
    Guided GradCAM & 0 & n & \textbf{4.301} & \textbf{-0.053} & \textbf{0.857} & 0.161 & \textbf{1.459} & 1.047 \\
    Guided GradCAM & 26 & n & 5.130 & -2.635 & 0.843 & \textbf{0.179} & 1.415 & \textbf{0.947} \\
    \cmidrule(lr){1-9}
    GradCAM & 0 & n & 6.006 & \textbf{0.356} & \textbf{0.638} & 0.060 & 0.899 & 1.123 \\
    GradCAM & 14 & n & \textbf{5.555} & 0.306 & 0.580 & \textbf{0.136} & \textbf{1.525} & \textbf{0.903} \\
    \cmidrule(lr){1-9}
    LayerCAM & 14 & - & 2.792 & 0.576 & 0.798 & \fboxsep=0.5pt\colorbox{red!20}{\textbf{0.317}} & \fboxsep=0.5pt\colorbox{red!20}{\textbf{2.474}} & \textbf{2.656} \\
    LayerCAM & 2 & - & \textbf{3.128} & \textbf{0.580} & \textbf{0.855} & 0.300 & 2.191 & 5.508 \\
    \cmidrule(lr){1-9}
    LRP $\gamma=0.25$ & 0 & n & \textbf{3.189} & \textbf{0.615} & \textbf{0.680} & 0.218 & 1.918 & 0.519 \\
    LRP $\gamma=0.25$ & 14 & n & 3.593 & 0.577 & 0.611 & \textbf{0.279} & \textbf{2.255} & \textbf{0.491} \\
    \midrule
    \multicolumn{9}{c}{\textit{Baseline Methods}} \\
    \midrule
    Baseline: Binary Mask & 0 & y & 6.011 & 1.000 & 1.000 & 0.091 & 1.054 & -- \\
    Baseline: Original Image & 0 & y & 8.770 & -1.699 & 0.572 & 0.061 & 0.619 & -- \\
    Random Baseline & 0 & y & 3.570 & 0.425 & 0.458 & 0.203 & 1.585 & -- \\
    \midrule
\end{longtable}

%% file: appendix/table_appendix_generated_res_full_eval_sorted.tex
\begin{longtable}{p{4cm}|cc|cccccc}
    \caption{Comprehensive XAI Evaluation Results (ResNet50). The abbreviations are similar to \cref{tab:eval_res_main}. For each method configuration we highlight the best metrics values across the considered layer configurations in bold.}
    \label{tab:xai_eval_comprehensive_res} \\
    \toprule
    \textbf{Method} & \textbf{Layer} & \textbf{Abs} & \textbf{Select.} & \textbf{Attr. Loc.} & \textbf{Point.} & \textbf{Pixel Fl.} & \textbf{Pixel Fl.} & \textbf{Max Sens.} \\
     & & & \textbf{$\downarrow$} & \textbf{$\uparrow$} & \textbf{$\uparrow$} & \textbf{AUC@5 $\uparrow$} & \textbf{AUC@20 $\uparrow$} & \textbf{$\downarrow$} \\
    \midrule
    \endfirsthead
    
    \multicolumn{9}{c}%
    {{\tablename\ \thetable{} -- continued from previous page}} \\
    \toprule
    \textbf{Method} & \textbf{Layer} & \textbf{Abs} & \textbf{Select.} & \textbf{Attr. Loc.} & \textbf{Point.} & \textbf{Pixel Fl.} & \textbf{Pixel Fl.} & \textbf{Max Sens.} \\
     & & & \textbf{$\downarrow$} & \textbf{$\uparrow$} & \textbf{$\uparrow$} & \textbf{AUC@5 $\uparrow$} & \textbf{AUC@20 $\uparrow$} & \textbf{$\downarrow$} \\
    \midrule
    \endhead
    
    \midrule
    \multicolumn{9}{r}{{Continued on next page}} \\
    \midrule
    \endfoot
    
    \bottomrule
    \endlastfoot
    
    \multicolumn{9}{c}{\textit{SplitCAM Methods}} \\
    \midrule
    SplitCAM (sc, $\alpha\ 0.3$, wta) & (1.0.2, g) & n & \fboxsep=0.5pt\colorbox{red!20}{\textbf{3.503}} & 0.487 & 0.747 & \textbf{0.168} & \textbf{1.395} & 1.277 \\
    SplitCAM (sc, $\alpha\ 0.3$, wta) & (4.1.2, g) & n & 5.654 & \fboxsep=0.5pt\colorbox{red!20}{\textbf{0.673}} & \fboxsep=0.5pt\colorbox{red!20}{\textbf{0.972}} & 0.081 & 0.880 & \textbf{0.233} \\

    \cmidrule(lr){1-9}

    SplitCAM (sc, $\alpha\ 0.5$, wta) & (2.0.1, g) & n & \textbf{3.682} & -1.370 & 0.887 & \textbf{0.158} & \textbf{1.447} & 1.078 \\
    SplitCAM (sc, $\alpha\ 0.5$, wta) & (4.1.1, +, g) & y & 6.526 & \textbf{0.611} & \textbf{0.956} & 0.066 & 0.737 & \textbf{0.187} \\

    \midrule
    \multicolumn{9}{c}{\textit{SplitGrad Methods}} \\
    \midrule
    SplitGrad ($\alpha\ 0.3$, wta) & (1.0.2, g) & y & \textbf{4.449} & 0.513 & 0.777 & \textbf{0.149} & \textbf{1.266} & 0.813 \\
    SplitGrad ($\alpha\ 0.3$, wta) & (4.1.2, g) & y & 5.663 & \textbf{0.649} & \textbf{0.972} & 0.082 & 0.882 & \textbf{0.219} \\

    \cmidrule(lr){1-9}

    SplitGrad ($\alpha\ 0.5$, wta) & (3.1.1, g) & y & \textbf{3.588} & 0.559 & 0.880 & \textbf{0.130} & \textbf{1.366} & 0.859 \\
    SplitGrad ($\alpha\ 0.5$, wta) & (4.1.2, g) & y & 5.690 & \textbf{0.654} & \textbf{0.972} & 0.083 & 0.886 & \textbf{0.230} \\

    \midrule
    \multicolumn{9}{c}{\textit{SplitLRP Methods}} \\
    \midrule
    SplitLRP (sc, wta) & (conv1, pos) & y & \textbf{6.313} & \textbf{0.506} & \textbf{0.587} & \textbf{0.056} & \textbf{0.697} & 0.755 \\
    SplitLRP (sc, wta) & (1.0.2, pos) & y & 6.785 & 0.483 & 0.567 & 0.050 & 0.686 & \textbf{0.696} \\
    \midrule
    \multicolumn{9}{c}{\textit{Classical Methods (No Layer Optimization)}} \\
    \midrule
    Deconvolution & conv1 & y & 5.474 & 0.478 & 0.666 & 0.151 & 1.177 & 0.618 \\
    Gradient SHAP & conv1 & y & 5.269 & 0.527 & 0.779 & 0.166 & 1.227 & 1.074 \\
    GradCAM++ & 4.1.2 & n & 6.054 & 0.556 & 0.945 & 0.077 & 0.838 & 0.140 \\
    Integrated Gradients & conv1 & y & 5.265 & 0.528 & 0.777 & 0.167 & 1.226 & 0.994 \\
    Feature Ablation & conv1 & n & 4.655 & 0.136 & 0.684 & 0.188 & 1.388 & 1.077 \\
    Occlusion & conv1 & n & 5.300 & 0.179 & 0.726 & 0.130 & 1.030 & 0.785 \\
    Deconvolution & conv1 & n & 3.934 & 0.484 & 0.666 & 0.147 & 1.203 & 0.751 \\
    \midrule
    \multicolumn{9}{c}{\textit{Classical Methods (Layer Optimization)}} \\
    \midrule
    GradCAM & 2.0.1 & y & \textbf{4.003} & 0.450 & 0.678 & 0.052 & \textbf{1.024} & 0.738 \\
    GradCAM & 4.1.2 & n & 6.070 & \textbf{0.554} & \textbf{0.945} & \textbf{0.076} & 0.829 & \fboxsep=0.5pt\colorbox{red!20}{\textbf{0.131}} \\
    \cmidrule(lr){1-9}
    Guided GradCAM & 4.0.2 & n & 4.929 & \fboxsep=0.5pt\colorbox{red!20}{\textbf{0.742}} & 0.866 & 0.119 & 1.008 & 0.687 \\
    Guided GradCAM & 4.1.2 & n & \textbf{3.843} & 0.717 & \textbf{0.883} & \textbf{0.129} & \textbf{1.096} & \textbf{0.612} \\
    \cmidrule(lr){1-9}
    LayerCAM & 2.0.2 & n & \textbf{3.690} & 0.549 & 0.865 & \textbf{0.144} & \textbf{1.428} & 0.913 \\
    LayerCAM & 4.1.1 & n & 5.661 & \textbf{0.664} & \textbf{0.971} & 0.083 & 0.890 & \textbf{0.227} \\
    \cmidrule(lr){1-9}
    LRP $\gamma=0.25$ & conv1 & y & \textbf{3.550} & \textbf{0.656} & 0.852 & \fboxsep=0.5pt\colorbox{red!20}{\textbf{0.213}} & \textbf{1.737} & 0.521 \\
    LRP $\gamma=0.25$ & 4.1.1 & n & 5.649 & 0.630 & \textbf{0.965} & 0.081 & 0.877 & \textbf{0.200} \\
    \cmidrule(lr){1-9}
    Guided Backprop & conv1 & n & \textbf{3.517} & \textbf{0.621} & 0.867 & \textbf{0.212} & \fboxsep=0.5pt\colorbox{red!20}{\textbf{1.757}} & 0.459 \\
    Guided Backprop & 3.1.2 & n & 3.874 & 0.582 & \textbf{0.938} & 0.121 & 1.307 & \textbf{0.213} \\
    \midrule
    \multicolumn{9}{c}{\textit{Baseline Methods}} \\
    \midrule
    Baseline: Original Image & conv1 & y & 8.351 & -1.699 & 0.572 & 0.049 & 0.435 & -- \\
    Baseline: Binary Mask & conv1 & y & 5.964 & 1.000 & 1.000 & 0.067 & 0.739 & -- \\
    Random Baseline & - & y & 4.071 & 0.425 & 0.440 & 0.147 & 1.067 & -- \\
    \midrule
    \end{longtable}

%% file: appendix/randomization_sanity_check_table.tex
\begin{table}[htbp]
\centering
\caption{Randomization sanity check: Mean values across the custom test subset of ImageNet-S. The abbreviations are similar to \cref{tab:eval_vgg_main}.}
\label{tab:randomization_sanity_check}
\begin{tabular}{l c c c}
\toprule
Method & HOG $\uparrow$ & Spearman $\uparrow$ & SSIM $\uparrow$ \\
\midrule
\multicolumn{4}{l}{\textbf{Baselines}} \\
Deconvolution & 0.985 & 0.999 & 0.481 \\
DeepLift & 0.365 & 0.274 & 0.716 \\
GradientShap & 0.286 & 0.081 & 0.571 \\
input\_x\_gradient & 0.313 & 0.082 & 0.621 \\
integrated\_gradients & 0.303 & 0.102 & 0.584 \\
layer\_cam\_features\_14\_relu & 0.688 & 0.123 & 0.417 \\
layer\_cam\_features\_2\_relu & 0.542 & 0.114 & 0.475 \\
saliency & 0.188 & 0.416 & 0.353 \\

\midrule
\multicolumn{4}{l}{\textbf{SplitCAM}} \\
SplitCAM (sc, $\alpha \ 0.4$) (12, g) & 0.736 & 0.729 & 0.202 \\
SplitCAM (sc, $\alpha \ 0.4$) (14, g) & 0.724 & 0.194 & 0.174 \\
SplitCAM (sc, $\alpha \ 0.4$, wta) (14, g) & 0.724 & 0.389 & 0.303 \\
SplitCAM (sc, $\alpha \ 0.4$, wta) (26, +, g) abs & 0.728 & 0.342 & 0.319 \\
SplitCAM (sc, $\alpha \ 0.5$, wta) (14, g) & 0.728 & 0.096 & 0.292 \\
SplitCAM (sc, $\alpha \ 0.5$, wta) (26, +, g) & 0.804 & 0.186 & 0.148 \\

\midrule
\multicolumn{4}{l}{\textbf{SplitGrad}} \\
SplitGrad ($\alpha \ 0.4$) (12, +, g) & 0.677 & 0.711 & 0.174 \\
SplitGrad ($\alpha \ 0.4$) (14, +, g) abs & 0.704 & 0.197 & 0.091 \\
SplitGrad ($\alpha \ 0.4$, wta) (12, +, g) & 0.706 & 0.454 & 0.174 \\
SplitGrad ($\alpha \ 0.4$, wta) (14, +, g) abs & 0.721 & 0.211 & 0.199 \\
SplitGrad ($\alpha \ 0.5$, wta) (12, +, g) abs & 0.725 & 0.238 & 0.206 \\
SplitGrad ($\alpha \ 0.5$, wta) (26, +, g) abs & 0.795 & 0.149 & 0.306 \\

\midrule
\multicolumn{4}{l}{\textbf{SplitLRP}} \\
SplitLRP (7, pos) & 0.997 & 1.000 & 0.471 \\
SplitLRP (19, pos) & 0.998 & 1.000 & 0.562 \\
SplitLRP (wta) (7, pos) & 0.998 & 1.000 & 0.456 \\
SplitLRP (wta) (26, pos) abs & 0.994 & 0.999 & 0.446 \\
\bottomrule
\end{tabular}
\end{table}

%% file: appendix/table_timing_comparison.tex
\begin{longtable}{l|cc}
    \caption{Mean Computation Time Comparison (seconds)}
    \label{tab:timing_comparison} \\
    \toprule
    \textbf{Method} & \textbf{ResNet18} & \textbf{VGG16} \\
    \midrule
    \endfirsthead

    \multicolumn{3}{c}%
    {{\tablename\ \thetable{} -- continued from previous page}} \\
    \toprule
    \textbf{Method} & \textbf{ResNet18} & \textbf{VGG16} \\
    \midrule
    \endhead

    \midrule
    \multicolumn{3}{r}{{Continued on next page}} \\
    \midrule
    \endfoot

    \bottomrule
    \endlastfoot

    Model Splitting Time & 0.0331 & 0.0295 \\
    \midrule
    \multicolumn{3}{c}{\textit{Split Methods}} \\
    \midrule
    SplitCAM & 0.4368 & 0.3348 \\
    SplitGrad & 0.4284 & 0.3309 \\
    SplitLRP & 0.5711 & 0.3204 \\
    \midrule
    \multicolumn{3}{c}{\textit{Classical Methods}} \\
    \midrule
    GradCAM & 0.0216 & 0.0232 \\
    GradCAM++ & 0.0157 & 0.0155 \\
    LayerCAM & 0.0149 & 0.0124 \\
    Guided GradCAM & 0.3102 & 0.2808 \\
    LRP $\gamma=0.25$ & 0.0599 & 0.0507 \\
    DeepLift & 0.0203 & 0.0210 \\
    Integrated Gradients & 0.0070 & 0.0121 \\
    Gradient SHAP & 0.0084 & 0.0082 \\
    Guided Backprop & 0.0253 & 0.0258 \\
    Deconvolution & 0.0087 & 0.0074 \\
    Saliency & 0.0060 & 0.0047 \\
    Input $\times$ Gradient & 0.0055 & 0.0046 \\
    Feature Ablation & 231.6631 & 192.2602 \\
    Occlusion & 238.2302 & 201.0051 \\
\end{longtable}

%% file: appendix/ablation_studies.tex
\section{Ablation Studies}
\label{app:ablation}

In this appendix, we present ablation studies examining the effect of the proposed stabilization techniques for the forward and backward pass, as well as the influence of the selected max-pooling implementation, see \cref{subsec:other_components}.

\subsection{Backward Pass Stabilization Ablation}

\input{appendix/table_ppendix_alpha_combined.tex}

In this subsection, we present the full ablation study referenced in \cref{subsec:ablation}. 
\Cref{tab:alpha_comparison_by_method} reports the $\alpha$--ablation results for SplitCAM 
and SplitGrad across selected configurations.

Overall, we observe that performance generally improves as $\alpha$ decreases below $0.5$, 
while all metrics---except Maximum Sensitivity---exhibit a sharp performance drop at 
$\alpha = 0.0$. This highlights that the stabilization mechanism in the backward pass plays 
a crucial role in the effectiveness of both methods. While the improvements in the 
Pixel~Flipping metric are modest, the gains observed in the two localization metrics and 
the robustness metric are substantial.

The table further provides pairwise comparisons between configurations using max-pooling 
in convex mode versus the ``winner-takes-it-all'' mode, see \cref{subsec:other_components}. For $\alpha$ values close to $0.5$, 
the differences across the SplitCAM and SplitGrad variants are minimal, indicating that the 
impact of max-pooling on SplitCAM's forward pass is minor in this regime. We emphasize that, 
since split-pair gradients coincide with the scaled version of the original gradients for $\alpha = 0.5$, the choice 
of max-pooling operation has no influence on the gradients in this case. The impact 
becomes most apparent for $\alpha \in \{0.30, 0.35\}$, where the ``winner-takes-it-all'' mode 
shows mostly stronger performance than the monotone mode across all methods and metrics.

\subsection{Forward Pass Stabilization Ablation}

In this subsection we want to investigate the influence of the forward stabilization procedure on SplitLRP's performance of layer configurations selected by the evaluation framework \cref{subsec:eval_framework} as well as performance of SplitCAM on manually selected layer configurations.
As a baseline, we introduce a shifting mode that effectively shifts the activations of positive neurons to half of the respective original activation and the activations of negative neurons to the negative half of it.
We also tested a SplitCAM variant with a forward pass without stabilization, where we observe an activation explosion in late VGG16 layers in the order of $10^{18}$ resulting in numerical errors in the order of $10^9$ despite trying to correct the activations in every layer such that their difference equals the original activation.
For SplitLRP, the absence of the forward stabilization produced nearly entirely white saliency maps with only small, random, and semantically meaningless artifacts. Since these results were clearly unusable, we did not compute quantitative metrics for this setting.

\subsubsection{SplitCAM}

\Cref{tab:alpha_comparison_none} summarizes the performance of all SplitCAM configurations---corresponding to those in \cref{tab:alpha_comparison_shift}---\emph{without} forward-pass stabilization.
\Cref{tab:alpha_comparison_shift} presents the results \emph{with} the shifting-based stabilization procedure.

\subsubsection{SplitLRP}

\Cref{tab:lrp_scale_shift_comparison} directly compares SplitLRP using shifting versus scaling stabilization in the forward pass.
For both layer-7 configurations, replacing scaling with shifting leads to substantial performance drops across all metrics.
For the later-layer configurations, the picture is mixed: while shifting yields slightly higher Attribution Localization scores, the scaling variants achieve markedly better Pointing Game performance and significantly improved robustness (Maximum Sensitivity).

These findings highlight the clear advantage of using scaled split-pair activations -- and not the original activations -- for LRP within the DC-decomposition framework.

\input{appendix/table_lrp_scale_shift_comparison}
\input{appendix/table_alpha_none}
\input{appendix/table_alpha_shift}

\subsection{Layer Dependence Analysis}

\begin{figure}[htbp]
\centering
\includegraphics[width=0.5\textwidth]{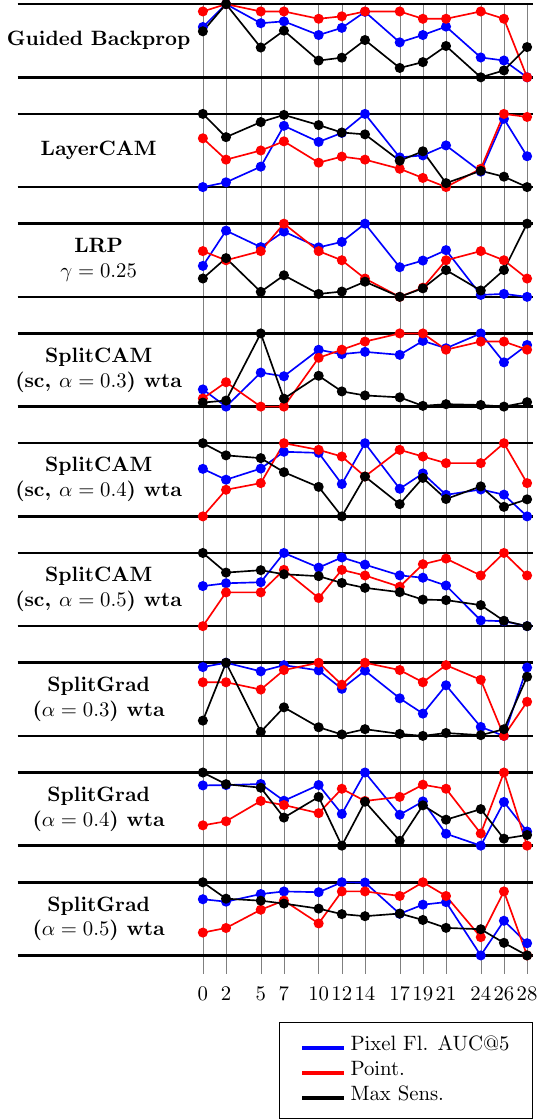}
\caption{Layer dependence for different attribution methods. Each method is normalized independently to show relative performance across layers.}
\label{fig:layer_dependence}
\end{figure}

Figure~\ref{fig:layer_dependence} shows the layer dependence of different attribution methods.
Each method is normalized independently to visualize the relative performance across different layers. Unlike the ablations of the previous section, here we only evaluated across the custom validation subset of ImageNet-S.

Overall, the plots reveal a strong layer dependence of all considered methods and metrics.
Surprisingly, the $\gamma$-LRP does not show the strongest performance in the first layer.
There are clear patterns for each of the metrics, valid across different methods.
Pixel Flipping performs exceptionally well in layer 14 across almost all methods, except SplitCAM and SplitGrad with $\alpha=0.3$.
Maximum sensitivity performs best in later layers, while exploding in the last layers for LRP and SplitGrad with $\alpha=0.3$.

Figure~\ref{fig:layer_dependence} resembles a characteristic fingerprint of a trained classification model.
It reveals which layers localize most accurately, which exhibit the highest robustness to small input perturbations, and which activation or gradient pathways most strongly influence the Pixel Flipping behavior.
These layer-wise patterns may provide insights into how evidence is processed throughout the network and could potentially guide future diagnostics or architectural analysis.
However, a deeper investigation of these research questions lies beyond the scope of this work.

%% file: appendix/table_ppendix_alpha_combined.tex
\renewcommand\cellalign{cc}

\begin{table*}[h!]
    \centering
    \footnotesize
    \setlength{\tabcolsep}{5pt}
    \renewcommand{\arraystretch}{1.2}

    \begin{tabular}{|c|c|c|c|c|c|c|c|}
    \noalign{\global\arrayrulewidth=1pt}
    \hline
    \noalign{\global\arrayrulewidth=0.4pt}
    \textbf{Method} & \textbf{Metric} & $\alpha \ 0.50$ & $\alpha \ 0.45$ & $\alpha \ 0.40$ & $\alpha \ 0.35$ & $\alpha \ 0.30$ & $\alpha \ 0.00$ \\
    \noalign{\global\arrayrulewidth=1pt}
    \hline
    \noalign{\global\arrayrulewidth=0.4pt}
    \multirow{4}{*}{\makecell{\textbf{SplitGrad (wta)}\\(26, g), abs}} & Point. $\uparrow$ & 0.8975 & 0.9064 & 0.9470 & \textbf{0.9558} & 0.8887 & 0.4011 \\
    \cline{2-8}
     & Attr. Loc. $\uparrow$ & 0.5434 & 0.5476 & 0.6051 & \textbf{0.6169} & 0.5346 & 0.4486 \\
    \cline{2-8}
     & Max Sens. $\downarrow$ & 0.6160 & 0.6071 & 0.4399 & 0.2382 & 0.3344 & \textbf{0.0283} \\
    \cline{2-8}
     & Pixel Fl. @20 $\uparrow$ & 1.5270 & 1.5337 & \textbf{1.5808} & 1.4476 & 1.3171 & 1.2471 \\
    \noalign{\global\arrayrulewidth=1pt}
    \hline
    \noalign{\global\arrayrulewidth=0.4pt}
    \multirow{4}{*}{\makecell{\textbf{SplitGrad}\\(26, g), abs}} & Point. $\uparrow$ & 0.8975 & 0.9064 & \textbf{0.9470} & 0.9099 & 0.2827 & 0.6413 \\
    \cline{2-8}
     & Attr. Loc. $\uparrow$ & 0.5434 & 0.5476 & \textbf{0.6037} & 0.5216 & 0.4245 & 0.4563 \\
    \cline{2-8}
     & Max Sens. $\downarrow$ & 0.6171 & 0.6102 & 0.4426 & 0.3864 & 0.0622 & \textbf{0.0079} \\
    \cline{2-8}
     & Pixel Fl. @20 $\uparrow$ & 1.5272 & 1.5333 & \textbf{1.5876} & 1.3702 & 0.2431 & 0.5755 \\
    \noalign{\global\arrayrulewidth=1pt}
    \hline
    \noalign{\global\arrayrulewidth=0.4pt}
    \multirow{4}{*}{\makecell{\textbf{SplitGrad (wta)}\\(14, g), abs}} & Point. $\uparrow$ & 0.8198 & 0.8198 & 0.8198 & \textbf{0.8322} & 0.8127 & 0.7898 \\
    \cline{2-8}
     & Attr. Loc. $\uparrow$ & \textbf{0.5403} & 0.5403 & 0.5402 & 0.5006 & 0.4936 & 0.4857 \\
    \cline{2-8}
     & Max Sens. $\downarrow$ & 0.8579 & 0.8597 & 0.8584 & 0.0875 & 0.0677 & \textbf{0.0481} \\
    \cline{2-8}
     & Pixel Fl. @20 $\uparrow$ & 2.1887 & \textbf{2.1894} & 2.1869 & 1.8740 & 1.8304 & 1.7684 \\
    \noalign{\global\arrayrulewidth=1pt}
    \hline
    \noalign{\global\arrayrulewidth=0.4pt}
    \multirow{4}{*}{\makecell{\textbf{SplitGrad}\\(14, g), abs}} & Point. $\uparrow$ & 0.8198 & \textbf{0.8198} & 0.8180 & 0.7951 & 0.8004 & 0.7968 \\
    \cline{2-8}
     & Attr. Loc. $\uparrow$ & 0.5403 & \textbf{0.5403} & 0.5330 & 0.5054 & 0.4926 & 0.4847 \\
    \cline{2-8}
     & Max Sens. $\downarrow$ & 0.8618 & 0.8584 & 0.8478 & 0.1103 & 0.0662 & \textbf{0.0301} \\
    \cline{2-8}
     & Pixel Fl. @20 $\uparrow$ & \textbf{2.1894} & 2.1891 & 2.1185 & 1.3425 & 1.2959 & 1.1927 \\
    \noalign{\global\arrayrulewidth=1pt}
    \hline
    \noalign{\global\arrayrulewidth=0.4pt}
    \multirow{4}{*}{\makecell{\textbf{SplitCAM (sc, wta)}\\(26, +, g), abs}} & Point. $\uparrow$ & 0.8675 & 0.8675 & \textbf{0.9382} & 0.9329 & 0.6131 & 0.8551 \\
    \cline{2-8}
     & Attr. Loc. $\uparrow$ & 0.5501 & 0.5535 & 0.6347 & \textbf{0.6627} & 0.4033 & 0.5133 \\
    \cline{2-8}
     & Max Sens. $\downarrow$ & 0.6906 & 0.6820 & 0.4573 & 0.2645 & 0.2511 & \textbf{0.1196} \\
    \cline{2-8}
     & Pixel Fl. @20 $\uparrow$ & 1.6067 & \textbf{1.6085} & 1.5849 & 1.4989 & 0.8534 & 1.5071 \\
    \noalign{\global\arrayrulewidth=1pt}
    \hline
    \noalign{\global\arrayrulewidth=0.4pt}
    \multirow{4}{*}{\makecell{\textbf{SplitCAM (sc)}\\(26, +, g), abs}} & Point. $\uparrow$ & 0.8940 & 0.8905 & \textbf{0.9346} & 0.6466 & 0.8251 & 0.8604 \\
    \cline{2-8}
     & Attr. Loc. $\uparrow$ & 0.5595 & 0.5636 & \textbf{0.6363} & 0.4015 & 0.4977 & 0.5114 \\
    \cline{2-8}
     & Max Sens. $\downarrow$ & 0.6738 & 0.6640 & 0.4520 & 0.3729 & 0.1102 & \textbf{0.1062} \\
    \cline{2-8}
     & Pixel Fl. @20 $\uparrow$ & 1.5902 & \textbf{1.5978} & 1.5906 & 0.9076 & 1.2784 & 1.3605 \\
    \noalign{\global\arrayrulewidth=1pt}
    \hline
    \noalign{\global\arrayrulewidth=0.4pt}
    \multirow{4}{*}{\makecell{\textbf{SplitCAM (sc, wta)}\\(14, g)}} & Point. $\uparrow$ & 0.8251 & 0.8251 & \textbf{0.8269} & 0.7880 & 0.7809 & 0.7703 \\
    \cline{2-8}
     & Attr. Loc. $\uparrow$ & 0.5363 & 0.5363 & \textbf{0.5435} & 0.5338 & 0.5286 & 0.5225 \\
    \cline{2-8}
     & Max Sens. $\downarrow$ & 1.0566 & 1.0569 & 1.0489 & \textbf{0.1620} & 0.1625 & 0.1630 \\
    \cline{2-8}
     & Pixel Fl. @20 $\uparrow$ & 2.4128 & 2.4130 & \textbf{2.4135} & 1.9014 & 1.8613 & 1.8731 \\
    \noalign{\global\arrayrulewidth=1pt}
    \hline
    \noalign{\global\arrayrulewidth=0.4pt}
    \multirow{4}{*}{\makecell{\textbf{SplitCAM (sc)}\\(14, g)}} & Point. $\uparrow$ & 0.8251 & 0.8251 & \textbf{0.8428} & 0.7756 & 0.7686 & 0.7756 \\
    \cline{2-8}
     & Attr. Loc. $\uparrow$ & 0.5363 & 0.5363 & \textbf{0.5626} & 0.5363 & 0.5271 & 0.5208 \\
    \cline{2-8}
     & Max Sens. $\downarrow$ & 1.0552 & 1.0593 & 1.0136 & 0.1629 & 0.1585 & \textbf{0.1555} \\
    \cline{2-8}
     & Pixel Fl. @20 $\uparrow$ & 2.4171 & \textbf{2.4172} & 2.4065 & 1.5032 & 1.5082 & 1.5139 \\
    \noalign{\global\arrayrulewidth=1pt}
    \hline
    \noalign{\global\arrayrulewidth=0.4pt}
    \end{tabular}

    \caption{Alpha ablation as performed in \cref{subsec:ablation} across four SplitGrad and SplitCAM configuration of layers $14$ and $26$ respectively. The notation is similar to \cref{tab:eval_vgg_main}.}
    \label{tab:alpha_comparison_by_method}
\end{table*}

%% file: appendix/table_lrp_scale_shift_comparison.tex
\begin{longtable}{p{4.5cm}|c|cccccc}
    \caption{Scale vs Shift Comparison for SplitLRP Methods on VGG16.}
    \label{tab:lrp_scale_shift_comparison} \\
    \toprule
    \textbf{Method} & \textbf{Type} & \textbf{Select.} & \textbf{Attr. Loc.} & \textbf{Point.} & \textbf{Pixel Fl.} & \textbf{Pixel Fl.} & \textbf{Max Sens.} \\
     & & \textbf{$\downarrow$} & \textbf{$\uparrow$} & \textbf{$\uparrow$} & \textbf{AUC@5 $\uparrow$} & \textbf{AUC@20 $\uparrow$} & \textbf{$\downarrow$} \\
    \midrule
    \endfirsthead
    
    \multicolumn{8}{c}%
    {{\tablename\ \thetable{} -- continued from previous page}} \\
    \toprule
    \textbf{Method} & \textbf{Type} & \textbf{Select.} & \textbf{Attr. Loc.} & \textbf{Point.} & \textbf{Pixel Fl.} & \textbf{Pixel Fl.} & \textbf{Max Sens.} \\
     & & \textbf{$\downarrow$} & \textbf{$\uparrow$} & \textbf{$\uparrow$} & \textbf{AUC@5 $\uparrow$} & \textbf{AUC@20 $\uparrow$} & \textbf{$\downarrow$} \\
    \midrule
    \endhead
    
    \midrule
    \multicolumn{8}{r}{{Continued on next page}} \\
    \midrule
    \endfoot
    
    \bottomrule
    \endlastfoot
    
    \multirow{2}{*}{SplitLRP (wta), 7} & sc & \textbf{3.235} & \textbf{-0.036} & \textbf{0.836} & \textbf{0.296} & \textbf{2.291} & \textbf{0.866} \\
     & sh & 3.619 & -0.610 & 0.698 & 0.260 & 2.070 & 0.914 \\
    \cmidrule(lr){1-8}

    \multirow{2}{*}{SplitLRP (wta), 26, abs} & sc & \textbf{5.168} & 0.588 & 0.871 & 0.151 & 1.473 & \textbf{0.282} \\
     & sh & 5.168 & \textbf{0.628} & \textbf{0.877} & \textbf{0.161} & \textbf{1.478} & 0.468 \\
    \cmidrule(lr){1-8}

    \multirow{2}{*}{SplitLRP, 7} & sc & \textbf{3.235} & \textbf{-0.036} & \textbf{0.836} & \textbf{0.296} & \textbf{2.291} & \textbf{0.866} \\
     & sh & 3.619 & -0.695 & 0.698 & 0.260 & 2.070 & 0.914 \\
    \cmidrule(lr){1-8}

    \multirow{2}{*}{SplitLRP, 19} & sc & 5.168 & 0.588 & \textbf{0.871} & 0.151 & 1.473 & \textbf{0.282} \\
     & sh & \textbf{4.955} & \textbf{0.629} & 0.853 & \textbf{0.197} & \textbf{1.692} & 0.507 \\
    \midrule
\end{longtable}

%% file: appendix/table_alpha_none.tex
\renewcommand\cellalign{cc}

\begin{table}[htbp]
    \centering
    \footnotesize
    \setlength{\tabcolsep}{5pt}
    \renewcommand{\arraystretch}{1.2}

    \begin{tabular}{|c|c|c|c|c|c|c|c|}
    \noalign{\global\arrayrulewidth=1pt}
    \hline
    \noalign{\global\arrayrulewidth=0.4pt}
    \textbf{Method} & \textbf{Metric} & $\alpha \ 0.50$ & $\alpha \ 0.45$ & $\alpha \ 0.40$ & $\alpha \ 0.35$ & $\alpha \ 0.30$ & $\alpha \ 0.00$ \\
    \noalign{\global\arrayrulewidth=1pt}
    \hline
    \noalign{\global\arrayrulewidth=0.4pt}
    \multirow{4}{*}{\makecell{\textbf{SplitCAM (sc, wta)}\\(26, +, g), abs}} & Point. $\uparrow$ & 0.7686 & 0.7686 & 0.8632 & \textbf{0.8712} & 0.3441 & 0.5453 \\
    \cline{2-8}
     & Attr. Loc. $\uparrow$ & 0.5374 & 0.5412 & 0.5943 & \textbf{0.6118} & 0.3595 & 0.4569 \\
    \cline{2-8}
     & Max Sens. $\downarrow$ & \cellcolor{green!20}0.6578 & \cellcolor{green!20}0.6507 & 0.4841 & 0.2873 & \cellcolor{green!20}0.1918 & \cellcolor{green!20}\textbf{0.0135} \\
    \cline{2-8}
     & Pixel Fl. @20 $\uparrow$ & 1.2744 & 1.2859 & \textbf{1.3209} & 1.1792 & 0.4430 & 0.6527 \\
    \noalign{\global\arrayrulewidth=1pt}
    \hline
    \noalign{\global\arrayrulewidth=0.4pt}
    \multirow{4}{*}{\makecell{\textbf{SplitCAM (sc)}\\(26, +, g), abs}} & Point. $\uparrow$ & 0.7565 & 0.7565 & \textbf{0.8451} & 0.3783 & 0.4507 & 0.4970 \\
    \cline{2-8}
     & Attr. Loc. $\uparrow$ & 0.5346 & 0.5384 & \textbf{0.5897} & 0.3663 & 0.4420 & 0.4545 \\
    \cline{2-8}
     & Max Sens. $\downarrow$ & \cellcolor{green!20}0.6604 & \cellcolor{green!20}0.6512 & 0.4985 & \cellcolor{green!20}0.3077 & \cellcolor{green!20}0.0279 & \cellcolor{green!20}\textbf{0.0043} \\
    \cline{2-8}
     & Pixel Fl. @20 $\uparrow$ & 1.2680 & 1.2856 & \textbf{1.3295} & 0.5912 & 0.4906 & 0.5563 \\
    \noalign{\global\arrayrulewidth=1pt}
    \hline
    \noalign{\global\arrayrulewidth=0.4pt}
    \multirow{4}{*}{\makecell{\textbf{SplitCAM (sc, wta)}\\(14, g)}} & Point. $\uparrow$ & \cellcolor{green!20}\textbf{0.8491} & \cellcolor{green!20}0.8491 & \cellcolor{green!20}0.8491 & 0.7867 & \cellcolor{green!20}0.7847 & 0.7686 \\
    \cline{2-8}
     & Attr. Loc. $\uparrow$ & 0.5324 & 0.5321 & \textbf{0.5406} & 0.5306 & 0.5261 & 0.5200 \\
    \cline{2-8}
     & Max Sens. $\downarrow$ & 1.0834 & 1.0814 & 1.0885 & \textbf{0.2271} & 0.2309 & 0.2384 \\
    \cline{2-8}
     & Pixel Fl. @20 $\uparrow$ & \cellcolor{green!20}2.4454 & \cellcolor{green!20}2.4482 & \cellcolor{green!20}\textbf{2.4486} & \cellcolor{green!20}1.9588 & \cellcolor{green!20}1.9193 & \cellcolor{green!20}1.9349 \\
    \noalign{\global\arrayrulewidth=1pt}
    \hline
    \noalign{\global\arrayrulewidth=0.4pt}
    \multirow{4}{*}{\makecell{\textbf{SplitCAM (sc)}\\(14, g)}} & Point. $\uparrow$ & 0.7746 & 0.7767 & \textbf{0.7847} & 0.5855 & 0.5594 & 0.5573 \\
    \cline{2-8}
     & Attr. Loc. $\uparrow$ & 0.4003 & 0.4144 & \textbf{0.5517} & 0.5300 & 0.5209 & 0.5137 \\
    \cline{2-8}
     & Max Sens. $\downarrow$ & 1.5733 & 1.5627 & 1.5443 & \textbf{1.2661} & 1.3939 & 1.4272 \\
    \cline{2-8}
     & Pixel Fl. @20 $\uparrow$ & 2.1298 & \textbf{2.1384} & 2.1371 & \cellcolor{green!20}1.6669 & \cellcolor{green!20}1.6793 & \cellcolor{green!20}1.6740 \\
    \noalign{\global\arrayrulewidth=1pt}
    \hline
    \noalign{\global\arrayrulewidth=0.4pt}
    \end{tabular}

    \caption{Alpha ablation for scale normalization across four SplitCAM configurations of layers 14 and 26. Here, in the forward pass no stabilization is applied. Notation and structure is similar to \cref{tab:alpha_comparison_by_method}. Green cells indicate better performance of the method compared to the default variant with scale stabilization in the forward pass.}
    \label{tab:alpha_comparison_none}
\end{table}

%% file: appendix/table_alpha_shift.tex
\renewcommand\cellalign{cc}

\begin{table}[htbp]
    \centering
    \footnotesize
    \setlength{\tabcolsep}{5pt}
    \renewcommand{\arraystretch}{1.2}

    \begin{tabular}{|c|c|c|c|c|c|c|c|}
    \noalign{\global\arrayrulewidth=1pt}
    \hline
    \noalign{\global\arrayrulewidth=0.4pt}
    \textbf{Method} & \textbf{Metric} & $\alpha \ 0.50$ & $\alpha \ 0.45$ & $\alpha \ 0.40$ & $\alpha \ 0.35$ & $\alpha \ 0.30$ & $\alpha \ 0.00$ \\
    \noalign{\global\arrayrulewidth=1pt}
    \hline
    \noalign{\global\arrayrulewidth=0.4pt}
    \multirow{4}{*}{\makecell{\textbf{SplitCAM (sc, wta)}\\(26, +, g), abs}} & Point. $\uparrow$ & 0.8330 & 0.8149 & 0.7968 & \textbf{0.9235} & 0.6036 & \cellcolor{green!20}0.8592 \\
    \cline{2-8}
     & Attr. Loc. $\uparrow$ & 0.5242 & 0.5203 & 0.5147 & \textbf{0.6549} & 0.3968 & 0.5093 \\
    \cline{2-8}
     & Max Sens. $\downarrow$ & \cellcolor{green!20}0.6180 & \cellcolor{green!20}0.6182 & 0.6509 & 0.3104 & 0.2791 & \textbf{0.1665} \\
    \cline{2-8}
     & Pixel Fl. @20 $\uparrow$ & \cellcolor{green!20}\textbf{1.6804} & \cellcolor{green!20}1.6646 & 1.5065 & 1.4962 & \cellcolor{green!20}0.8896 & \cellcolor{green!20}1.5405 \\
    \noalign{\global\arrayrulewidth=1pt}
    \hline
    \noalign{\global\arrayrulewidth=0.4pt}
    \multirow{4}{*}{\makecell{\textbf{SplitCAM (sc)}\\(26, +, g), abs}} & Point. $\uparrow$ & 0.8330 & 0.8149 & 0.7988 & 0.5654 & \cellcolor{green!20}0.8431 & \cellcolor{green!20}\textbf{0.8652} \\
    \cline{2-8}
     & Attr. Loc. $\uparrow$ & \textbf{0.5242} & 0.5203 & 0.5076 & 0.3736 & 0.4949 & 0.5094 \\
    \cline{2-8}
     & Max Sens. $\downarrow$ & \cellcolor{green!20}0.6129 & \cellcolor{green!20}0.6168 & 0.6494 & 0.3820 & 0.1686 & \textbf{0.1664} \\
    \cline{2-8}
     & Pixel Fl. @20 $\uparrow$ & \cellcolor{green!20}\textbf{1.6804} & \cellcolor{green!20}1.6646 & 1.5105 & 0.8714 & \cellcolor{green!20}1.4289 & \cellcolor{green!20}1.4765 \\
    \noalign{\global\arrayrulewidth=1pt}
    \hline
    \noalign{\global\arrayrulewidth=0.4pt}
    \multirow{4}{*}{\makecell{\textbf{SplitCAM (sc, wta)}\\(14, g)}} & Point. $\uparrow$ & \cellcolor{green!20}\textbf{0.8310} & \cellcolor{green!20}0.8310 & \cellcolor{green!20}0.8310 & \cellcolor{green!20}0.7887 & \cellcolor{green!20}0.7827 & \cellcolor{green!20}0.7726 \\
    \cline{2-8}
     & Attr. Loc. $\uparrow$ & 0.5312 & 0.5312 & \textbf{0.5376} & 0.5315 & 0.5264 & 0.5203 \\
    \cline{2-8}
     & Max Sens. $\downarrow$ & \cellcolor{green!20}1.0562 & \cellcolor{green!20}1.0521 & 1.0649 & \textbf{0.1633} & 0.1641 & 0.1642 \\
    \cline{2-8}
     & Pixel Fl. @20 $\uparrow$ & \cellcolor{green!20}\textbf{2.4412} & \cellcolor{green!20}2.4412 & \cellcolor{green!20}2.4400 & \cellcolor{green!20}1.9306 & \cellcolor{green!20}1.8953 & \cellcolor{green!20}1.9059 \\
    \noalign{\global\arrayrulewidth=1pt}
    \hline
    \noalign{\global\arrayrulewidth=0.4pt}
    \multirow{4}{*}{\makecell{\textbf{SplitCAM (sc)}\\(14, g)}} & Point. $\uparrow$ & \cellcolor{green!20}0.8310 & \cellcolor{green!20}0.8310 & \cellcolor{green!20}\textbf{0.8431} & 0.7746 & \cellcolor{green!20}0.7706 & \cellcolor{green!20}0.7807 \\
    \cline{2-8}
     & Attr. Loc. $\uparrow$ & 0.5311 & 0.5312 & \textbf{0.5580} & 0.5337 & 0.5245 & 0.5183 \\
    \cline{2-8}
     & Max Sens. $\downarrow$ & 1.0554 & \cellcolor{green!20}1.0568 & 1.0198 & 0.1641 & 0.1587 & \textbf{0.1570} \\
    \cline{2-8}
     & Pixel Fl. @20 $\uparrow$ & \cellcolor{green!20}\textbf{2.4412} & \cellcolor{green!20}2.4411 & \cellcolor{green!20}2.4318 & \cellcolor{green!20}1.5187 & \cellcolor{green!20}1.5222 & \cellcolor{green!20}1.5329 \\
    \noalign{\global\arrayrulewidth=1pt}
    \hline
    \noalign{\global\arrayrulewidth=0.4pt}
    \end{tabular}

    \caption{Alpha ablation for scaling normalization across four SplitCAM configurations of layers 14 and 26. Here, in the forward pass we apply shift normalization. Notation and structure is similar to \cref{tab:alpha_comparison_by_method}. Green cells indicate better performance of the method compared to the default variant with scale stabilization in the forward pass.}
    \label{tab:alpha_comparison_shift}
\end{table}

%% file: appendix/dcinns.tex
\section{Difference of Networks}
\label{app:explaining_io_convex_nns}

In this appendix, we empirically investigate two questions using a simple image classification task:
\begin{enumerate}
    \item[(A)] Are the faithful saliency maps observed in \cref{sec:experiments} a general property of monotone neural networks?
    \item[(B)] Can we directly train a \emph{difference-of-two monotone neural networks}, rather than obtaining such a decomposition by splitting a general architecture? 
\end{enumerate}

We introduced the two resulting architectures as DMs and DICs as well as their inverted-input variations DMINs and DICINs in \cref{subsec:dc_network_training_methods}. \citetapp{AppSankaranarayanan2021CDINN} proposed an architecture similar to DICs as Convex Difference Neural Network (\emph{CDiNN}) in the context of optimal control.
Our experiments provide evidence that montone neural networks are inherently self-explanatory: ICNNs consistently focus on the most informative present and missing visual features in the image classification task. We remark that here we view ICNNs as monotone networks on an affine transformation of the input embodied by the first network layer. The gradients of the DICIN models clearly separate their concerns: The gradients of the $g$-model provide factual, while the gradients of the $h$-model provide counterfactual explanations.
Moreover, all proposed network architectures achieve excellent test accuracy on MNIST.

\subsection{Experimental Setup} \label{subsec:dic_experiment}

We evaluate two representative architectures -- as rank-2 Maxout networks consistently outperformed standard ReLU networks in the difference setting, we settled with the Maxout variants of two classical neural network architectures. We remark that this choice aligns with the fact that the split-streams in \cref{sec:method} effectively are rank-2 Maxout networks.

\paragraph{\texttt{cnn\_maxout}}
A convolutional neural network with Maxout activations.
It consists of four convolutional layers with 32, 64, 128, and 128 channels, each using $3\times3$ kernels with padding 1 and batch normalization after every layer.
After flattening the network applies three fully connected layers with dimensions $6272 \rightarrow 1024 \rightarrow 256 \rightarrow 10$, applying rank-2 Maxout activations after each hidden layer.
In total, it contains approximately 1.2M parameters, including the Maxout expansion.

\paragraph{\texttt{mlp\_maxout}}
A wide multilayer perceptron that also employs Maxout units.
It comprises four fully connected layers with dimensions $784 \rightarrow 800 \rightarrow 800 \rightarrow 800 \rightarrow 10$, using rank-2 Maxout activations after each hidden layer and batch normalization after the first three layers.
This network contains approximately 4.2M parameters, including the Maxout expansion.

\paragraph{Restricting the weights}

To enforce non-negativity we replace weight matrices $W$ with trainable parameters $\tilde{W}$ and apply a positivity transform $p$ such that the effective weights are $p(\tilde{W})$.
We explored several choices for $p$ (\eg \emph{softplus}, exponential, square) and found that $p=\text{ReLU}$ yielded the best test accuracy when all $\tilde{W}$ were initialized by sampling from a Gaussian distribution and taking the absolute value.
As a technical note, we also apply $p$ to the scaling and shifting parameters of all \emph{batch normalization} layers, as well as to all biases.

Because non-negative weights can cause exploding activations, we use a small initialization scale for all base model weights and apply a weight decay of $10^{-3}$ in all experiments.

\subsection{Results} \label{subsec:dic_results}

\input{appendix/table_model_accuracy.tex}

\Cref{tab:cdinn_acc} reports the median test accuracies of all considered architectures across 10 trainings with different random seeds. As a baseline we also trained and evaluated the difference of two unrestricted models, denoted by 
Diff\_None as well as an unrestricted model, denoted by None. The highest test performances are observed from \texttt{cnn\_maxout} None and IC models, reaching a median accuracy of $99.19\%$ and $99.16\%$ respectively. We remark that all our models achieve strong test accuracies.

For the \texttt{mlp\_maxout} the split variants DIC and DM outperform their respective single model counterparts IC and Mon. We assume that the strong inductive biases underlying a single \texttt{cnn\_maxout} model that are favorable for the MNIST task are a reason why this does not translate to the \texttt{cnn\_maxout} base model.
For both basemodels the inverted-input mode boosts the difference-of-two architecture.

In case of \texttt{mlp\_maxout} the IC and Mon are competitive with the unrestricted base model, while the test results are more seed-dependent for the restricted models and show higher variance. All the difference-of-two variants outperform the base model.

In case of \texttt{cnn\_maxout}, the IC variant achieves a surprisingly high median test accuracy of $99.16 \%$ comparable to the performance of the unrestricted baseline, even showing better minimal test accuracy than the baseline.
All the difference-of-two variants achieve strong, but unambiguously worse test accuracies compared to the single-model variants.

\input{appendix/figures/mnist_ic_saliency.tex}

\cref{fig:mnist_ic_saliency} visualizes input gradients of the \texttt{cnn\_maxout} DIC model on representative test samples.
The gradients are mostly positive on the strokes and negative on the background, indicating that adding white pixels in certain regions and deleting stroke pixels would decrease the predicted confidence.
We observe that the hole of the ``0'', the gaps between the arches of ``3'' (distinguishing it from ``8'' and ``9''), the area between the belly of ``6'' and the arch of ``8'', and the vertical line and lower curve of ``7'' and ``8'' respectively, are all critical for classification.
It is remarkable that the positive gradient aligns precisely with the digit’s strokes.

However, the gradients of the \texttt{cnn\_maxout} IC do not exhibit class-dependent directional variation, in contrast to those of the \texttt{cnn\_maxout} DIC model.
\Cref{fig:mnist_split_ic_saliency_main} shows overlays of $g$- and $h$-network gradients (red and blue, respectively) for all classes on two test samples with correctly predicted ground-truth label ``3''.
Since the $h$-model receives $1 - x$ as input, a high gradient value (bright blue pixel) indicates that adding a white pixel at that location would increase the corresponding class logit.

Class ``1'' shows no $g$-gradient activity above the threshold, while the $g$-gradient for the ground-truth class ``3'' is strongest and follows the digit strokes most accurately.
For classes ``6'', ``8'', and ``9'', the $h$-gradients are dense between the arches of the ``3'', indicating that those missing pixels are discriminative for these classes.
In several cases, red gradients also align with the strokes of competing classes within the original digit structure.

\subsection{Conclusion}

Considering our results with small models on MNIST, we can answer both questions raised at the beginning of this appendix affirmatively.
Concerning question (B), we observed strong performance of our proposed DM and DIC models, in case of the small MLP even outperforming all single-model variants.
Concerning question (A), we even report better interpretability of the monotone and input-convex models in their single-model and difference-of-two setups compared to the DC decompositions of pretrained network architectures.
Indeed, we were not able to rigorously prove the disentanglement of concerns to factual and counterfactual features in the split-pairs of the pretrained VGG16 and ResNet18 models.
It is future work to translate our ideas from the proof of concept presented in this appendix to larger network architectures.

%% file: appendix/table_model_accuracy.tex
\begin{table}[h!]
\centering
\caption{Test accuracy across restriction modes (\%). Values show median with [min, max] range over 10 runs.}
\label{tab:cdinn_acc}
\begin{tabular}{lcccccccc}
\toprule
Model & IC & Mon & None & DIC & DM & Diff\_None & DICIN & DMIN \\
\midrule
\texttt{mlp\_maxout} & \makecell{98.36 \\ \scriptsize [98.07, 98.69]} & \makecell{98.30 \\ \scriptsize [97.64, 98.49]} & \makecell{98.39 \\ \scriptsize [98.07, 98.63]} & \makecell{98.56 \\ \scriptsize [98.33, 98.96]} & \makecell{98.55 \\ \scriptsize [98.23, 98.71]} & \makecell{98.47 \\ \scriptsize [98.19, 98.72]} & \makecell{98.58 \\ \scriptsize [98.41, 98.84]} & \makecell{98.54 \\ \scriptsize [98.36, 98.65]} \\
\texttt{cnn\_maxout} & \makecell{99.16 \\ \scriptsize [98.86, 99.27]} & \makecell{98.50 \\ \scriptsize [98.19, 98.79]} & \makecell{99.19 \\ \scriptsize [98.61, 99.44]} & \makecell{97.94 \\ \scriptsize [97.81, 98.04]} & \makecell{98.06 \\ \scriptsize [97.99, 98.19]} & \makecell{97.94 \\ \scriptsize [97.72, 98.45]} & \makecell{98.17 \\ \scriptsize [97.96, 98.34]} & \makecell{98.28 \\ \scriptsize [98.17, 98.54]} \\
\bottomrule
\end{tabular}
\end{table}

%% file: appendix/figures/mnist_ic_saliency.tex
\begin{figure}[H]
    \centering
    \begin{subfigure}[b]{0.09\textwidth}
        \centering
        \includegraphics[width=\textwidth]{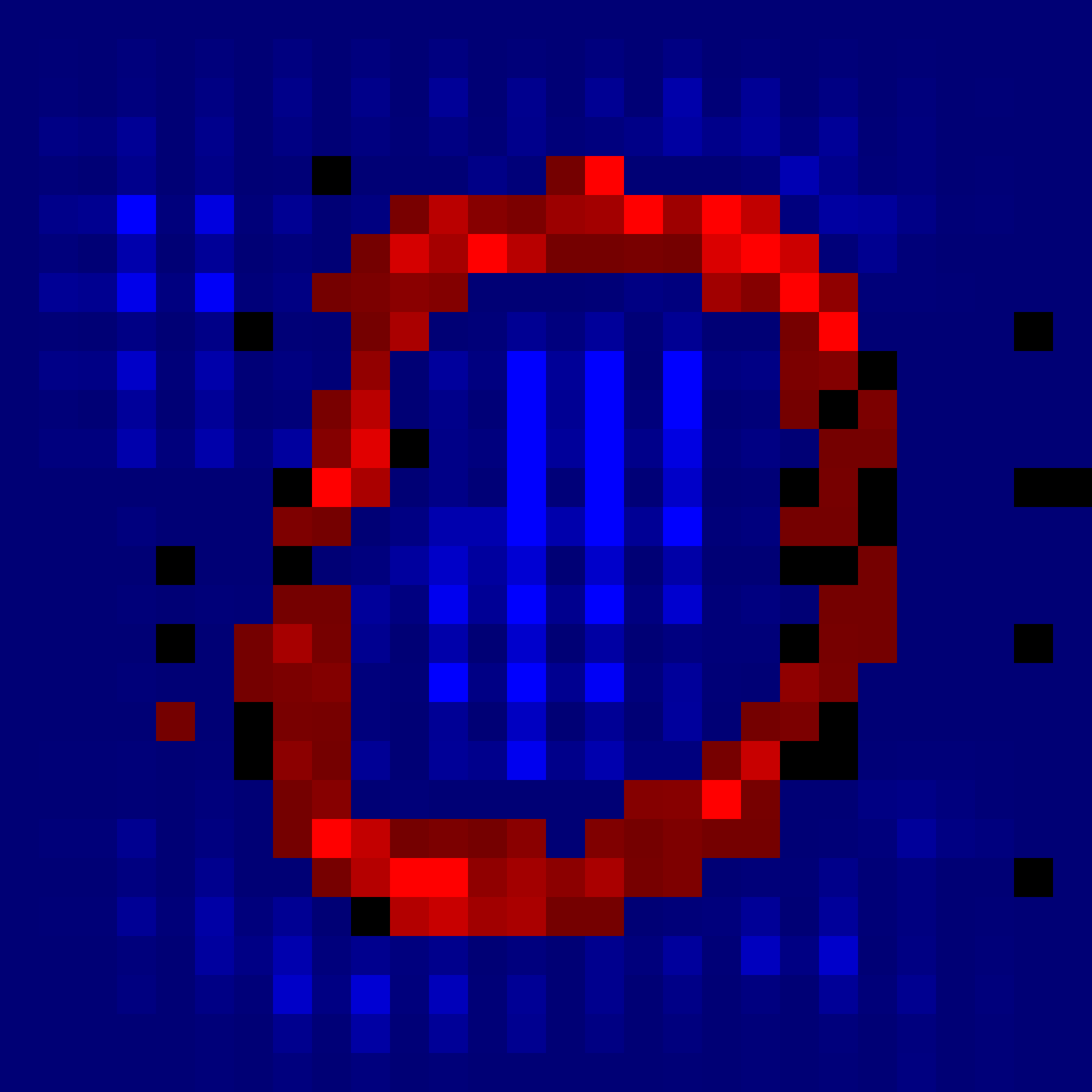}
    \end{subfigure}
    \begin{subfigure}[b]{0.09\textwidth}
        \centering
        \includegraphics[width=\textwidth]{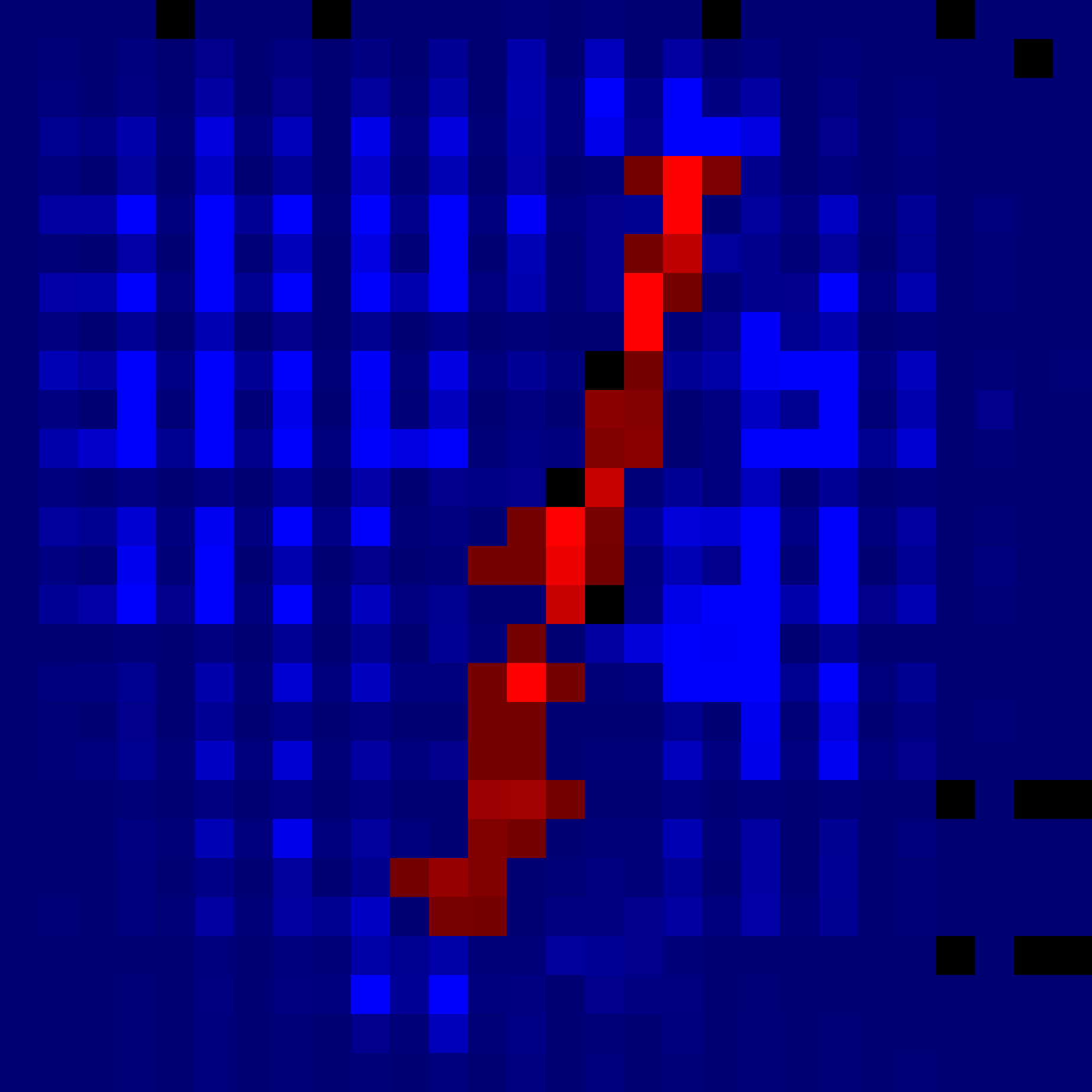}
    \end{subfigure}
    \begin{subfigure}[b]{0.09\textwidth}
        \centering
        \includegraphics[width=\textwidth]{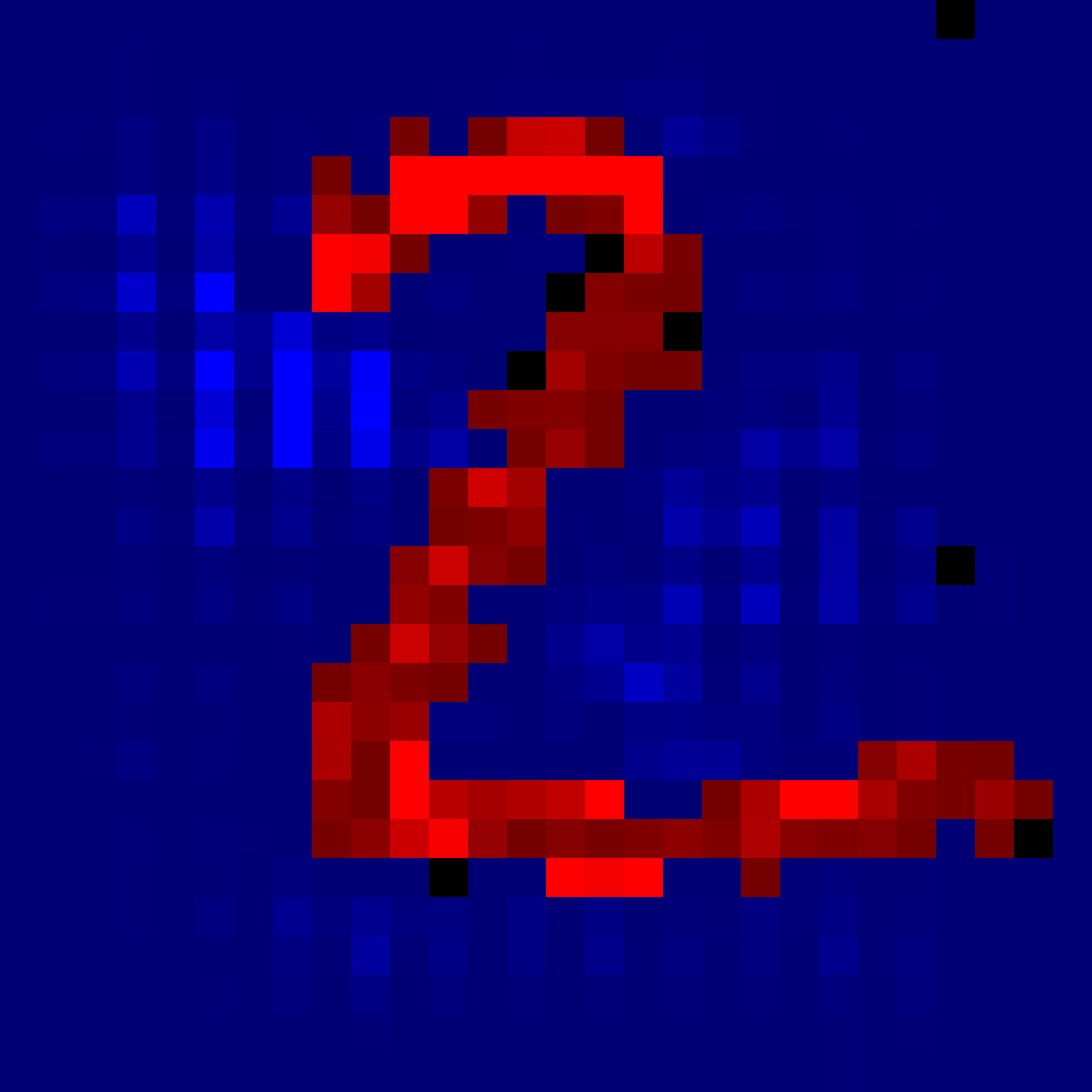}
    \end{subfigure}
    \begin{subfigure}[b]{0.09\textwidth}
        \centering
        \includegraphics[width=\textwidth]{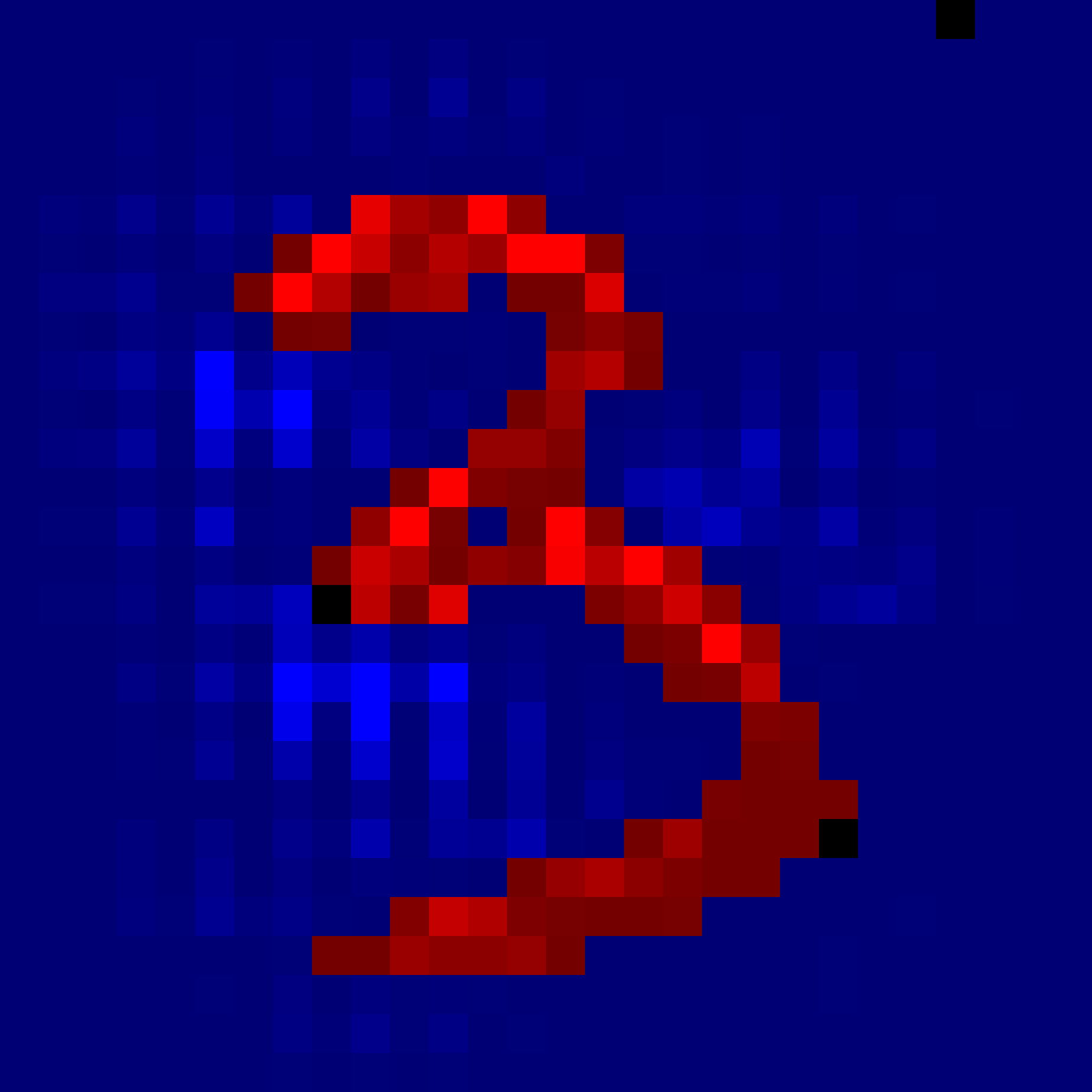}
    \end{subfigure}
    \begin{subfigure}[b]{0.09\textwidth}
        \centering
        \includegraphics[width=\textwidth]{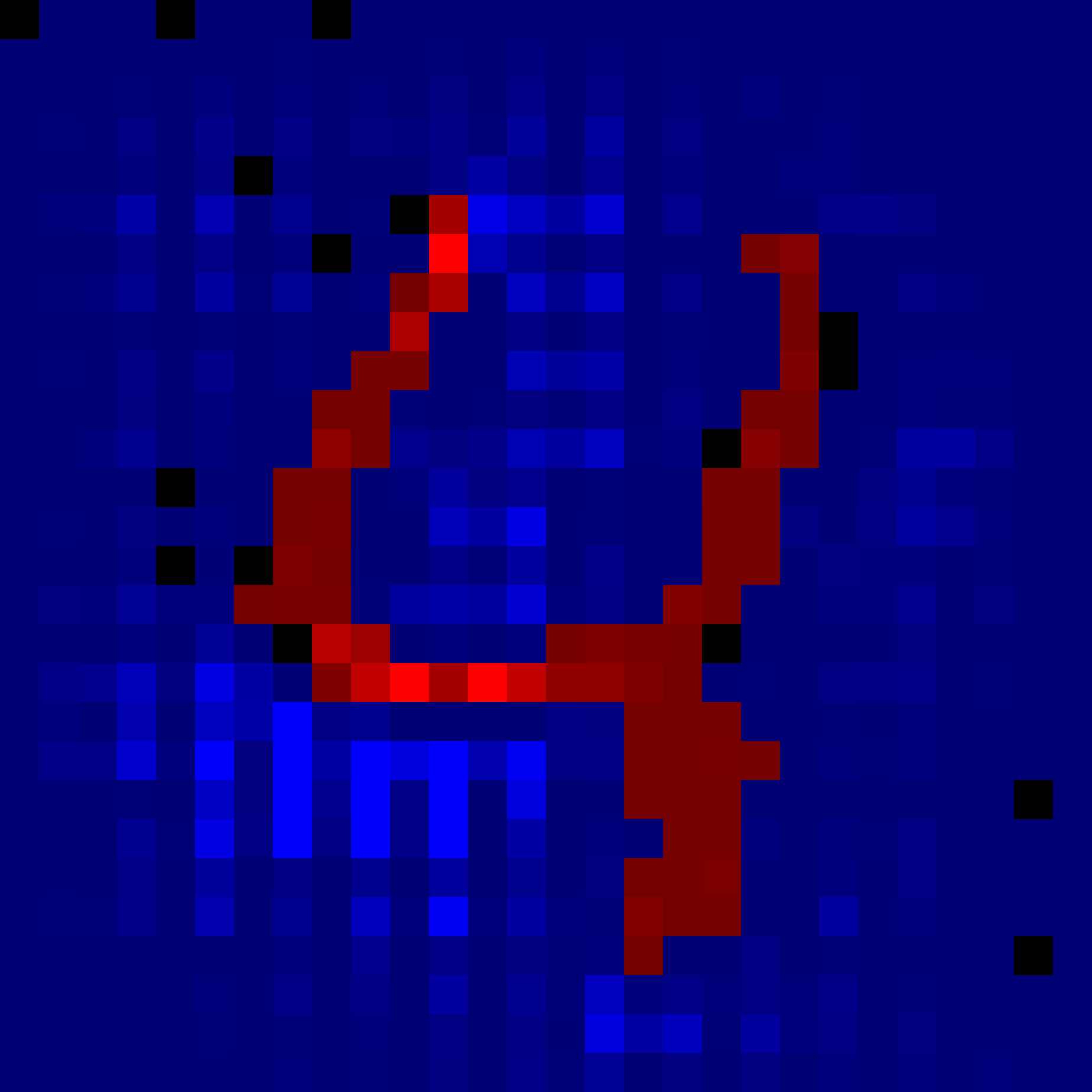}
    \end{subfigure}
    \begin{subfigure}[b]{0.09\textwidth}
        \centering
        \includegraphics[width=\textwidth]{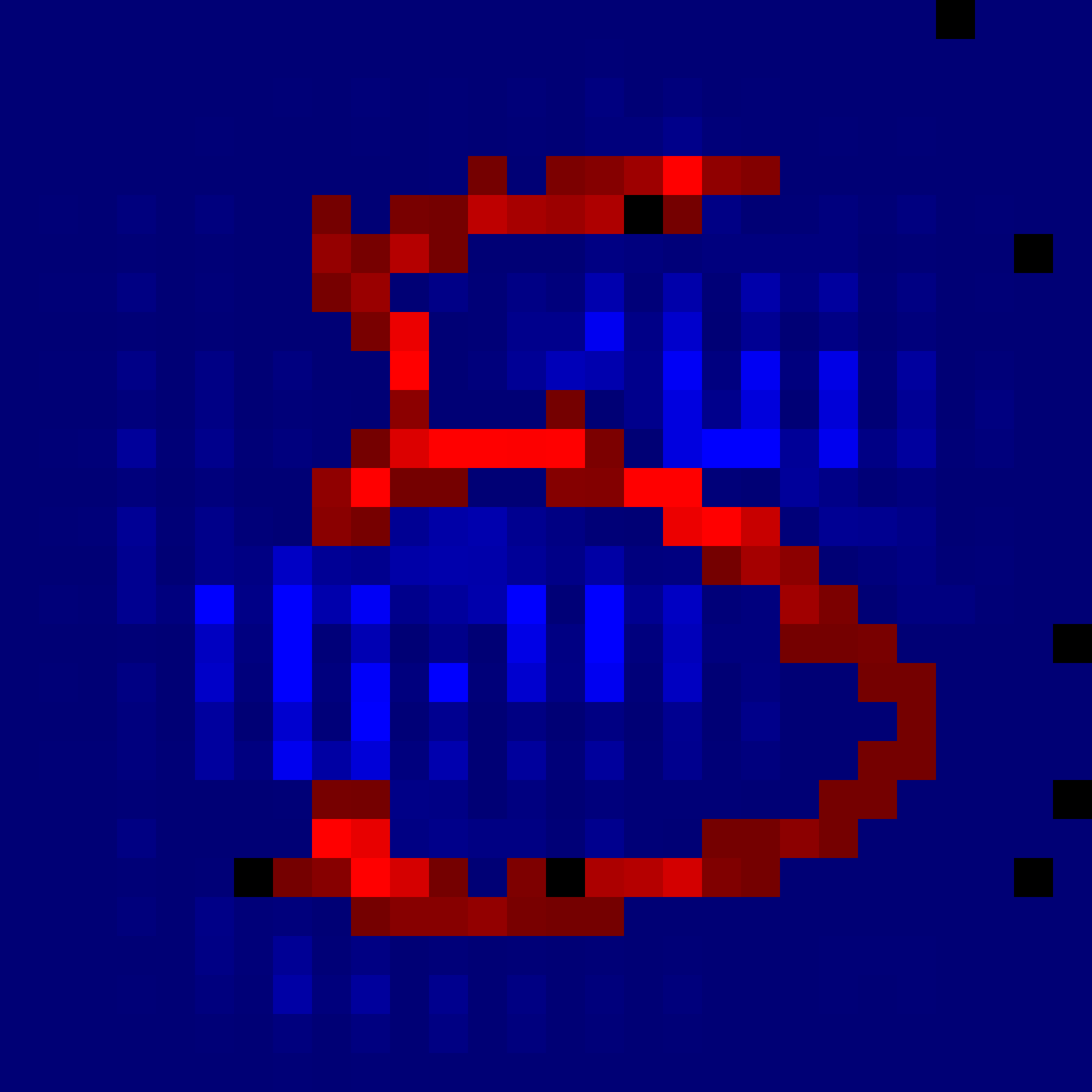}
    \end{subfigure}
    \begin{subfigure}[b]{0.09\textwidth}
        \centering
        \includegraphics[width=\textwidth]{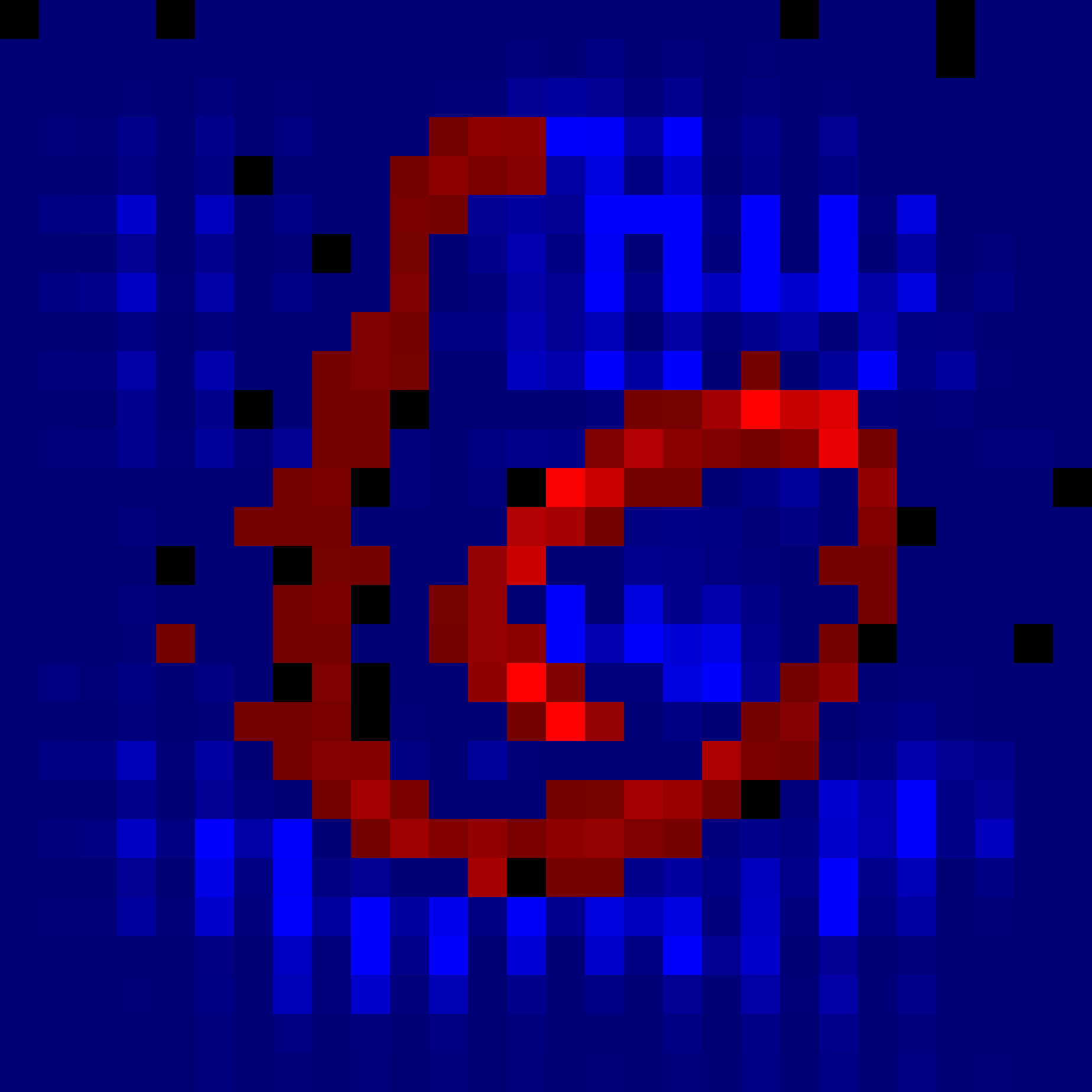}
    \end{subfigure}
    \begin{subfigure}[b]{0.09\textwidth}
        \centering
        \includegraphics[width=\textwidth]{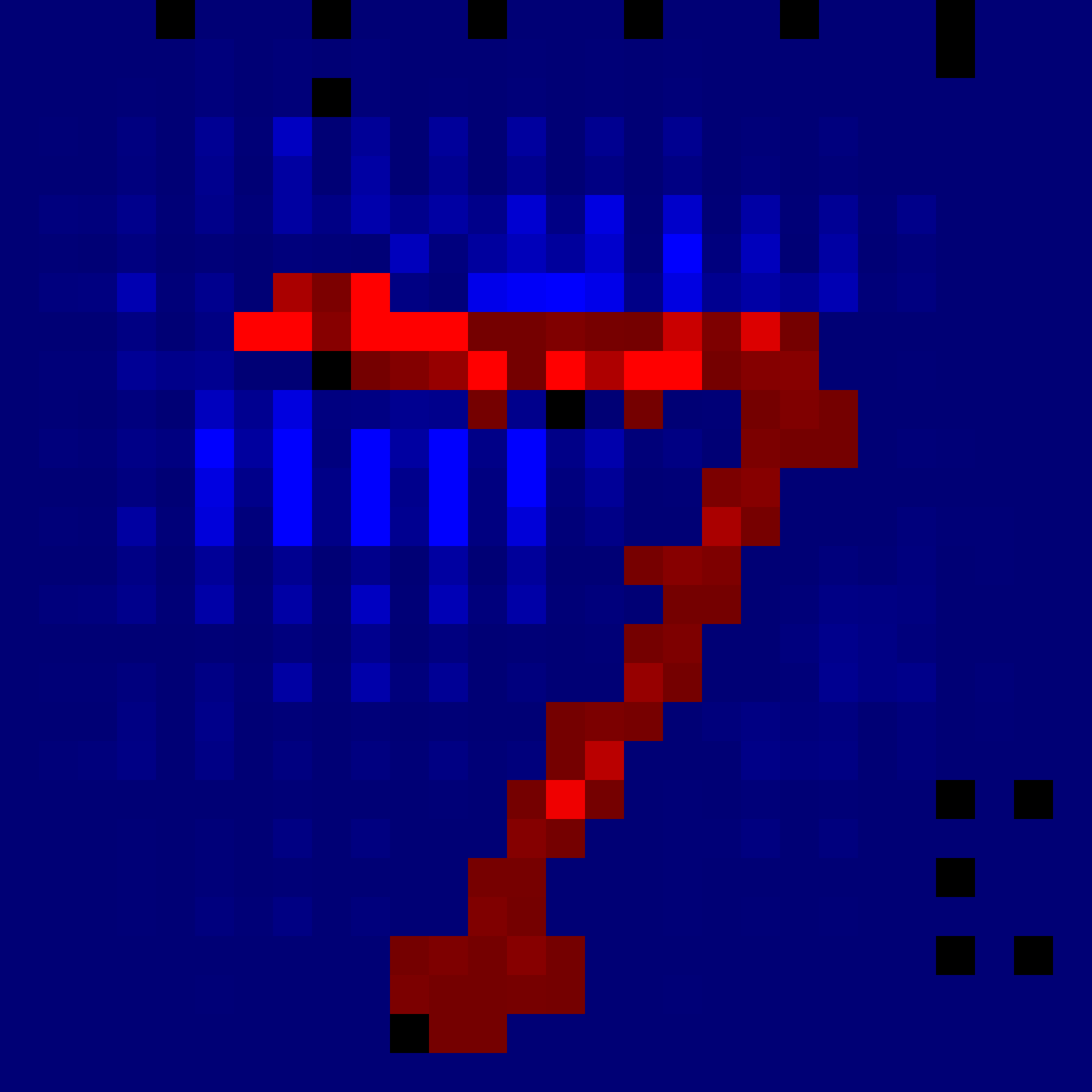}
    \end{subfigure}
    \begin{subfigure}[b]{0.09\textwidth}
        \centering
        \includegraphics[width=\textwidth]{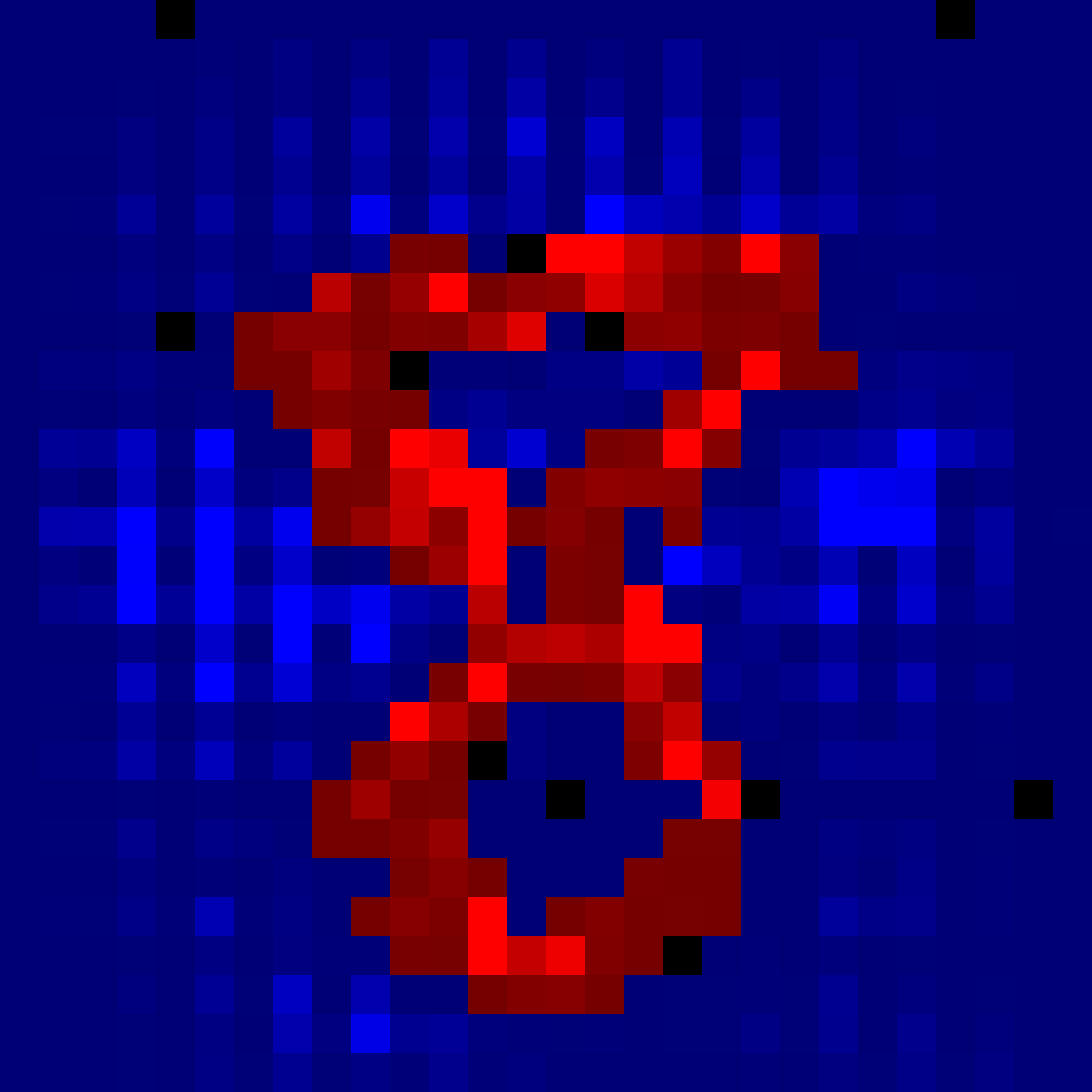}
    \end{subfigure}
    \begin{subfigure}[b]{0.09\textwidth}
        \centering
        \includegraphics[width=\textwidth]{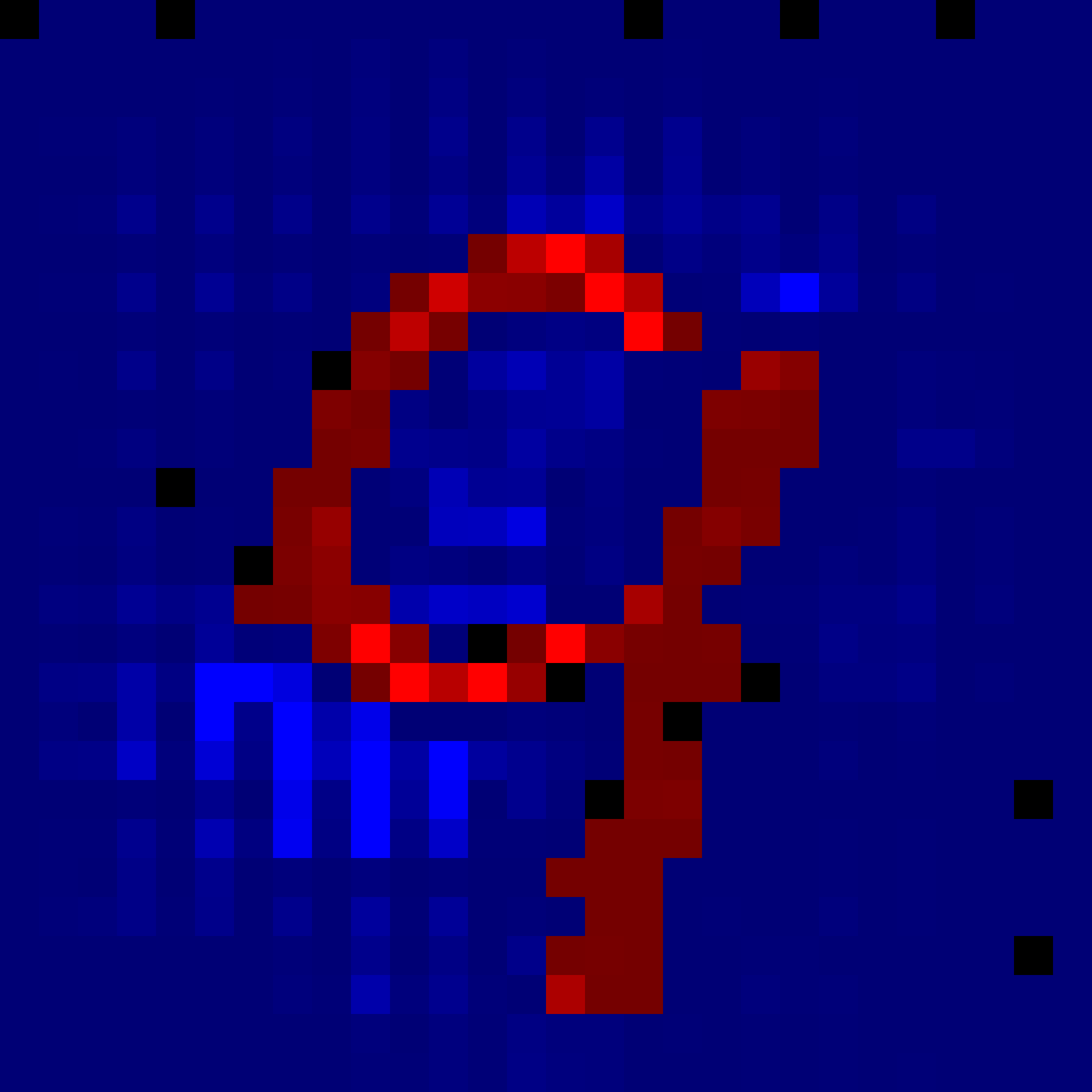}
    \end{subfigure}

    \vspace{0.5cm}

    \begin{subfigure}[b]{0.09\textwidth}
        \centering
        \includegraphics[width=\textwidth]{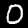}
        \caption*{\scriptsize Label 0}
    \end{subfigure}
    \begin{subfigure}[b]{0.09\textwidth}
        \centering
        \includegraphics[width=\textwidth]{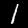}
        \caption*{\scriptsize Label 1}
    \end{subfigure}
    \begin{subfigure}[b]{0.09\textwidth}
        \centering
        \includegraphics[width=\textwidth]{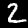}
        \caption*{\scriptsize Label 2}
    \end{subfigure}
    \begin{subfigure}[b]{0.09\textwidth}
        \centering
        \includegraphics[width=\textwidth]{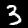}
        \caption*{\scriptsize Label 3}
    \end{subfigure}
    \begin{subfigure}[b]{0.09\textwidth}
        \centering
        \includegraphics[width=\textwidth]{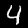}
        \caption*{\scriptsize Label 4}
    \end{subfigure}
    \begin{subfigure}[b]{0.09\textwidth}
        \centering
        \includegraphics[width=\textwidth]{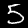}
        \caption*{\scriptsize Label 5}
    \end{subfigure}
    \begin{subfigure}[b]{0.09\textwidth}
        \centering
        \includegraphics[width=\textwidth]{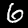}
        \caption*{\scriptsize Label 6}
    \end{subfigure}
    \begin{subfigure}[b]{0.09\textwidth}
        \centering
        \includegraphics[width=\textwidth]{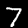}
        \caption*{\scriptsize Label 7}
    \end{subfigure}
    \begin{subfigure}[b]{0.09\textwidth}
        \centering
        \includegraphics[width=\textwidth]{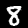}
        \caption*{\scriptsize Label 8}
    \end{subfigure}
    \begin{subfigure}[b]{0.09\textwidth}
        \centering
        \includegraphics[width=\textwidth]{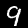}
        \caption*{\scriptsize Label 9}
    \end{subfigure}
    \caption{Input gradient visualizations of the \texttt{cnn\_maxout} IC model of ten digits with respect to their ground truth label. Top row: saliency maps. Bottom row: original images for each label. We colored positive and negative split-stream gradient values above threshold 0.2 red and blue respectively, where we colorized the ranking of the absolute values with an exponential decay.}

    \label{fig:mnist_ic_saliency}
\end{figure}

%% file: appendix/maxout_networks.tex
\section{Generalization of the Split to Multiplicative Maxout Networks} \label{app:splitting_multiplicative_maxout_networks}

In this chapter we lay the theoretical groundwork to split neural networks that use gating, SE-blocks and the attention mechanism into two convex neural networks.
For full generality let us switch to an abstract neural network model, containing so-called addition, multiplication and input neurons.
Furthermore, we consider rank-$k$ Maxout networks, as they are a generalization of ReLU networks.
For notational convenience we do not use bias, as it can be simply simulated by an additional input neuron of constant activation equal to $1$.
The neurons and their connections, so-called arcs, form a directed, weighted and acyclic graph---we choose to not model neural network layers for conceptual simplicity.

For an addition or multiplication neuron $v$ we denote the set of its input neurons by $\delta^{\text{in}}v$ and its incoming weights by $w^{uv} \in \NN \ (u \in \delta^{\text{in}}v)$.
The weights of incoming archs are $\mathbb{R}^k$-vectors in case of addition neurons and natural numbers in case of multiplication neurons.
Any neuron $v$ has a one-dimensional activation $a^v \in \RR$. In case of input neurons we consider it constant. In case of a multiplication neuron it calculates as
\begin{align}
    a^v &= \text{max}_{i \in [k]} z^v_i, \ \text{where the preactivations are given by} \label{eq:addition_neuron_activation}\\
    z^v_i &= \sum_{u \in \delta^{\text{in}}v} w_i^{uv} \cdot a^u. \label{eq:addition_neuron_input_activation}
\end{align}

The activation of a multiplication neuron calculates as
\begin{align}
    a^v &= \prod_{u \in \delta^{\text{in}}v} \left( a^u \right) ^ {w^{uv}}. \label{eq:multiplication_neuron_activation}
\end{align}
We remark that we restrict input weights of multiplication neurons to natural numbers, as we do not want to handle division by zero or complex outputs.
We do not use an activation function for the multiplication nodes, as one can combine multiplication neurons with addition neurons and obtain the same behaviour.

We call a network made of such multiplication and addition neurons a rank-$k$ Maxout multiplication network.
We call such a multiplication network \emph{monotone} in case all its weights are non-negative and \emph{input-convex} in case all weights are non-negative, except weights of arcs incident to input neurons.
The following Theorem is reminiscent of the maximum representation of the support function in \cref{eq:support_function}.
Recall that a posynomial is a polynomial with only positive coefficients.

\begin{theorem} \label{thm:monotone_maxout_main}
    An input-convex rank-$k$ Maxout multiplication network represents the maximum over a finite set of natural-exponent polynomials in the input variables.
    A monotone rank-$k$ Maxout multiplication network represents the maximum over a finite set of natural-exponent posynomials in the input variables.
\end{theorem}

\begin{proofref}{Theorem}{thm:monotone_maxout_main}
    We only show the claim about the monotone network, as the claim about an input-convex model follows analogously. Let $N$ be a monotone rank-$k$ Maxout multiplication network. It is an acyclic graph of input, addition and multiplication neurons.
    We prove the stronger statement that the activation $a^v$ of each neuron is the maximum of posynomials.
    The base cases are trivial as the activation $a^v$ of a single input neuron $v$ is the projection of the activations of all network input neurons on exactly that network input neuron's activation.

    Observe that the set of functions represented by the maximum of posynomials is closed under scaling with positive constants, addition, taking the maximum and multiplication.
    This observation yields the induction steps and therefore closes the proof of \cref{thm:monotone_maxout_main}.
\end{proofref}

For convenience let us define the simple addition neuron that just sums weighted incoming inputs without applying an activation function. It is equivalent to a rank-$k$ Maxout neuron with $k$-times copied weights.
The proof of the following Theorem follows and extends the idea of Theorem in \citeapp{AppHertrich2025}.
Weight matrices are split similar to the approach for ReLU networks in \cref{subsec:relu_mlp-split} into positive and negative parts while the Maxout units are decomposed using the idea introduced for Maxpooling in \cref{subsec:other_components}.
Namely, for $a^+, \ a^- \in \RR^k$ we observe
\begin{align}
    \max_{j\in[k]}(a_j^+ - a_j^-) = \max_{j\in[k]}\Bigl\{a_j^+ + \sum_{i\neq j} a_i^- \Bigr\} - \sum_{i} a_i^-.
\end{align}
We split multiplication neurons by multiplying out the differences of the activations of positive and negative input neurons respectively.

\begin{theorem} \label{thm:main}
    Let $f: \ \RR^d \longrightarrow \RR$ be represented by a rank-$k$ Maxout multiplication network.
    Then it can be written as the difference $f = g - h$ of two convex functions $g$ and $h$ that are both representable with an input-convex rank-$k$ Maxout multiplication network.
    In case that the input count to each multiplication neuron is bounded, the neuron counts of the monotone and input-convex networks are linear in the neuron count of the original network.
\end{theorem}

\begin{proofref}{Theorem}{thm:main}

    Let $N_0$ be the Maxout multiplication network representing $f$ and let
    $$
    v_1, v_2, \dots, v_s
    $$
    be a topological order of the neurons of $N_0$.

    We construct a sequence of equivalent neural networks $N_1, \dots, N_{s-1}$ and the final decomposition network $N_s$, where in each step we split a neuron into multiple neurons including $v^+, v^-$ such that all new neurons have only positive incoming weights.
    Notice that in the intermediate networks we also make use of helping labels on input weights to multiplication neurons, changing their functionality.
    After the algorithm, in $N_s$ all neurons are going to be split, there are no helping labels and no negative weights left. 
    Setting the network outputs to $g = a^{(v^{+}_s)}$ and $h = a^{(v^-_s)}$ we obtain the final decomposition.

    \textbf{case} $v \coloneqq v_p$ is an input neuron.
        We want to ensure that all the inputs to addition and multiplication neurons are positive.
        For this reason, we add addition neurons $v^+$ and $v^-$ and connect them only to $v$, where we set all weight entries to zero, except one weight entry that is set to $1$ in case of $v^+$ and $-1$ in case of $v^-$.
        We note, that this simulates two ReLU activations -- $v^+$ transmits positive inputs and $v^-$ transmits negative inputs.

        Then, we connect $v^+$ to all outputs of $v$ with the original weights and $v^-$ to all outputs of $v$ with signs flipped for weights that go to addition neurons.
        Lastly, the output weights of $v^+$ and $v^-$ get a unique label, changing how multiplication nodes process their input. Namely, they consider $(a^{v^+} - a^{v^-})$ as a single input.

    \textbf{case} $v$ is an addition neuron.
        Let us consider the incoming weights $w^{uv}_i$ of $v \coloneqq v_p$. We split them into a positive and negative part, that is,
        $$
        w^{uv}_i = a^{uv}_i - b^{uv}_i, \quad \text{with } a^{uv}_i = \text{max} \{w^{uv}_i, 0\}, \quad b^{uv}_i = \text{max} \{-w^{uv}_i, 0\}
        $$
        
        Observe that
        \begin{align}
                z^v = \max_{i = 1, \dots, k} \left\{ \sum_{u \in \delta^{\text{in}} v} \left(a^{uv}_i + \sum_{j \neq i} b^{uv}_j \right) z_u \right\} - \sum_{u \in \delta^{\text{in}} v} \sum_{j} b^{uv}_j z_u \label{eq:addition_correct}.
        \end{align}
        
        Let us split $v$ into $v^+$ and $v^-$ where $v^+$ has input weights $w_i^{uv^+} \coloneqq a^{uv}_i + \sum_{j \neq i} b^{uv}_j$ and the same output weights as $v$.
        $v^-$ is a simple addition neuron and has input weights $w^{uv^-} \coloneqq\sum_{j=1}^k b^{uv}_j$. Furthermore, it has the same output weights as $v$, only with a flipped sign towards addition neurons.
        Again, the output weights of $v^+$ and $v^-$ get a unique label, changing how multiplication nodes process their input.

    \textbf{case} $v \coloneqq v_p$ is a multiplication neuron.
        We recall the following extension of the binomial Theorem.
        \begin{align}
            a^v = \prod_{u \in \delta^{\text{in}}v} (a^{u^+} - a^{u^-}) ^ {w^{uv}}
                = \sum_{\underset{\forall u \in  \delta^{\text{in}}v: \ 0 \leq q_u \leq w_{uv}}{q \in \mathbb{N}^{\delta^{\text{in}}v}}} \sigma_q \prod_{ u \in \delta^{\text{in}}v } \binom{w_{uv}}{q_u} (a^{u^+}) ^ {(w_{uv}-q_u)} \cdot (a^{u^-}) ^ {q_u} \label{eq:multiplication_correct},
        \end{align}
        where $\sigma_q \coloneqq 2 \cdot \mathbf{1} \{ \sum_{u \in \delta^{\text{in}}v} q_u \ \text{is even} \} - 1$.

        Let us introduce one multiplication neuron for each of the summands.
        To avoid exponential neuron explosion, we introduce two simple addition neurons $v^+$ and $v^-$ that have multiplication neurons as input with weight $1$, one for all the positive and one for all the negative summands, respectively.
        Now, similarly as before, we connect $v^+$ to all outputs of $v$ with the original weights and $v^-$ to all outputs of $v$ with sign flipped weights towards addition neurons. Moreover, we label the outputs of $v^+$ and $v^-$ uniquely, so they are correctly processed by multiplicative neurons.

    \begin{claim}[Correctness] \label{claim:main_correct}
        For any $0 \leq j < s, \ N_j$ represents $f$ and in $N_s$ the positive and negative output neuron represent $f$ by their difference, namely $f = g - h$.
    \end{claim}

    \begin{proofref}{Claim}{claim:main_correct}
        We need to show that our input, addition and multiplication neuron transformations are not changing the function represented by the network.
        Then it is also easy to see that the claim $f = g - h$ holds.
        Indeed, the split of the input neurons into positive and negative input combined with the sign flip of weights towards addition neurons as well as the labels towards multiplication neurons does not change the function represented by the network.
        The same holds for addition neurons, where we use \cref{eq:addition_correct}. Lastly, using \cref{eq:multiplication_correct} it is easy to see that also the multiplication neuron transformation does not change the function represented by the network.
    \end{proofref}

    Let $g'$ and $h'$ be the functions represented by positive and negative output neuron respectively, where we consider the positive and negative neurons $V'_{\text{in}}$ created when splitting the network input neurons as the new input neurons.
    Let us call these networks $N_{g'}$ and $N_{h'}$.

    As by \cref{thm:monotone_maxout_main} $g'$ and $h'$ are maxima of posynomials and therefore monotone and convex on positive input, and as the activations of $V'_{\text{in}}$ in $N_s$ are convex in the network input, we conclude, that $g$ and $h$ are convex.

    Further remark, that indeed the neural networks representing $g$ and $h$ are input-convex rank-$k$ Maxout multiplication networks.
    Now, together with Claim \ref{claim:main_correct} this concludes the proof of \cref{thm:main}.
\end{proofref}

Finally, let us remark that by not splitting the input neurons as done in the proof of \cref{thm:main}, one obtains two monotone networks in the split. However, these networks represent functions only monotone on positive input, as posynomials are only monotone on positive input.
As a fix, one could simply manually add a big constant $C$ to all network inputs and circumvent by shifting the bias of the original model. This way, the split networks represent monotone and convex functions on $\mathbb{R}_{\geq-C}^d$.

%% file: appendix/proofs.tex
\section{Further Proofs}

In this appendix we rigorously define and analyze the shifting procedures proposed in \cref{sec:method} and provide the proofs to Theorems referred to in the main paper. First, in \cref{app:fundamental} we give proofs about monotonicity and convexity of weight-restricted neural networks and about the correctness of the proposed DC decomposition algorithm for ReLU networks.
Secondly, in \cref{app:shift_forward} we provide some intuition for the shift procedure in the forward pass and use these insights to illuminate the relationship between the different local sensitivities defined in \cref{sec:backward}.
Lastly, in \cref{app:closed_formulas} we rigorously define the shifting procedure in the backward pass and develop closed formulas for the stabilized gradients of the split-pair. The latter demonstrate how our shifting procedure effectively keeps the absolute values of the stabilized split-pair gradients within a numerically stable range and how it can potentially be used to emphasize the splitting information of different layers in the calculated gradients.

\subsection{Proof of the Fundamental Theorems} \label{app:fundamental}

In this section we give the simple proofs of Theorems referred to in \cref{sec:method} for completeness.
The following Theorem is already discussed in \citetapp{AppAmos2017}.
\begin{theorem} \label{thm:convexity_monotonicity}
    An input-convex neural network represents a convex function. A monotone neural network represents a monotone function.
\end{theorem}

\begin{proofref}{Theorem}{thm:convexity_monotonicity}
    Let $N_{\text{mon}}$ and $N_{\text{ic}}$ be a monotone and input-convex neural network respectively. Let us denote their activation function by $\text{act}_{\text{mon}}$ and $\text{act}_{\text{ic}}$.
    For convenience we denote both their weights by $W^{(1)},\dots,W^{(L)}$.
    We want to prove the monotonicity and convexity of $f_{\text{mon}}^{(l)}$ and $f_{\text{ic}}^{(l)}$ by induction on $l=L,\dots,0$.

    \textbf{base} $l=L$.
        Since $f_{\text{mon}}^{(l)}$ and $f_{\text{ic}}^{(l)}$ are the activation functions themselves, the induction base case is trivial.

    \textbf{step} Let $0 \leq l < L$.
        \begin{align}
            f_{\text{mon}}^{(l)}(x) &= f_{\text{mon}}^{(l+1)}(x) \left( \text{act}_{\text{mon}}\left( W^{(l+1)} x \right) \right), \\
            f_{\text{ic}}^{(l)}(x) &= f_{\text{ic}}^{(l+1)}(x) \left( \text{act}_{\text{ic}}\left( W^{(l+1)} x \right) \right).
        \end{align}

        It is easy to see that $f_{\text{mon}}^{(l)}$ is monotone as the composition of monotone functions.
        Similarly, it is easy to see that $f_{\text{ic}}^{(l)}(x)$ is convex as the composition of a linear, and two convex and monotone functions.
        We remark that the monotonicity of $\text{act}_{\text{ic}}$ and $f_{\text{ic}}^{(l+1)}$ is necessary - the composition of two convex functions is not convex in general.
        However, as we can see by the inductive argument, we do not require $f_{\text{ic}}^{(0)}$ to be monotone, so we can allow negative weights in the first layer for the argument to run through.
\end{proofref}

We remark that monotonicity of the activation function for the input-convex neural networks is indeed necessary for its convexity.

\begin{remark}[Bias] \label{rem:bias}
    It is easy to see that we can add bias -- that is not restricted to be non-negative -- to a monotone or input-convex neural network without breaking the properties guaranteed by Theorem \ref{thm:convexity_monotonicity}.
    For this insight let us briefly view bias as a shift of the activation function. Indeed, shifting the activation function neither breaks its convexity, its monotonicity nor its CPWL property.
\end{remark}

Now let us prove the correctness of the DC decomposition as described in \cref{sec:method}.

\begin{proofref}{Theorem}{thm:split_correct}
    We prove the Theorem by induction on $l=L,\dots,0$.
    
    \textbf{base} $l=L$.
        Observe $g^{(L)}\left( x^+, x^- \right) = \text{max}\left\{ x^+, x^- \right\}$ and $h^{(L)}\left( x^+, x^- \right) = x^-$. We can verify correctness by
        \begin{align}
            f^{(L)}\left( x^+ - x^- \right)
                = \text{ReLU}\left( x^+ - x^- \right)
                = \text{max}\left\{ x^+, x^- \right\} - x^-
                = g^{(L)}\left( x^+ \right) - h^{(L)}\left( x^- \right).
        \end{align}
    
    \textbf{step} Let $0 \leq l < L$.
        Observe that $g^{(l)}\left( x^+, x^- \right) = g^{(l+1)}\left( a^{(l+1, +)}, a^{(l+1, -)} \right)$, where
        \begin{align}
            a^{(l+1, +)} &= \text{max}\left\{ W^{(l+1, +)} x^+ + W^{(l+1, -)} x^-, W^{(l+1, +)} x^- + W^{(l+1, -)} x^+ \right\}, \ \text{and} \\
            a^{(l+1, -)} &= W^{(l+1, +)} x^- + W^{(l+1, -)} x^+.
        \end{align}
        Using the induction hypothesis we obtain
        \begin{align}
            f^{(l)}\left( x^+ - x^- \right)
            &= f^{(l+1)} \left( \text{ReLU}\left( \left( W^{(l+1, +)} - W^{(l+1, -)} \right) \left( x^+ - x^- \right) \right) \right) \\
            &= f^{(l+1)} \left( \text{max}\left\{ W^{(l+1, +)} x^+ + W^{(l+1, -)} x^-, W^{(l+1, -)} x^- + W^{(l+1, +)} x^- \right\} - \left( W^{(l+1, -)} x^+ +  W^{(l+1, +)} x^- \right) \right) \\
            &= g^{(l+1)} \left( a^{(l+1,+)}, a^{(l+1,-)} \right) - h^{(l+1)}\left( a^{(l+1,+)}, a^{(l+1,-)} \right) \\
            &= g^{(l)}\left( x^+, x^- \right) - h^{(l)}\left( x^+, x^- \right).
        \end{align}

        This completes the proof of \cref{thm:split_correct}.
\end{proofref}

\subsection{Shifting in the forward pass} \label{app:shift_forward}

Shifting the activation of the positive and negative neurons in layer $l$ in the forward pass by the same amount $c$, shifts both the output of the positive and the negative stream by the same amount, which is linear in $c$. For notational convenience, let us define the product of the absolute weight matrices of layer $L,\dots,l+1$ by
\begin{align}
    W_{\text{abs}}^{[l, \rightarrow]} \coloneqq \prod_{ j = L, \dots, l+1} \left|W^{(j)}\right|. \label{eq:w_abs}
\end{align}

\begin{theorem} \label{thm:shift_forward}
    For any $0 \leq l \leq L$ and any $c, \ a^{(+)}, \ a^{(-)} \in \RR ^ {d^{(l)}}$ we have
    \begin{align}
        g^{(l)}\!\left( \begin{bmatrix} a^{(+)} + c \\ a^{(-)} + c \end{bmatrix} \right)
            &= g^{(l)}\!\left( \begin{bmatrix} a^{(+)} \\ a^{(-)} \end{bmatrix} \right)
            + W_{\text{abs}}^{[l, \rightarrow]} \cdot c \\
        h^{(l)}\!\left( \begin{bmatrix} a^{(+)} + c \\ a^{(-)} + c \end{bmatrix} \right)
            &= h^{(l)}\!\left( \begin{bmatrix} a^{(+)} \\ a^{(-)} \end{bmatrix} \right)
            + W_{\text{abs}}^{[l, \rightarrow]} \cdot c.
    \end{align}
\end{theorem}

\begin{proofref}{Theorem}{thm:shift_forward}
    We show the claim for $g^{(l)}$ by induction on $l$. It follows analogously for $h^{(l)}$.

    The base case $l = L$ is trivial, as $W_{\text{abs}}^{[L, \rightarrow]}$ is an empty product and therefore the identity.
    For the step, let $0 \leq l < L$.
    As the positive and negative neurons in layer $l+1$ are connected to the previous layer with the same weights,
    only that $W^{(l+1, +)}$ and $W^{(l+1, -)}$ are flipped, uniformly shifting the positive and negative activations of layer $l$ results in a uniform shift in layer $l+1$, see \cref{fig:splitstreams} for visualization.
    \[
    a^{(l+1, +)} 
      = \max\!\left\{
        W^{(l+1, +)} a^{(+)} + W^{(l+1, -)} a^{(-)}, \,
        W^{(l+1, -)} a^{(+)} + W^{(l+1, +)} a^{(-)}
      \right\} 
      + \left|W^{(l+1)}\right| c
    \]
    and
    \[
    a^{(l+1, -)} 
      = W^{(l+1, -)} a^{(+)} + W^{(l+1, +)} a^{(-)} 
      + \left|W^{(l+1)}\right| \cdot c,
    \]
    using the induction hypothesis we obtain
    \begin{align}
        g^{(l)}\!\left( \begin{bmatrix} a^{(+)} + c \\ a^{(-)} + c \end{bmatrix} \right)
            = g^{(l+1)}\!\left( \begin{bmatrix} a^{(l+1, +)} \\ a^{(l+1, -)} \end{bmatrix} \right)
            = g^{(l)}\!\left( \begin{bmatrix} a^{(+)} \\ a^{(-)} \end{bmatrix} \right) 
               + W_{\text{abs}}^{[l, \rightarrow]} c,
    \end{align}
    which completes the proof of \cref{thm:shift_forward}.
\end{proofref}

This observation has consequences on the connection of the local sensitivities $\delta^{(l, +, g)}$ and $\delta^{(l, -, g)}$ as well as $\delta^{(l, +, h)}$ and $\delta^{(l, -, h)}$.

\begin{theorem} \label{thm:delta_pos_neg_connection}
    \begin{align}
        \delta^{(l,+,g)} &= W_{\text{abs}}^{[l, \rightarrow]} - \delta^{(l,-,g)}, \\
        \delta^{(l,+,h)} &= W_{\text{abs}}^{[l, \rightarrow]} - \delta^{(l,-,h)}
    \end{align}
\end{theorem}

\begin{proofref}{Theorem}{thm:delta_pos_neg_connection}
    For $j \in [d^{(l)}]$ \cref{thm:shift_forward} yields
    \begin{align}
        \delta_j^{(l,+,g)}
            = \text{lim}_{\epsilon \rightarrow 0} \frac{g^{(l)}(a^{(l,+)} + \epsilon \cdot e_j, a^{(l,-)})}{\epsilon}
            = \text{lim}_{\epsilon \rightarrow 0} \frac{g^{(l)}(a^{(l,+)}, a^{(l,-)} - \epsilon \cdot e_j) + \epsilon W_{\text{abs}}^{[l, \rightarrow]} e_j}{\epsilon}
            = W_{\text{abs}}^{[l, \rightarrow]} e_j - \delta_j^{(l,-,g)}
    \end{align}
    The claim about $\delta^{(l,+,h)}$ follows analogously, closing the proof of \cref{thm:delta_pos_neg_connection}.
\end{proofref}

\subsection{Stabilizing the split-pair backward pass} \label{app:closed_formulas}

\subsubsection{The algorithm}

In order to stay in numerically stable regimes we shift the local sensitivities during the backward pass in each layer uniformly by $\text{shift}_\delta^{(l, g)}$ and $\text{shift}_\delta^{(l, h)}$. We compute these modified saliency tensors $\delta_{\text{shift}}^{(l, +, g)}, \delta_{\text{shift}}^{(l, -, g)}, \delta_{\text{shift}}^{(l, +, h)}, \delta_{\text{shift}}^{(l, -, h)}$ in an iterative way: using the regular gradient formulas, and the already modified local sensitivities inflowing from subsequent network layers, we compute intermediate local sensitivities $\delta_{\text{intermediate}}^{(l,+,g)}, \ \delta_{\text{intermediate}}^{(l,-,g)}, \ \delta_{\text{intermediate}}^{(l,+,h)}, \ \delta_{\text{intermediate}}^{(l,-,h)}$. In a second step, we calculate the modifed version by
\begin{align}
    \delta_{\text{shift}}^{(l, +, g)} &\coloneqq \delta_{\text{intermediate}}^{(l, +, g)} - \text{shift}_\delta^{(l, g)} \\
    \delta_{\text{shift}}^{(l, -, g)} &\coloneqq \delta_{\text{intermediate}}^{(l, -, g)} - \text{shift}_\delta^{(l, g)} \\
    \delta_{\text{shift}}^{(l, +, h)} &\coloneqq \delta_{\text{intermediate}}^{(l, +, h)} - \text{shift}_\delta^{(l, h)} \\
    \delta_{\text{shift}}^{(l, -, h)} &\coloneqq \delta_{\text{intermediate}}^{(l, -, h)} - \text{shift}_\delta^{(l, h)}
\end{align}

For the shifting we use
\begin{align}
    \text{shift}_\delta^{(l, g)} & \coloneqq \alpha^{(l)} \cdot \left( \delta_{\text{intermediate}}^{(l, +, g)} + \delta_{\text{intermediate}}^{(l, -, g)} \right), \\
    \text{shift}_\delta^{(l, h)} & \coloneqq \alpha^{(l)} \cdot \left( \delta_{\text{intermediate}}^{(l, +, h)} + \delta_{\text{intermediate}}^{(l, -, h)} \right).
\end{align}

It turns out, that indeed these shifts are just $\alpha$-multiples of the matrices $W_{\text{abs}}^{[l, \rightarrow]}$ defined in \cref{eq:w_abs}.
We provide the closed formulas of the stabilized split-pair gradients at the end of the next section in \cref{thm:closed_formula_shift}.

\subsubsection{Closed formulas for the stabilized split-pair gradients}

In standard backpropagation for $0 \leq l < L$ the chain rule yields
\begin{align}
    \delta^{(l)}
        \coloneqq \frac{\partial f^{(l)}}{\partial a^{(l)}}
        = \frac{\partial f^{(l)}}{\partial a^{(l+1)}} \frac{\partial a^{(l+1)}}{\partial z^{(l+1)}} \frac{\partial z^{(l+1)}}{\partial a^{(l)}}
        = \delta^{(l+1)} \text{diag} \left( \mathbb{1}\left\{ z^{(l+1)} \geq 0 \right\} \right) W^{(l+1)}. \label{eq:std_bckprop}
\end{align}

In order to define the derivatives of $g$ and $h$ with respect to layer activations of the original network, we need the following helper function.
\begin{align}
    \tilde{g}^{(l)} :\ \RR^{d^{(l)}} &\longrightarrow \RR, \\
     a^{(l)} &\longmapsto g^{(l)}\!\left( \begin{bmatrix} \frac{a^{(l)}}{2} \\ -\frac{a^{(l)}}{2} \end{bmatrix} \right).
\end{align}
Defining $\tilde{h}^{(l)}$ similarily we now can rigorously define the split-streams gradients with respect to a unified input vector.
\begin{align}
    \delta^{(l, g)} \coloneqq \frac{\partial \tilde{g}^{(l)}}{\partial a^{(l)}}, \qquad
    \delta^{(l, h)} \coloneqq \frac{\partial \tilde{h}^{(l)}}{\partial a^{(l)}}, \qquad
\end{align}

To calculate $\delta^{(l, g)}$ directly, we use the chain rule:
\begin{align}
    \delta^{(l, g)}
        = \frac{\partial \tilde{g}^{(l)}}{\partial a^{(l)}} 
        = \frac{\partial g^{(l)}}{\partial a^{(l,+)}} \frac{\partial a^{(l,+)}}{\partial a^{(l)}} + \frac{\partial g^{(l)}}{\partial a^{(l,-)}} \frac{\partial a^{(l,-)}}{\partial a^{(l)}}
        = \frac{\delta^{(l,+,g)} - \delta^{(l,-,g)}}{2}.
\end{align}
Similarly, we obtain
\begin{align}
\delta^{(l, h)} = \frac{\delta^{(l,+,h)} - \delta^{(l,-,h)}}{2}.
\end{align}

We want to find a closed formula for all the derivatives. For this purpose let us define
\begin{alignat}{2}
    m^{(l,+)} &\;\coloneqq\; \mathbb{1}\!\left\{\,z^{(l,+)} \ge z^{(l,-)}\,\right\} = \mathbb{1}\!\left\{\,z^{(l)} \ge 0\,\right\},
    &\qquad M^{(l,+)} &\;\coloneqq\; \text{diag}\!\left( m^{(l,+)} \right) \\[6pt]
    m^{(l,-)} &\;\coloneqq\; \mathbb{1}\!\left\{\,z^{(l,-)} > z^{(l,+)}\,\right\} = \mathbb{1}\!\left\{\,z^{(l)} < 0\,\right\},
    &\qquad M^{(l,-)} &\;\coloneqq\; \text{diag}\!\left( m^{(l,-)} \right)
\end{alignat}
understood elementwise. For ease of notation we define
\begin{align}
    W^{(l, x)} &\coloneqq M^{(l, +)} W^{(l)}
\end{align}

It is easy to see from \cref{eq:std_bckprop} that $\delta^{(l)} = \prod_{j = L,\dots,l+1} W^{(j,x)}$ and analogously for any $0 \leq l_1 < l_2 \leq L$
\begin{align}
    \frac{\partial z^{(l_2)}}{\partial a^{(l_1)}} = \prod_{j = l_2,\dots,(l_1 + 1)} W^{(j, x)} \label{eq:general_derivative_product}
\end{align}
This formula also holds for $l_1 = l_2$ in case we interpret $\frac{\partial z^{(l)}}{\partial a^{(l)}}$ as the derivative of a network where input and output layer coincide; in this case the derivative is the identity.

Before looking at the shifted gradients let us develop some understanding of the plain gradients of the split-streams.
In a first step the proof of the following Theorem calculates recursive formulas for those gradients and in a second step unfolds them into large summation formulas.

\begin{theorem} \label{thm:closed_formula}
    \begin{align}
        \delta^{(l, g)}
            &= \frac{1}{2} \cdot \left[ \delta^{(l)}- \sum_{l \leq j < L} W_{\text{abs}}^{[j+1, \rightarrow]} \cdot \left( M^{(j+1,-)}W^{(j+1)} \right) \cdot \prod_{i = (j-1),\dots, l} W^{(i+1, x)} \right], \\
        \delta^{(l, h)}
            &= -\frac{1}{2} \cdot \left[ \delta^{(l)} + \sum_{l \leq j < L} W_{\text{abs}}^{[j+1, \rightarrow]} \cdot \left( M^{(j+1,-)}W^{(j+1)} \right) \cdot \prod_{i = (j-1),\dots, l} W^{(i+1, x)} \right].
    \end{align}
\end{theorem}

\begin{proofref}{Theorem}{thm:closed_formula}  
    First, let us find recursive formulations of $\delta^{(l, g)}$ and $\delta^{(l, h)}$.
    Let us calculate
    \begin{alignat}{5}
        \frac{\partial a^{(l+1,+)}}{\partial a^{(l,+)}}
            &={} \frac{\partial a^{(l+1,+)}}{\partial z^{(l+1,+)}} 
                   \frac{\partial z^{(l+1,+)}}{\partial a^{(l,+)}}
            &+{}& \frac{\partial a^{(l+1,+)}}{\partial z^{(l+1,-)}} 
                   \frac{\partial z^{(l+1,-)}}{\partial a^{(l,+)}}
            &={}& M^{(l+1,+)} W^{(l+1,+)} + M^{(l+1,-)} W^{(l+1,-)}, \\[6pt]
        \frac{\partial a^{(l+1,-)}}{\partial a^{(l,+)}}
            &={} \frac{\partial a^{(l+1,-)}}{\partial z^{(l+1,+)}} 
                   \frac{\partial z^{(l+1,+)}}{\partial a^{(l,+)}}
            &+{}& \frac{\partial a^{(l+1,-)}}{\partial z^{(l+1,-)}} 
                   \frac{\partial z^{(l+1,-)}}{\partial a^{(l,+)}}
            &={}& 0 \cdot W^{(l+1,+)} + W^{(l+1,-)}, \\[6pt]
        \frac{\partial a^{(l+1,+)}}{\partial a^{(l,-)}}
            &={} \frac{\partial a^{(l+1,+)}}{\partial z^{(l+1,+)}} 
                   \frac{\partial z^{(l+1,+)}}{\partial a^{(l,-)}}
            &+{}& \frac{\partial a^{(l+1,+)}}{\partial z^{(l+1,-)}} 
                   \frac{\partial z^{(l+1,-)}}{\partial a^{(l,-)}}
            &={}& M^{(l+1,+)} W^{(l+1,-)} + M^{(l+1,-)} W^{(l+1,+)}, \\[6pt]
        \frac{\partial a^{(l+1,-)}}{\partial a^{(l,-)}}
            &={} \frac{\partial a^{(l+1,-)}}{\partial z^{(l+1,+)}} 
                   \frac{\partial z^{(l+1,+)}}{\partial a^{(l,-)}}
            &+{}& \frac{\partial a^{(l+1,-)}}{\partial z^{(l+1,-)}} 
                   \frac{\partial z^{(l+1,-)}}{\partial a^{(l,-)}}
            &={}& 0 \cdot W^{(l+1,-)} + W^{(l+1,+)}.
    \end{alignat}
    
    Again the chain rule yields the iterative formulas
    \begin{align}
        \delta^{(l,+,g)}
            &= \frac{\partial g^{(l+1)}}{\partial a^{(l+1,+)}}
                \frac{\partial a^{(l+1,+)}}{\partial a^{(l,+)}}
             + \frac{\partial g^{(l+1)}}{\partial a^{(l+1,-)}}
                \frac{\partial a^{(l+1,-)}}{\partial a^{(l,+)}} \notag\\
            &= \delta^{(l+1,+,g)} \left( M^{(l+1, +)} W^{(l+1, +)} + M^{(l+1, -)} W^{(l+1, -)} \right)
                + \delta^{(l+1,-,g)} W^{(l+1, -)} \label{eq:delta_pg_recursion}
    \end{align}
    and further
    \begin{align}
        \delta^{(l,-,g)}
            &= \frac{\partial g^{(l+1)}}{\partial a^{(l+1,+)}}
               \frac{\partial a^{(l+1,+)}}{\partial a^{(l,-)}}
             + \frac{\partial \hat g^{(l+1)}}{\partial a^{(l+1,-)}}
               \frac{\partial a^{(l+1,-)}}{\partial a^{(l,-)}} \notag\\
            &= \delta^{(l+1,+,g)} \left(M^{(l+1, +)} W^{(l+1, -)} + M^{(l+1, -)} W^{(l+1, +)} \right)
                + \delta^{(l+1,-,g)} W^{(l+1, +)} \label{eq:delta_ng_recursion}.
    \end{align}
    
    With this, we obtain
    \begin{align}
        2 \cdot \delta^{(l, g)}
            &= \delta^{(l,+,g)} - \delta^{(l,-,g)} \notag\\
            &= \delta^{(l+1,+,g)} \left(M^{(l+1, +)} \cdot W^{(l+1)} - M^{(l+1, -)} \cdot W^{(l+1)} \right)
                - \delta^{(l+1,-,g)} W^{(l+1)} \notag\\
            &= \left( \delta^{(l+1,+,g)} - \delta^{(l+1,-,g)} \right) \cdot M^{(l+1, +)} W^{(l+1)}
                - \left( \delta^{(l+1,+,g)} + \delta^{(l+1,-,g)} \right) M^{(l+1, -)} W^{(l+1)} \notag\\
            &= 2 \cdot \delta^{(l+1, g)} \cdot W^{(l+1, x)} - W_{\text{abs}}^{[l+1, \rightarrow]} M^{(l+1, -)} W^{(l+1)} \label{eq:delta_g_recursion},
    \end{align}
    where we have used in the penultimate step that $M^{(l+1, +)} + M^{(l+1, -)} = \mathbb{1}$.
    Similarly, we obtain
    \begin{align}
        2 \cdot \delta^{(l, h)}
            &= 2 \cdot \delta^{(l+1, h)} \cdot W^{(l+1, x)} - W_{\text{abs}}^{[l+1, \rightarrow]} M^{(l+1, -)} W^{(l+1)} \label{eq:delta_h_recursion}.
    \end{align}

    Let us sanity check the recursive formulas \cref{eq:delta_g_recursion} and \cref{eq:delta_h_recursion} by subtracting them.
    We obtain $\delta^{(l, g)} - \delta^{(l, h)} = \left( \delta^{(l+1,g)} -\delta^{(l+1,h)} \right) W^{(l+1, x)}$, which turns out to be just \cref{eq:std_bckprop}.

    Finally, it is easy to obtain the closed formulas by the recursion.
    The induction base $l=L$ is correct, as $\delta^{(L,g)} = \frac{\mathbb{1}}{2} = \frac{\delta^{(L)}}{2}$ and $\delta^{(L,h)} = - \frac{\mathbb{1}}{2} = -\frac{\delta^{(L)}}{2}$.
    For the step, we plug in the induction hypothesis.
    \begin{align}
        \delta^{(l, g)}
            &= \delta^{(l+1, g)} \cdot W^{(l+1, x)} - \frac{1}{2} \cdot W_{\text{abs}}^{[l+1, \rightarrow]} \cdot \left( M^{(l+1,-)}W^{(l+1)} \right) \\
            &= \frac{1}{2} \cdot \left[ \delta^{(l+1)} - \sum_{l+1 \leq j < L} W_{\text{abs}}^{[j+1, \rightarrow]} \cdot \left( M^{(j+1,-)}W^{(j+1)} \right) \cdot \prod_{i = (j-1),\dots, l+1} W^{(i+1, x)} \right] \cdot W^{(l+1, x)}  \\ &\qquad \mathllap{-} \frac{1}{2} \cdot W_{\text{abs}}^{[l+1, \rightarrow]} \cdot \left( M^{(l+1, -)} W^{(l+1)} \right) \\
            &= \frac{1}{2} \cdot \left[ \delta^{(l)} - \sum_{l+1 \leq j < L} W_{\text{abs}}^{[j+1, \rightarrow]} \cdot \left( M^{(j+1,-)}W^{(j+1)} \right) \cdot \prod_{i = (j-1),\dots, l} W^{(i+1), x} \right] \\ &\qquad \mathllap{-} \frac{1}{2} \cdot W_{\text{abs}}^{[l+1, \rightarrow]} \cdot \left( M^{(l+1, -)} W^{(l+1)} \right) \\
            &= \frac{1}{2} \cdot \left[ \delta^{(l)} - \sum_{l \leq j < L} W_{\text{abs}}^{[j+1, \rightarrow]} \cdot \left( M^{(j+1,-)}W^{(j+1)} \right) \cdot \prod_{i = (j-1),\dots, l} W^{(i+1, x)} \right]
    \end{align}
    One can prove the claim about $\delta^{(l, h)}$ analogously, where the only difference is a sign flip in the induction base case.
    This concludes the proof of \cref{thm:closed_formula}.
\end{proofref}

Finally, we want to give a mathematical interpretation of what the shifted gradients actually calculate. Let us introduce the notation
\begin{align}
    \delta_{\text{shift}}^g
        &\coloneqq \frac{\delta_{\text{shift}}^{(0, +, g)} -\delta_{\text{shift}}^{(0, -, g)}}{2}, \label{eq:stable_def_pos}\\
    \delta_{\text{shift}}^h
        &\coloneqq \frac{\delta_{\text{shift}}^{(0, +, h)} -\delta_{\text{shift}}^{(0, -, h)}}{2}\label{eq:stable_def_neg}.
\end{align}
The closed formulas for $\delta_{\text{shift}}^g$ and $\delta_{\text{shift}}^h$ have an interesting interpretation using the following network.
Let $Q[l]$ be the function represented by the original network where all ReLU activations from the $l$-th layer onwards are replaced with a $\text{min}\left\{ x, 0 \right\}$
and the weights in layers $l+1,\dots,L$ are replaced by their absolute values.
Indeed, we observe that the summands arising in the formulas of \cref{thm:closed_formula} turn out to be just the respective gradients of $Q[l]$.
\begin{align}
    \frac{\partial Q[l]}{\partial x}
        = W_{\text{abs}}^{[l, \rightarrow]} \cdot \left( M^{(l,-)}W^{(l)} \right) \cdot \prod_{i = (l-1),\dots,1} W^{(i,x)} \label{eq:special_network_derivative},
\end{align}
where the first product term corresponds to the derivative of the last layers with only positive weights and $\text{min}\left\{ x,0 \right\}$-activation. Note that since the output of the $l$-th layer is negative and all the following weights are positive, all upcoming activations are negative. This means that the minimum activation functions let all activations pass unchanged.

\begin{theorem}[Closed formulas for shifted saliency maps] \label{thm:closed_formula_shift}
    With $\overline{\alpha}_j \coloneqq \prod_{j \leq i \leq L} \left( 1 - 2 \alpha^{(i)} \right)$ we obtain
    \begin{align}
        \delta_{\text{shift}}^g &= \frac{1}{2} \left[ \delta - \sum_{0 \leq j < L} \ \overline{\alpha}_j \frac{\partial Q[j+1]}{\partial x} \right], \label{eq:main_shifted_grad_pos}\\
        \delta_{\text{shift}}^h &= - \frac{1}{2} \left[ \delta + \sum_{0 \leq j < L} \ \overline{\alpha}_j  \frac{\partial Q[j+1]}{\partial x} \right]. \label{eq:main_shifted_grad_neg}
    \end{align}
\end{theorem}

In case, we do not shift, this simply means $\overline{\alpha}_j = 1$.
Observe that, while not depending on shifts in the forward pass the stable gradients with respect to $g$ and $h$ do depend on the shifts in the backward pass and also on how we split the input $x$ to the positive and negative stream inputs $z^{(0,+)}$ and $z^{(0,-)}$.
In our definition, we choose to split them half-half, resulting in a half-half distribution of $\delta$ among $\delta^g$ and $\delta^h$.

\begin{remark}[Shifting]
    As shown in the following proof, 
    the gradient $\frac{\partial Q[j]}{\partial x}$ in \cref{eq:main_shifted_grad_pos} 
    captures the contribution of the split information flow in layer~$j$.
    We observe that $\frac{\partial Q[j]}{\partial x}$ tends to grow rapidly as $j$ approaches~$0$.
    Consequently, the shifting mechanism provides an effective lever to prevent gradient explosion 
    and to regulate the influence of split related gradient information originating from the different layers.
\end{remark}

\begin{proofref}{Theorem}{thm:closed_formula_shift}
    As in \cref{thm:closed_formula}, we want to prove the following claim by induction on $l$.
   \begin{align}
        \delta^{(l, g)}
            &= \frac{1}{2} \left[ \delta^{(l)}- \sum_{l \leq j < L} \ \prod_{j \leq i \leq L} \left( 1 - 2\alpha^{(i)} \right) \ W_{\text{abs}}^{[j+1, \rightarrow]} \cdot \left( M^{(j+1,-)}W^{(j+1)} \right) \cdot \prod_{i = (j-1),\dots, l} W^{(i+1, x)} \right], \\
        \delta^{(l, h)}
            &= -\frac{1}{2} \left[ \delta^{(l)} + \sum_{l \leq j < L} \ \prod_{j \leq i \leq L} \left( 1 - 2\alpha^{(i)} \right) \ W_{\text{abs}}^{[j+1, \rightarrow]} \cdot \left( M^{(j+1,-)}W^{(j+1)} \right) \cdot \prod_{i = (j-1),\dots, l} W^{(i+1, x)} \right].
    \end{align}
    This in the case $l=0$, together with \cref{eq:special_network_derivative} yields the formula stated in the Theorem.
    We defined
    \begin{align}
        \delta_{\text{intermediate}}^{(l,+,g)}
            &= \delta_{\text{shift}}^{(l+1,+,g)} \left( M^{(l+1, +)} W^{(l+1, +)} + M^{(l+1, -)} W^{(l+1, -)} \right)
                + \delta_{\text{shift}}^{(l+1,-,g)} W^{(l+1, -)} \label{eq:delta_shift_pos_g} \\
        \delta_{\text{intermediate}}^{(l,-,g)}
            &= \delta_{\text{shift}}^{(l+1,+,g)} \left(M^{(l+1, +)} W^{(l+1, -)} + M^{(l+1, -)} W^{(l+1, +)} \right)
                + \delta_{\text{shift}}^{(l+1,-,g)} W^{(l+1, +)} \label{eq:delta_shift_neg_g}
    \end{align}
    
    Similarly to the calculation in \cref{eq:delta_g_recursion} we obtain
    \begin{align}
        2 \cdot \delta_{\text{shift}}^{(l,g)}
            &\coloneqq \delta_{\text{shift}}^{(l,+,g)} - \delta_{\text{shift}}^{(l,-,g)} \notag\\
            &= \delta_{\text{intermediate}}^{(l,+,g)} - \delta_{\text{intermediate}}^{(l,-,g)} \notag\\
            &= \left( \delta_{\text{shift}}^{(l+1,+,g)} - \delta_{\text{shift}}^{(l+1,-,g)} \right) \cdot W^{(l+1, x)}
                - \left( \delta_{\text{shift}}^{(l+1,+,g)} + \delta_{\text{shift}}^{(l+1,-,g)} \right) M^{(l+1, -)} W^{(l+1)} \label{eq:delta_g_shif}
    \end{align}
    
    To simplify the expression we use the following Claim.
    \begin{claim} \label{claim:sum_delta_shif}
        \begin{align}
            \delta_{\text{shift}}^{(l,+,g)} + \delta_{\text{shift}}^{(l,-,g)}
                = \prod_{l \leq j \leq L} \left( 1 - 2 \alpha^{(j)} \right) W_{\text{abs}}^{[l, \rightarrow]}.
        \end{align}
    \end{claim}

    \begin{proofref}{Claim}{claim:sum_delta_shif}
        The induction base solves
        \begin{align}
            \delta_{\text{shift}}^{(L,+,g)} + \delta_{\text{shift}}^{(L,-,g)}
                &= \left( 1 - 2 \alpha^{(L)} \right) \left( \delta_{\text{intermediate}}^{(L,+,g)} + \delta_{\text{intermediate}}^{(L,-,g)} \right) \notag\\
                &= \left( 1 - 2 \alpha^{(L)} \right) \left( \delta^{(L,+,g)} + \delta^{(L,-,g)} \right) \notag\\
                &= \left( 1 - 2 \alpha^{(L)} \right) \mathbb{1} \notag\\
                &= \left( 1 - 2 \alpha^{(L)} \right) W_{\text{abs}}^{[L,\rightarrow]}.
        \end{align}
        For $0 \leq l < L$ adding \cref{eq:delta_shift_pos_g} and \cref{eq:delta_shift_neg_g} yields
        \begin{align}
            \delta_{\text{intermediate}}^{(l,+,g)} + \delta_{\text{intermediate}}^{(l,-,g)}
                &= \delta_{\text{shift}}^{(l+1,+,g)} \left( M^{(l+1, +)} + M^{(l+1, -)} \right) \left|W^{(l+1)}\right| + \delta_{\text{shift}}^{(l+1,-,g)} \left|W^{(l+1)}\right| \notag\\
                &= \left( \delta_{\text{shift}}^{(l+1,+,g)} + \delta_{\text{shift}}^{(l+1,-,g)} \right) \left|W^{(l+1)}\right| \label{eq:sum_shifted}
        \end{align}
        
        Now using \cref{eq:sum_shifted}, induction yields
        \begin{align}
            \delta_{\text{shift}}^{(l,+,g)} + \delta_{\text{shift}}^{(l,-,g)}
                &= \left( 1 - 2 \alpha^{(l)} \right) \left( \delta_{\text{intermediate}}^{(l,+,g)} + \delta_{\text{intermediate}}^{(l,-,g)} \right) \notag\\
                &= \left( 1 - 2 \alpha^{(l)} \right) \left( \delta_{\text{shift}}^{(l+1,+,g)} + \delta_{\text{shift}}^{(l+1,-,g)} \right) \left|W^{(l+1)}\right| \notag\\
                &= \left( 1 - 2 \alpha^{(l)} \right) \prod_{l+1 \leq j \leq L} \left( 1 - 2 \alpha^{(j)} \right) W_{\text{abs}}^{[l+1, \rightarrow]} \left|W^{(l+1)}\right| \notag\\
                &= \prod_{l \leq j \leq L} \left( 1 - 2 \alpha^{(j)} \right) W_{\text{abs}}^{[l, \rightarrow]}.
        \end{align}
        which concludes the proof of Claim \ref{claim:sum_delta_shif}.
    \end{proofref}

    Using the Claim and \cref{eq:delta_g_shif} we obtain the recursion
    \begin{align}
        2 \cdot \delta_{\text{shift}}^{(l,g)}
            = 2 \cdot \delta_{\text{shift}}^{(l+1, g)}
                - \prod_{l+1 \leq j \leq L} \left( 1 - 2 \alpha^{(j)} \right) W_{\text{abs}}^{[l+1, \rightarrow]} M^{(l+1, -)} W^{(l+1)} \label{eq:delta_g_shif_recursion}
    \end{align}

    Unfolding this recursion one by one as in the inductive proof of \cref{thm:closed_formula}, this closes the proof of \cref{thm:closed_formula_shift}.
\end{proofref}